\documentclass[1t0pt]{article}

\usepackage[margin=1in]{geometry}
\usepackage{graphicx}
\usepackage{amsmath,amsfonts,amssymb,amsthm}
\usepackage{bm}
\usepackage{color}
\usepackage{setspace}
\usepackage{listings}
\usepackage[font=small, labelfont=bf]{caption}
\usepackage{subcaption}
\usepackage{parskip}
\usepackage{float}
\usepackage{booktabs}
\usepackage[symbol]{footmisc}

\definecolor{dkgreen}{rgb}{0,0.6,0}
\definecolor{gray}{rgb}{0.5,0.5,0.5}
\definecolor{mauve}{rgb}{0.58,0,0.82}

\numberwithin{equation}{section}

\usepackage[toc,page]{appendix}
\usepackage{soul}
\usepackage{url}
\usepackage{physics}
\usepackage{multirow}
\usepackage{enumitem}
\usepackage[dvipsnames]{xcolor}
\usepackage[section]{placeins}
\usepackage{algorithm}
\usepackage[noend]{algpseudocode}
\usepackage{multicol}

\usepackage{hyperref}
\usepackage{adjustbox}

\newtheorem{remark}{Remark}

\begin{document}
\title{Neural Green's Operators for \\Parametric Partial Differential Equations}

\author{H.A. Melchers$^*$ \\
	Department of Mathematics and Computer Science\\
	Eindhoven University of Technology\\
	\texttt{h.a.melchers@tue.nl}\\
    \and 
	J.H.M. Prins$^*$ \\
	Department of Applied Physics\\
	Eindhoven University of Technology\\
	\texttt{j.h.m.prins@tue.nl}\\
   \and 
	M.R.A. Abdelmalik \\
	Department of Mechanical Engineering\\
	Eindhoven University of Technology\\
	\texttt{m.abdel.malik@tue.nl}
}
\date{}
\maketitle

\section*{Abstract}\footnotetext{$^*$ These authors contributed equally to this work}
This work introduces a paradigm for constructing parametric neural operators that are derived from finite-dimensional representations of Green's operators for linear partial differential equations (PDEs). We refer to such neural operators as \emph{Neural Green's Operators} (NGOs). Our construction of NGOs preserves the linear action of Green's operators on the inhomogeneity fields, while approximating the nonlinear dependence of the Green's function on the coefficients of the PDE using neural networks that take weighted averages of such coefficients as input. This construction reduces the complexity of the problem from learning the entire solution operator and its dependence on all parameters to only learning the Green's function and its dependence on the PDE coefficients. Moreover, taking weighted averages, rather than point samples, of input functions decouples the network size from the number of sampling points, enabling efficient resolution of multiple scales in the input fields. Furthermore, we show that our explicit representation of Green's functions enables the embedding of desirable mathematical attributes in our NGO architectures, such as symmetry, spectral, and conservation properties. Through numerical benchmarks on canonical PDEs, we demonstrate that NGOs achieve comparable or superior accuracy to Deep Operator Networks, Variationally Mimetic Operator Networks, and Fourier Neural Operators with similar parameter counts, while generalizing significantly better when tested on out-of-distribution data. For parametric time-dependent PDEs, we show that NGOs that are trained on a single time step can produce pointwise-accurate dynamics in an auto-regressive manner over arbitrarily large numbers of time steps. For parametric nonlinear PDEs, we demonstrate that NGOs trained exclusively on solutions of corresponding linear problems can be embedded within iterative solvers to yield accurate solutions, provided a suitable initial guess is available. Finally, we show that we can leverage the explicit representation of Green's functions returned by NGOs to construct effective matrix preconditioners that accelerate iterative solvers for PDEs.

\section{Introduction}
In this paper, we consider the problem of inferring the solution operator, associated with a parametric family of partial differential equations (PDEs), from data. Generally, the task of approximating such solution operators plays a fundamental role in the development of computational methods for solving the underlying PDEs \cite{hughes1998variational, hughes2012finite} that appear in various areas of science and engineering. Typical examples include computational fluid dynamics \cite{anderson1995computational}, fluid-structure interaction \cite{bazilevs2013computational} and solid mechanics \cite{fung2001classical}. However, finding analytical expressions of such solution operators is generally not feasible, and numerical approximations of such operators can be computationally prohibitive due to their high-dimensional and nonlinear nature \cite{hughes1998variational}. Recent advances in operator learning techniques \cite{lu2022comprehensive, sanderse2024} address this challenge by using data to approximate solution operators to PDEs. Typically, the data comprises instantiations of solutions to the PDE and the corresponding PDE input functions, such as boundary conditions, PDE coefficients and external forcings. The inference problem is to find the operator that maps the PDE parameters to the solution. If such a mapping is inferred, we are able to evaluate it and recover solutions of the PDE for different input parameters.

By virtue of their generalization of standard neural networks \cite{hornik1989multilayer} to mappings between function spaces instead of between discrete vector spaces, neural operators (NOs) \cite{lu2022comprehensive} provide a suitable representation of PDE solution maps. Representation of operators by NOs can be roughly divided into two categories. The first approach expands the output function in terms of the dot product between two vectors that are returned from two sub-networks, the so-called \emph{branch} and \emph{trunk} networks, which accept as arguments samples of the input functions and spatial coordinates, respectively. The second approach adds an integral operator to the discrete affine transformation in a standard neural network which allows application of the NO to varying resolutions of the input and output functions. The former construction is due to work in \cite{chen1995universal} and was extended to Deep Operator Networks (DeepONets) in \cite{lu2021learning}, while the latter construction was developed for different choices of the nonlocal kernel functions such as the Fourier \cite{li2020fourier}, multipole \cite{li2020multipole} and graph \cite{kovachki2023neural}, neural operators (FNOs, MNOs and GNOs, respectively). Both such approaches received considerable research attention, and significant work has been devoted to their extensions and improvements \cite{lu2022comprehensive}. Moreover, both approaches have been successfully applied to approximating solutions to PDEs in various fields, such as PDEs for fluid mechanics \cite{kochkov2021machine}, fluid-structure-interaction \cite{xiao2024fourier}, electromagnetism \cite{backmeyer2024solving}, solid mechanics \cite{faroughi2024physics} and quantum mechanics \cite{he2022fourier}. While such NOs generalize well to unseen PDE parameters within the training distribution, extrapolation beyond that distribution may become unreliable \cite{zhu2023reliable}. The lack of reliability in extrapolation may preclude the adoption of NOs, particularly when training data may be limited, e.g. in applications where the PDE parameters vary across length and time scales while training data may only cover a subset of those scales. Moreover, such generalization errors may be compounded when such NOs are used in an auto-regressive manner, such as predicting long-time behavior when trained on short trajectories \cite{mccabe2023towards, mandl2025physics, bonev2023spherical}. Furthermore, standard application of such NOs \cite{patel2024variationally,lu2021learning,li2020fourier} require point samples of the PDE parameters as input which scales unfavorably with the number of trainable parameters. This may result in inefficient models for multi-scale systems where a dense sampling of input functions is required. This shortcoming can be addressed by constructing neural operators that act on functions in a discretization-independent manner to improve the interpretability and accuracy of the model \cite{bahmani2025resolution, huang2025resolution, bartolucci2023neural}.

In this work, we turn our attention to linear PDEs where the corresponding solution operator reduces to the Green's operator with the Green's function as the corresponding kernel \cite{riley2006mathematical}. We introduce a paradigm for constructing NOs that are derived from a finite-dimensional approximation of parametric Green's operators with a learnable Green's function. We refer to such NOs as \emph{Neural Green's Operators} (NGOs). Our construction of NGOs reduces the complexity of the operator learning problem by preserving the correct linear dependence of the solution on the PDE's inhomogeneity fields (such as forcing and boundary conditions). Moreover, taking weighted averages, rather than point samples, of input functions decouples the network size (in terms of the number of trainable parameters) from the number of sampling points, enabling efficient resolution of multiple scales in the input fields. Furthermore, our construction of NGOs is conducive to including additional mathematical structure, or inductive bias \cite{karniadakis2021physics}, to preserve salient properties of the solutions to the PDE, like symmetries, spectral properties and conservation laws. We demonstrate that:
\begin{enumerate}
    \item NGOs enable efficient approximations of PDE solutions in the sense that for roughly the same number of trainable network parameters, NGOs are on par with or more accurate than DeepONets, VarMiONs, and FNOs, and generalize significantly better to out-of-distribution problems.
    \item  NGOs that are trained on a single time step can produce pointwise-accurate dynamics of time-dependent PDEs in an auto-regressive manner over arbitrarily large numbers of time steps.
    \item NGOs trained exclusively on linearized parametric PDEs can be embedded within iterative nonlinear solvers to yield accurate solutions to associated nonlinear problems, provided a suitable initial guess is available.
    \item The Green's functions inferred from NGOs can be used to construct effective matrix preconditioners for numerical PDE solvers.
\end{enumerate}

\subsection{Other Related Work}
So far, our discussion has focused on neural operators for families of PDEs indexed by their coefficients, and the placement of neural Green's operators in such a class of networks. Since NGOs learn solution operators to parametric PDEs by mimicking the mathematical construction of Green's operators in the neural architecture, NGOs also learn the associated parametric Green's functions. This generalization to the parametric setting  distinguishes our work from prior research devoted to inferring Green’s functions \cite{boullé2022data, dai2024numerical, gin2021deepgreen, gu2025explainable, peng2023deep, teng2022learning, ichimura2020fast}. By learning the parametrization of the Green’s function with respect to the PDE coefficients, our learned Green's operator can be used as a versatile numerical building block; as demonstrated in Sections \ref{S time dependent diffusion} and \ref{S nonlinear}, the trained neural Green's operator can be embedded into iterative algorithms for time-dependent and nonlinear problems. For the problems considered in this work, such tasks would be infeasible with non-parametric Green's functions. Furthermore, we show in Section \ref{S Precon} that the inferred parametric Green's functions can be used to construct classical matrix preconditioners for parametric linear systems. Unlike hybrid neural approaches to preconditioning \cite{azulay2023multigrid, cui2025hybrid, lerer2024multigrid, zhang2024blending, kopaničáková2025deeponet, xiang2025unsupervised}, our learned matrix preconditioner can be applied algebraically to any Krylov subspace method. In the remainder of this section we review such related work, both to situate our contribution within the broader literature and to emphasize the distinctive capabilities enabled by our formulation.

Significant research work has been devoted to inferring an explicit representation of Green's functions for PDEs, for example in \cite{boullé2022data},\cite{peng2023deep}, \cite{teng2022learning} and \cite{ichimura2020fast}, Green's functions of linear PDEs are represented by neural networks; in \cite{gin2021deepgreen} autoencoders are used to infer a matrix representation of the linearization of a PDE operator, where the inverse of such a matrix represents Green's function; in \cite{gu2025explainable} a Green's function to a target PDE is represented by a trunk network of a DeepONet \cite{lu2021learning}, while the branch network of the DeepONet is used to approximate the auxiliary gradients of the Green’s function. In \cite{dai2024numerical} network architectures that are inspired by Green's function formulations are proposed to address challenges faced by some neural operators in accounting for boundary conditions.  Despite these advances, the aforementioned approaches do not account for the (generally nonlinear) dependence of the Green’s function on the PDE coefficients and are therefore restricted to fixed PDE operators. An exception is presented in \cite{praveen2023principled}, where low-rank approximations of Green's functions for fixed PDE coefficients are inferred from data, subsequently interpolating these representations across the PDE coefficient space. Nevertheless, this framework is restricted to self-adjoint operators, and its accuracy is critically contingent upon the regularity of the Green's function and its spectral decomposition relative to the PDE coefficients. Therefore, a general and flexible framework for learning Green’s functions that vary parametrically with the PDE coefficients remains lacking. Addressing this gap is the central focus of the present work.

Apart from predicting approximate solutions to PDEs, neural operators can also be used to accelerate iterative methods for PDEs. For example, Ruelmann et al.~\cite{ruelmann2018prospects} propose a neural network that maps system matrices to approximate inverses, thereby accelerating the solution of linear systems; however, this approach is limited to problems of a fixed size. Related efforts seek to learn mappings that output incomplete factorizations while preserving the sparsity pattern of the input matrix \cite{häusner2024neural, li2023learning, li2024deep, sappl2019deep}, though these methods are, as presented there, restricted to matrices that are symmetric and positive definite. A distinct category of hybrid preconditioning, exemplified by the HINTS framework~\cite{hu2025hybrid, zhang2024blending, kahana2023geometry}, integrates a solution-approximating neural operator directly into a fixed-point iteration. Similarly, neural operators have been embedded as preconditioners within Krylov-subspace solvers~\cite{xiang2025unsupervised, shpakovych2023neural, lerer2024multigrid, azulay2023multigrid}. A significant attribute of these existing works is their reliance on nonlinear models to approximate the action of what is fundamentally a linear operator, thus often meaning that Flexible GMRES (F-GMRES, \cite{saad1993flexible}) is the only Krylov method for which the approach is applicable. In contrast, this work proposes algorithms for constructing matrix preconditioners that preserve the linear structure of classical preconditioners, thereby ensuring algebraic compatibility with the full range of standard Krylov subspace methods.

The remainder of this paper is organized as follows. Section \ref{S Green's operators} establishes the mathematical foundation of Green's operators for parametric PDEs and introduces their corresponding finite-dimensional representations. In Section \ref{S The Neural Green's Operator}, we introduce the Neural Green’s Operator (NGO) and present architectural variants tailored to diverse modeling scenarios. Section \ref{S Test problem: steady diffusion} details the canonical elliptic PDE used as our primary benchmark, including the associated data generation and training protocols. Our core findings are presented in Section \ref{S Results}, where we evaluate model accuracy, computational efficiency, and the influence of critical hyperparameters such as quadrature schemes, basis function cardinality, and total parameter count. This section also demonstrates the utility of the learned Green's functions in constructing robust matrix preconditioners. To assess the framework's versatility, Section \ref{S Extend} explores the application of NGOs to non-elliptic, transient, and nonlinear problems. Finally, Section \ref{S Conclusion} provides a summary of our contributions and a discussion of future research directions.

\section{Parametric Green's Operators and Functions}\label{S Green's operators}
In this paper, we consider a linear boundary value problem in a bounded domain $\Omega\subset\mathbb{R}^d$ with a piecewise smooth boundary $\Gamma = \bigcup_{i} \Gamma_i$:
\begin{equation}\label{E PDE sbvp}
    \begin{aligned}
        \mathcal{L}[\theta]u(\vb{x}) &= f(\vb{x}), &&\forall\vb{x}\in \Omega, \\
        \mathcal{B}_i[\theta] u(\vb{x}) &=g_i(\vb{x}), &&\forall\vb{x}\in \Gamma_i,
    \end{aligned}
\end{equation}
where $u(\vb{x})$ denotes the unknown, $\mathcal{L}[\theta]$ is a linear differential operator, parameterized by some spatially varying coefficient $\theta(\vb{x})$, $f(\vb{x})$ is a source term, $\mathcal{B}_i[\theta]$ is a boundary operator, and $g_i$ is the data on the boundary $\Gamma_i$.

To find an expression for the unknown $u(\vb{x})$, we write \eqref{E PDE sbvp} in weak form by testing the first equation of \eqref{E PDE sbvp} against some suitable function $v$:
\begin{equation}\label{E PDE test}
    \int_\Omega v \mathcal{L}[\theta] u \, d\vb{x} = \int_\Omega v f \, d\vb{x}.
\end{equation}
Using integration by parts and substituting the boundary conditions we may arrive at the formulation
\begin{equation}\label{E final weak form sbvp}
    B[v,u] = L[v],
\end{equation}
where 
\begin{equation}
    B[v,u] = \int_\Omega u \mathcal{L}^*[\theta] v \, d\vb{x}
    -  \sum_i \int_{\Gamma \setminus \Gamma_i} (\mathcal{B}_i[\theta] u)(\tilde{\mathcal{B}}_i v) \, d\vb{x},
\end{equation}
and
\begin{equation}
    L[v] = \int_\Omega v f \, d\vb{x}
    + \sum_i\int_{\Gamma_i} g_i \tilde{\mathcal{B}}_i v \, d\vb{x}.
\end{equation}
Here, $\mathcal{L}^{*}$ is the adjoint operator of $\mathcal{L}$, and $\tilde{\mathcal{B}}_i$ are boundary operators that appear in the integration by parts. For example derivations of this form, see Appendix~\ref{sec:greens-function-derivation}.

We can derive an expression for the unknown $u(\vb{x})$ by setting $v$ in \eqref{E final weak form sbvp} to be the Green's function $G[\theta](\vb{x},\vb{x}')$ that satisfies
\begin{equation}\label{E definition G bvp}
    \begin{aligned}
        \mathcal{L}^*[\theta]G[\theta](\vb{x},\vb{x}') &= \delta(\vb{x} - \vb{x}'), &&\forall\vb{x}\in \Omega, \\
        \tilde{\mathcal{B}}_i G[\theta](\vb{x},\vb{x}') &= 0, &&\forall \vb{x}\in \Gamma\setminus\Gamma_i.
    \end{aligned}
\end{equation}
Then, \eqref{E final weak form sbvp} reduces to
\begin{equation}\label{E Green's operator BVP}
    u(\vb{x})= \mathcal{G}[\theta, f, g_i](\vb{x}) := \int_\Omega G[\theta]f \, d\vb{x}' +  \sum_i\int_{\Gamma_i} g_i \tilde{\mathcal{B}}_i G[\theta] \, d\vb{x}',
\end{equation}
and $\mathcal{G}$ denotes the Green's operator that maps the PDE coefficient $\theta(\vb{x}')$, forcing $f(\vb{x}')$ and boundary data $g_i(\vb{x}')$ onto the solution $u(\vb{x})$ \cite{riley2006mathematical}.

To derive a finite-dimensional approximation of the solution, we consider an expansion of the Green's function in terms of a test basis $\psi_n(\vb{x}')$ and trial basis $\phi_m(\vb{x})$:
\begin{equation}\label{E Galerkin approximation Green's function}
    G[\theta](\vb{x},\vb{x}') \approx \phi_m(\vb{x})A_{mn}[\theta]\psi_n(\vb{x}'),
\end{equation}
where $A_{mn}$ depends nonlinearly on $\theta$.
Substitution of \eqref{E Galerkin approximation Green's function} into  \eqref{E Green's operator BVP} gives the corresponding approximation of the Green's operator
\begin{equation}\label{E Galerkin approximation Green's operator}
    \mathcal{G}[\theta,f,g_i](\vb{x}) \approx \phi_m(\vb{x})A_{mn}[\theta] d_n[f, g_i] ,
\end{equation}
where 
\begin{equation}\label{E d}
    d_n[f,g_i] = \int_\Omega \psi_n f d\vb{x}' + \sum_i\int_{\Gamma_i} g_i \tilde{\mathcal{B}}_i \psi_n \, d\vb{x}'.
\end{equation}
 
\begin{remark}
In a Galerkin approximation setting, the matrix $A_{mn}$ in \eqref{E Galerkin approximation Green's function} represents the inverse of the system matrix. 
\end{remark}
\begin{remark}
While our approximation of the Green's function in \eqref{E Galerkin approximation Green's function} conforms to a Petrov-Galerkin method where the test functions $\psi$ and trial functions $\phi$ are not the same, we may opt for a (Bubnov-)Galerkin setting for our NGO, where $\phi$ and  $\psi$ are the same.
\end{remark}

\section{Neural Green's Operators (NGOs)}\label{S The Neural Green's Operator}
\begin{figure}[htb]
\centering
\includegraphics[width = 0.75\textwidth]{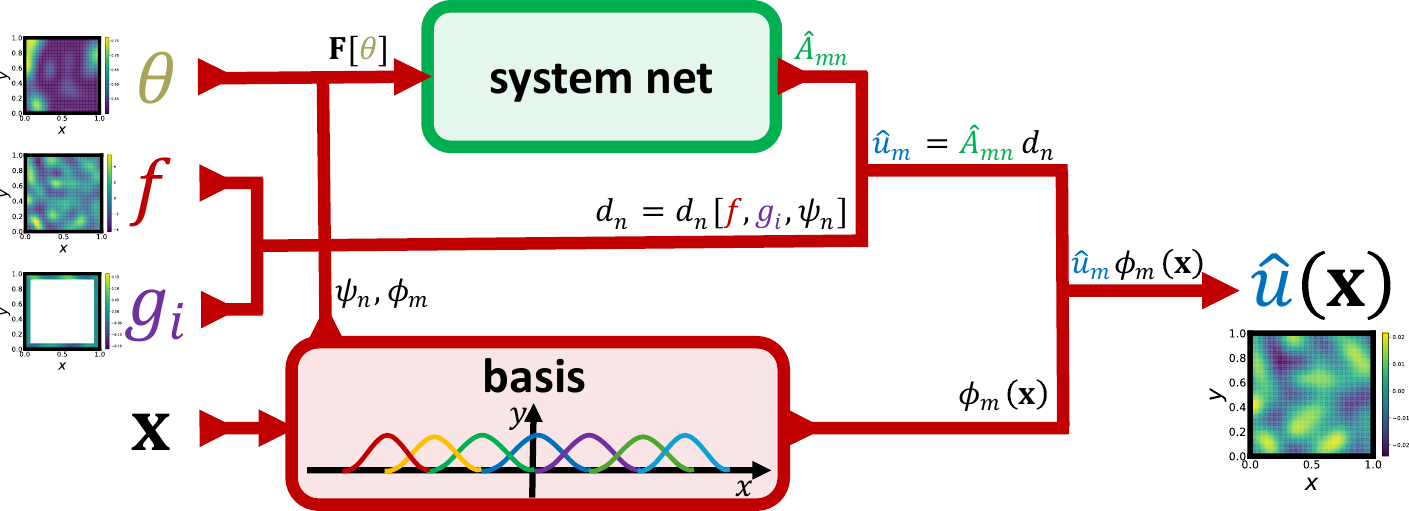}
\caption{\small{Architecture of the neural Green's operator (NGO), that maps the material parameter $\theta(\vb{x})$, forcing $f(\vb{x})$ and boundary conditions $g_i(\vb{x})$ onto the solution $u(\vb{x})$. For a model NGO, $\vb{F}$ is given by \eqref{E MNGO}, whereas for a data NGO, $\vb{F}$ is given by \eqref{E DNGO}. The system network, shown in blue, is the only trainable component in the NGO.}}
\label{F NGO}
\end{figure}

We now define a Neural Green's Operator, based on the approximation in \eqref{E Galerkin approximation Green's function} with a learnable matrix, and fixed test functions $\psi(\vb{x}')$ and trial functions $\phi(\vb{x})$:
\begin{equation}\label{E NGO}
    \hat{u}(\vb{x}) = \hat{\mathcal{G}}[\theta,f,g_i](\vb{x}) \equiv \phi_m(\vb{x})\hat{A}_{mn}(\vb{F}[\theta]) d_n[f, g_i],
\end{equation}
where `hats' indicate quantities approximated by a neural network, and $d_n$ is given by \eqref{E d}. The NGO is of the family of Chen\&Chen \cite{chen1995universal} neural operator architectures (DeepONets, VarMiONs), in the sense that the NGO utilizes an approximation of the solution in terms of basis functions $\phi_m(\vb{x})$ and coefficients $\hat{u}_m\equiv\hat{A}_{mn} d_n$. We summarize the NGO architecture in Figure \ref{F NGO}.
$\vb{F}[\theta]$ is the input to the system net, which is some functional of $\theta$. Based on the specific form of $\vb{F}[\theta]$, and the way the model is trained, we distinguish between three types of NGOs. 

\subsection{Different types of NGOs}
\label{S Different types of NGOs} 

\subsubsection{Model NGO}
The first type is the model NGO, where we set
\begin{equation}\label{E MNGO}
    F_{nm}[\theta] = \int_\Omega \phi_m \mathcal{L}^*[\theta] \psi_n \, d\vb{x}'
    -  \int_{\Gamma \setminus \Gamma_i} (\mathcal{B}_i[\theta] \phi_m)(\tilde{\mathcal{B}}_i\psi_n) \, d\vb{x}',
\end{equation}
which represents the system matrix corresponding to a discretization of \eqref{E final weak form sbvp}. A model NGO is trained on the loss function
\begin{equation}\label{E solution norm}
    L[\hat{u},u] = \norm{\hat{u} - u},
\end{equation}
where $\norm{.}$ is the norm of choice in which the user wants the model to be optimal (throughout this work, we use the \(L^2\) norm).
A model NGO can be used when the target PDE \eqref{E PDE sbvp} is known, and when training data is available. The theoretical lower bound of the error that an NGO may attain is given by $\norm{u^{\mathrm{p}}-u}$, where $u^{\mathrm{p}}$ is the projection of $u$ onto $\phi_m(\vb{x})$ in the norm $\norm{.}$.

\subsubsection{Data-free NGO}\label{S Data-free NGO}
The data-free NGO can be used in the case where the target PDE is known, but solution data of the PDE is unavailable (for example, because simulations are too expensive to do). The data-free NGO exploits the similarity between an NGO and a finite element method (FEM), that would approximate $\hat{\vb{A}}$ as $\hat{\vb{A}}\approx\vb{F}^{-1}$. The input to the system net of a data-free NGO is still the system matrix, given by \eqref{E MNGO}. However, the data-free NGO is trained on a loss function
\begin{equation}\label{E matrix norm}L[\hat{\vb{A}}]=\norm{\vb{F}\hat{\vb{A}}\vb{F} - \vb{F}},
\end{equation}
where $\norm{.}$ is a matrix norm of choice (we use the Frobenius norm). This loss function does not include solution data $u$, but instead, it steers the system net towards learning the pseudoinverse $\vb{F}^+$ of the matrix $\vb{F}$ ($\vb{F}^+=\vb{F}^{-1}$ if $\vb{F}$ is invertible). The lower bound of $L[\hat{\vb{A}}]$ is given by $L[\hat{\vb{A}}=\vb{F}^{-1}]=0$, which would correspond with the NGO predicting $\hat{u}=u^h$, where $u^h$ is the solution provided by a FEM. Note that $\norm{u^h-u}\geq \norm{u^{\mathrm{p}}-u}$ in the solution norm \eqref{E solution norm}. In other words, a data-free NGO has a higher error bound than a model NGO.

\subsubsection{Data NGO}
Lastly, we consider a third type of NGO, referred to as a data NGO, which may be used if the PDE that underlies the training data is not known, but solution training data is available (for example, when dealing with experimental data). The input to the system net of a data NGO is given by
\begin{equation}\label{E DNGO}
    F_{n}[\theta] = \int_\Omega \psi_n \theta d\vb{x}',
\end{equation}
and it is trained on the solution loss as given by \eqref{E solution norm}. Just like for a model NGO, the data NGO error lower bound in the norm of \eqref{E solution norm} is given by $\norm{u^{\mathrm{p}}-u}$.

\begin{figure}[htb]
\centering
\includegraphics[width = 0.75\textwidth]{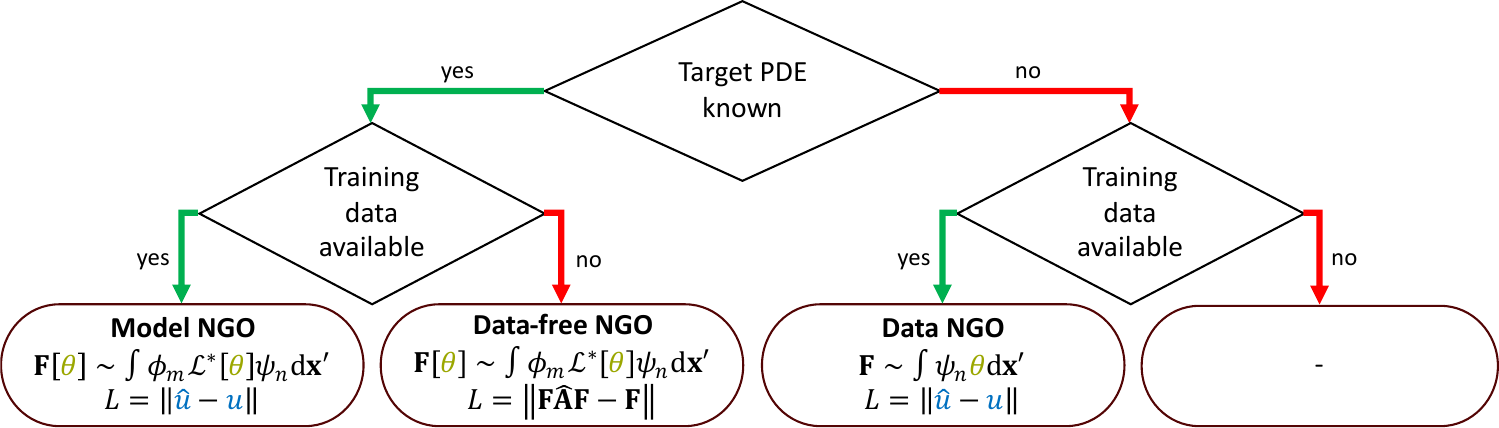}
\caption{\small{The use cases and characteristics of the model NGO, data-free NGO and data NGO.}}
\label{F modeldatafreedata}
\end{figure}

\begin{figure}[htb]
    \centering
    \includegraphics[width=0.25\linewidth]{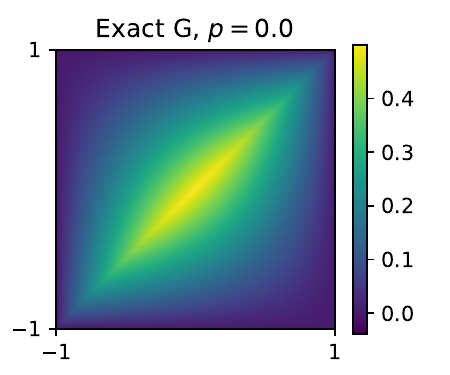}
    \includegraphics[width=0.25\linewidth]{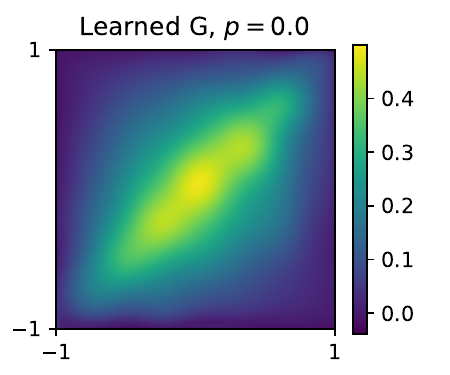}
    \includegraphics[width=0.25\linewidth]{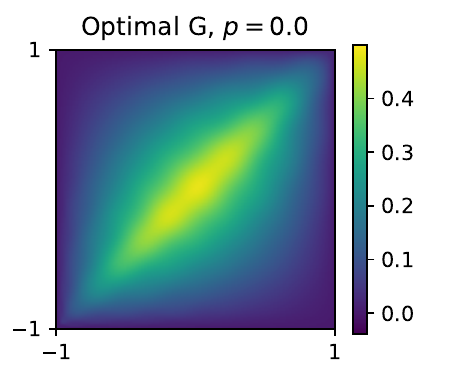} \\
    \includegraphics[width=0.25\linewidth]{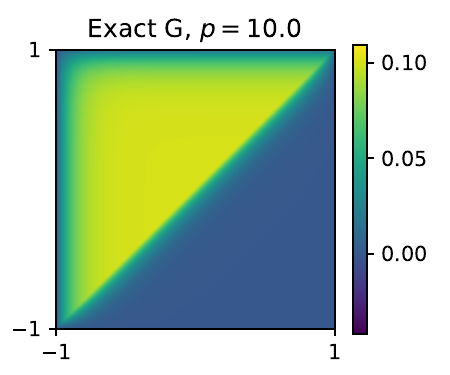}
    \includegraphics[width=0.25\linewidth]{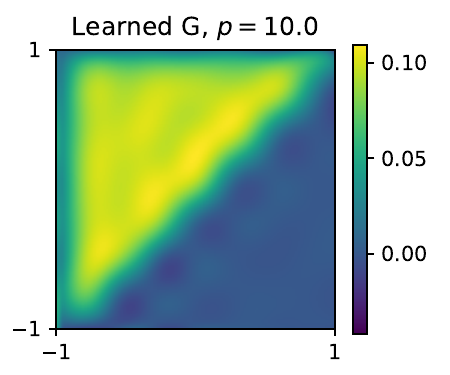}
    \includegraphics[width=0.25\linewidth]{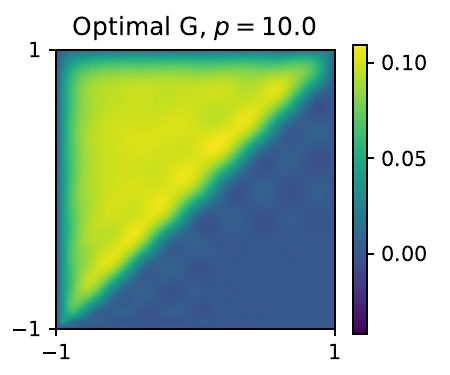}
    \caption{Left: the exact Green's function for the 1D advection-diffusion equation \(-u'' + pu' = f\), for two different values of the advection speed \(p\). Middle: the corresponding approximations to these Green's functions, approximated by an NGO. Right: the true Green's functions projected onto the NGO's basis, showing the best possible approximation of the Green's function by an NGO.}
    \label{fig:greens-function-example-1D}
\end{figure}

The use cases and characteristics of the model, data-free and data NGO are summarized in Figure \ref{F modeldatafreedata}.
An important property of all three types of NGO is that they learn an approximation to the Green's function in the form of Equation~\eqref{E Galerkin approximation Green's function}, which means that the learned approximation to the Green's function can be directly extracted from a trained model. Figure~\ref{fig:greens-function-example-1D} shows an example of the Green's function approximations as learned by an NGO. This figure shows exact Green's functions and their NGO approximations for a 1D PDE, and shows that NGOs typically learn a ``smoothed'' version of the Green's function, which is accurate in the low frequencies but less accurate in the high frequencies. NGOs are also not guaranteed to preserve properties of the original Green's function, such as positivity. The right-hand-side plots of Figure~\ref{fig:greens-function-example-1D} show that the optimal approximation, given by the projection of the Green's function onto a fixed basis, has the same limitations.

Table~\ref{tab:model-features} compares some key properties of the different models. NGOs are the only tested models for which (1) the input data can be given on an arbitrary  set of points (provided an accurate quadrature rule is available), (2) the output can be evaluated anywhere, and (3) the linear structure of the PDE is preserved. Note that other machine learning methods related to Green's functions \cite{boullé2022data, dai2024numerical, gin2021deepgreen, gu2025explainable, peng2023deep, teng2022learning}, while perhaps conceptually similar to NGOs, are not included in the comparison as those models learn solutions to a single linear or nonlinear PDE, with no dependence on PDE parameters such as varying diffusion coefficients or advection fields. As such, these models cannot be trained to approximate the mappings that NGOs and other included neural operators learn.
\begin{table}
    \centering
    \caption{An overview of some key features of different neural operators.}
    \label{tab:model-features}
    \begin{tabular}{l c c c}
        \toprule
        \phantom{[11] }Model & Input samples & Output samples & Linearity \\
        \midrule
        \cite{lu2021learning}
        DeepONet & Fixed samples       & Continuous    & No \\
        \cite{bahmani2025resolution}
        RINO     & Any quadrature rule & Continuous    & No \\
        \cite{patel2024variationally}
        VarMiON  & Fixed samples       & Continuous    & Yes \\
        \cite{li2020fourier}
        FNO      & Uniform grid        & Same as input & No \\
        \cite{ronneberger2015unet}
        U-Net    & Fixed uniform grid  & Same as input & No \\
        \cite{raonic2024convolutional}
        CNO      & Uniform grid        & Uniform grid  & No \\
        \midrule
        \phantom{[11]}
        NGOs     & Any quadrature rule & Continuous    & Yes \\
        \bottomrule
    \end{tabular}
\end{table}

\begin{remark}
While our NGO in \eqref{E NGO} uses fixed bases, we may opt to learn either the test function $\psi$ or the trial function $\phi$, or both.
\end{remark}

\begin{remark}[\textbf{Relation to CNOs}]
A neural operator architecture similar to NGOs is the Convolutional Neural Operator (CNO). The similarities and differences between NGOs and CNOs are explained in Appendix~\ref{appendix:relation-to-cnos}.
\end{remark}

\FloatBarrier
\section{Test Problem: Steady diffusion}\label{S Test problem: steady diffusion}
\begin{figure}[htb]
\centering
\includegraphics[width = 0.2\textwidth]{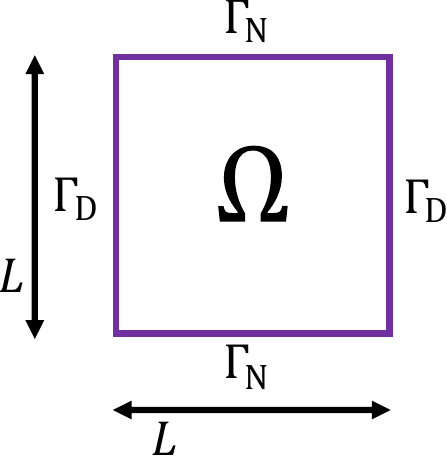}
\caption{The square domain $\Omega$, with Neumann boundaries $\Gamma_{\mathrm{N}}$ on the top and bottom, and Dirichlet boundaries $\Gamma_{\mathrm{D}}$ on the left and right. We used $L=1$.}
\label{F unitsquare}
\end{figure}

In this section, we compare the performance of the NGO with other models in literature, including DeepONets, VarMiONs, FNOs, CNOs, and U-Nets. The test problem is the steady diffusion equation defined in a unit square and parameterized by a spatially varying diffusion coefficient $\theta(\vb{x})$, the Neumann boundary conditions $\eta(\vb{x})$ on the upper and lower boundaries and the Dirichlet boundary conditions $g(\vb{x})$ on the left and right boundaries; The equations are given by
\begin{align}
    \begin{cases}
        \hfill -\nabla\cdot\left( \theta\nabla u \right) = f &\text{on } \Omega = (0, 1)^2, \\
        \hfill \theta \nabla u \cdot \vb{n} = \eta &\text{on } \Gamma_{\mathrm{N}} = (0, 1) \times \left\{ 0, 1 \right\}, \\
        \hfill \theta u = g &\text{on } \Gamma_{\mathrm{D}} = \left\{ 0, 1 \right\} \times (0, 1),
    \end{cases}
    \label{eq:darcy}
\end{align}
and a sketch of the geometry is given in Figure \ref{F unitsquare}.
The solution to \eqref{eq:darcy} can be expressed in terms of the Green's function as
\begin{equation}\label{E Green's operator Darcy}
    u(\vb{x}) = \int_\Omega G[\theta](\vb{x}, \vb{x}') f(\vb{x}') d\vb{x}' + \int_{\Gamma_{\mathrm{N}}} G[\theta](\vb{x}, \vb{x}') \eta(\vb{x}') d\vb{x}'  - \int_{\Gamma_{\mathrm{D}}} g \vu{n} \cdot \nabla_{\vb{x}'} G[\theta](\vb{x}, \vb{x}') d\vb{x}'.
\end{equation}
The derivation of Equation \ref{E Green's operator Darcy} can be found in Appendix \ref{sec:appendix-darcy-greens-function}.

\subsection{NGO architecture}\label{S NGO architecture steady diffusion}
Using the ansatz in \eqref{E Galerkin approximation Green's function} to approximate the Green's function in \eqref{E Green's operator Darcy}, with $\phi_m=\psi_m$, we can approximate the solution as
\begin{equation}\label{E GGO HC2D}
    u(\vb{x}) = \mathcal{G}[\theta,f,\eta,g](\vb{x}) \approx \phi_m(\vb{x})A_{mn}[\theta] d_n[f,\eta,g] ,
\end{equation}
where $d_n$ is given by
\begin{equation}\label{E d Darcy}
    d_n[f,\eta,g] = \int_{\Omega} \psi_n f d\vb{x}' + \int_{\Gamma_{\mathrm{N}}} \psi_n \eta d\vb{x}' - \int_{\Gamma_{\mathrm{D}}} g \vu{n} \cdot \nabla \psi_n d\vb{x}'.
\end{equation}
We define the NGO corresponding to \eqref{E GGO HC2D} as
\begin{equation}\label{E NGO HC2D}
    \hat{u}(\vb{x}) = \hat{\mathcal{G}}[\theta,f,\eta,g](\vb{x}) = \phi_m(\vb{x})\hat{A}_{mn}\left(\vb{F}[\theta]\right)d_n[f,\eta,g].
\end{equation}
Here, for model NGOs and data-free NGOs, $\vb{F}[\theta]$ is given by the system matrix 
\begin{equation}\label{E F MNGO sd}
F_{nm}[\theta]=-\int_\Omega \phi_m \nabla \cdot \theta \nabla \psi_n d\vb{x}' 
    - \int_{\Gamma\setminus \Gamma_{\mathrm{N}}} \theta \psi_n \vu{n}\cdot \nabla \phi_m d\vb{x}' 
    + \int_{\Gamma\setminus \Gamma_{\mathrm{D}}} \theta \phi_m \vu{n}\cdot \nabla \psi_n d\vb{x}', 
\end{equation}
which is equivalent to Equation \ref{E Darcy 2}. For data NGOs, $\vb{F}[\theta]$ is given by
\begin{equation}\label{E F DNGO sd}
    F_{n}[\theta] = \int_\Omega \psi_n  \theta d\vb{x}'.
\end{equation}
All neural operators were trained to learn the solution operator \(\left( \theta, f, \eta, g \right) \mapsto u\).
\begin{remark}\label{rem:nitsche}
The matrix $\vb{F}$ as given by \ref{E F MNGO sd} may not be coercive. Stabilization of such a formulation can be addressed using Nitsche's method\cite{Bazilevs_Hughes_2005} which adds the terms
\begin{equation}\label{E Nitsche F}
    F^{(\mathrm{s})}_{nm} = C_{\mathrm{s}}  \int_{\Gamma_{\mathrm{D}}} \psi_n \theta \phi_m d\vb{x}' 
\end{equation}
and
\begin{equation}\label{E Nitsche d}
    d^{(\mathrm{s})}_{n} = C_{\mathrm{s}}  \int_{\Gamma_{\mathrm{D}}} \psi_n g d\vb{x}'
\end{equation}
to the matrix $F_{nm}$ and right-hand-side vector $d_n$, respectively, where $C_\mathrm{s}>0$ is the Nitsche stabilization parameter. However, in our FEM tests in this section we observed that the stabilization is not needed. Therefore, we chose to not include the stabilization in the NGO formulation for the steady diffusion problem, because omitting it gave identical results, both for NGOs and FEM.
\end{remark}

\subsection{Inductive Bias: Preconditioning the System Net}\label{S Preconditioning the System Network}
\begin{figure}[htb]
\centering
\includegraphics[width = 0.75\textwidth]{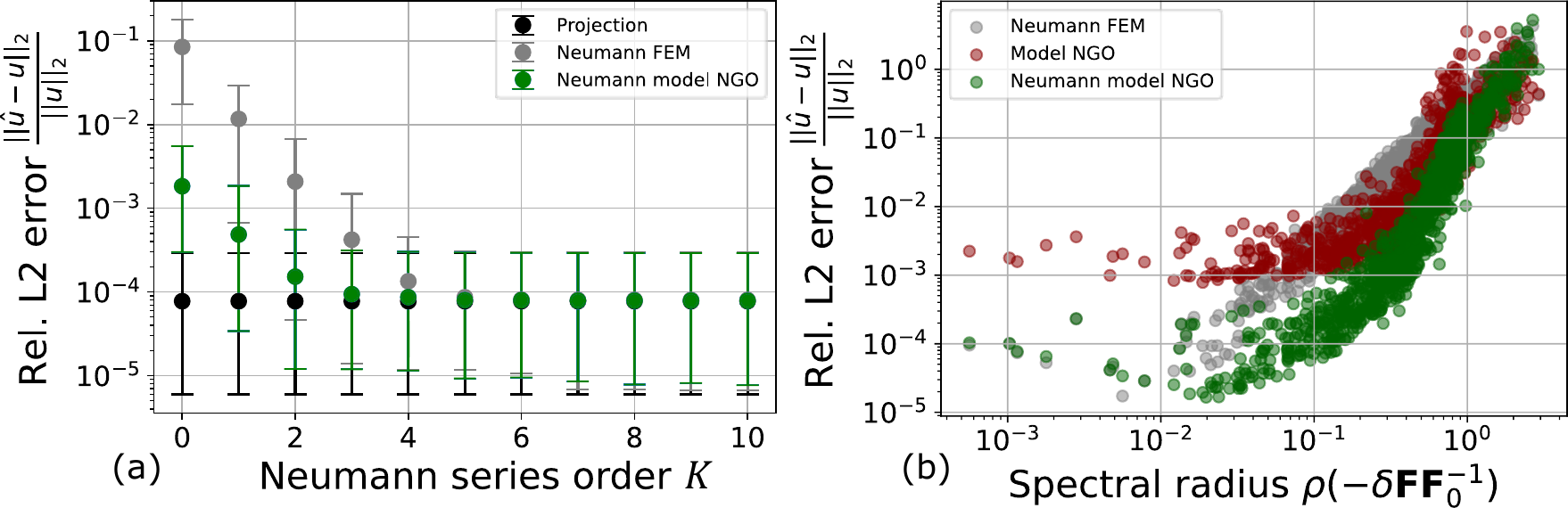}
\caption{\small{(a) Relative \(L^2\) test error versus the order $K$ of the truncated Neumann series of a FEM that uses an approximate matrix inversion using a Neumann series, and a Neumann model NGO as defined in \eqref{E Neumann series ansatz}.
(b) The relative \(L^2\) test error versus the spectral radius $\rho(-\delta \vb{F} \vb{F}_0^{-1})$ for a vanilla model NGO, and a Neumann model NGO. 
the models have been trained on dataset C (Appendix \ref{A data generation}). Points and error bars are, respectively, averages and 95\% confidence intervals on 1000 manufactured steady diffusion solutions, generated in the same way as dataset C.}}
\label{F neumannseries}
\end{figure}

The Green's formulation of the NGO allows us to introduce additional inductive bias that may ensure preservation of certain properties of the underlying PDE, and/or enhance model accuracy. In this section, we show an example of this, where we boost the NGO's accuracy by preconditioning the output of the system net. In Section \ref{S Data-free NGO}, we discussed the similarity between NGOs and finite element methods, and that finite element methods generally provide close to optimal approximations to the solution of a PDE. This suggests adding inductive bias to NGOs that steers the system net towards learning a matrix $\hat{\vb{A}}$ that is close to the inverse $\vb{F}^{-1}$ of the system matrix $\vb{F}$. For model and data-free NGOs, this can be done in the following way. Suppose that the material parameters $\theta(\vb{x})$ are centered around and reasonably close to some average value $\expval{\theta}\neq 0$. Then, we can split the material parameter as 
\begin{equation}
\theta(\vb{x})=\expval{\theta} + \delta \theta(\vb{x}),    
\end{equation}
and, correspondingly, the system matrix as
\begin{equation}
    \vb{F}[\theta] = \vb{F}_0 + \delta \vb{F},
\end{equation}
where $\vb{F}_0 \equiv \vb{F}[\expval{\theta}]$ and $\delta \vb{F} \equiv \vb{F}[\theta] - \vb{F}[\expval{\theta}]$. By factorizing out $\vb{F}_0$ as
\begin{equation}
    \vb{F}[\theta] = \left(\vb{I} + \delta \vb{F} \vb{F}_0^{-1}\right)\vb{F}_0
\end{equation}
and inverting the result, we get
\begin{equation}\label{E Neumann 1}
    \vb{F}^{-1}[\theta] = \vb{F}_0^{-1}\left(\vb{I} + \delta \vb{F} \vb{F}_0^{-1}\right)^{-1}.
\end{equation}
Equation \eqref{E Neumann 1} can be expressed in terms of a Neumann series as
\begin{equation}\label{E Neumann series}
    \vb{F}^{-1}[\theta] = \vb{F}_0^{-1}\sum_{k=0}^\infty \left(-\delta \vb{F}\vb{F}_0^{-1}\right)^k = \vb{F}_0^{-1}\left[\vb{I} + \left(-\delta \vb{F}\vb{F}_0^{-1}\right) + \left(-\delta \vb{F}\vb{F}_0^{-1}\right)^2 + \left(-\delta \vb{F}\vb{F}_0^{-1}\right)^3 +...\right],
\end{equation}
which converges if
\begin{equation}
    \rho\left(-\delta \vb{F}\vb{F}_0^{-1}\right) < 1,
\end{equation}
where $\rho$ is the spectral radius. Equation \eqref{E Neumann series} can be used as ansatz for $\hat{\vb{A}}$ as
\begin{equation}\label{E Neumann series ansatz}
    \hat{\vb{A}}\left(\vb{F}[\theta]\right) \approx \vb{F}_0^{-1}\left[ \sum_{k=0}^K \left(-\delta \vb{F}\vb{F}_0^{-1}\right)^k + \mathrm{NN}\left(-\delta \vb{F}\vb{F}_0^{-1}\right) \right],
\end{equation}
where NN is the system net. $\vb{F}_0^{-1}$ effectively acts as a preconditioner, and the system net within the Neumann series ansatz effectively learns a correction to a truncated Neumann series of order $K$. To minimize computational overhead, we choose $K=1$. Then, using the Neumann series ansatz in an NGO comes at the cost of computing a \emph{single, offline} matrix inverse $\vb{F}_0^{-1}$ and two online matrix-matrix products only. Figure \ref{F neumannseries}(a) shows that a Neumann model NGO effectively learns about two additional terms to a Neumann series truncated at order $K$, compared to a FEM that uses the same truncated Neumann series for approximate matrix inversion. Figure \ref{F neumannseries}(b) shows that the Neumann NGO error indeed scales with the spectral radius $\rho$ of $-\delta \vb{F} \vb{F}_0^{-1}$. However this turns out to be the case for both Neumann model NGOs, and vanilla model NGOs.

Another example of inductive bias that we embed in the NGO formulation is the scale equivariance of the steady diffusion equation \ref{eq:darcy} with respect to its material parameter $\theta$, which we can express as
\begin{equation}\label{PDE scale equivariance}
    \mathcal{L}[c\theta] = c\mathcal{L}[\theta], \;\;\;\;\; \forall c \in \mathbb{R}^+,
\end{equation}
where $c$ is a positive constant. In Appendix \ref{A Scale equivariance tests}, we show how to embed this symmetry into the NGO formulation, and we demonstrate its effect on the NGO's generalization error across different material parameter scales.

\subsection{Results}\label{S Results}
In the next section, we compare the performance of the NGO in \eqref{E NGO HC2D} to that of a number of other models: VarMiONs, DeepONets, FNOs, CNOs, and U-Nets.

\subsubsection{Generalization Errors of Different NOs}\label{S Generalization Errors of Different NOs}
\begin{figure}[htb]
    \centering
    \begin{subfigure}{0.49\textwidth}
        \includegraphics[width=\textwidth]{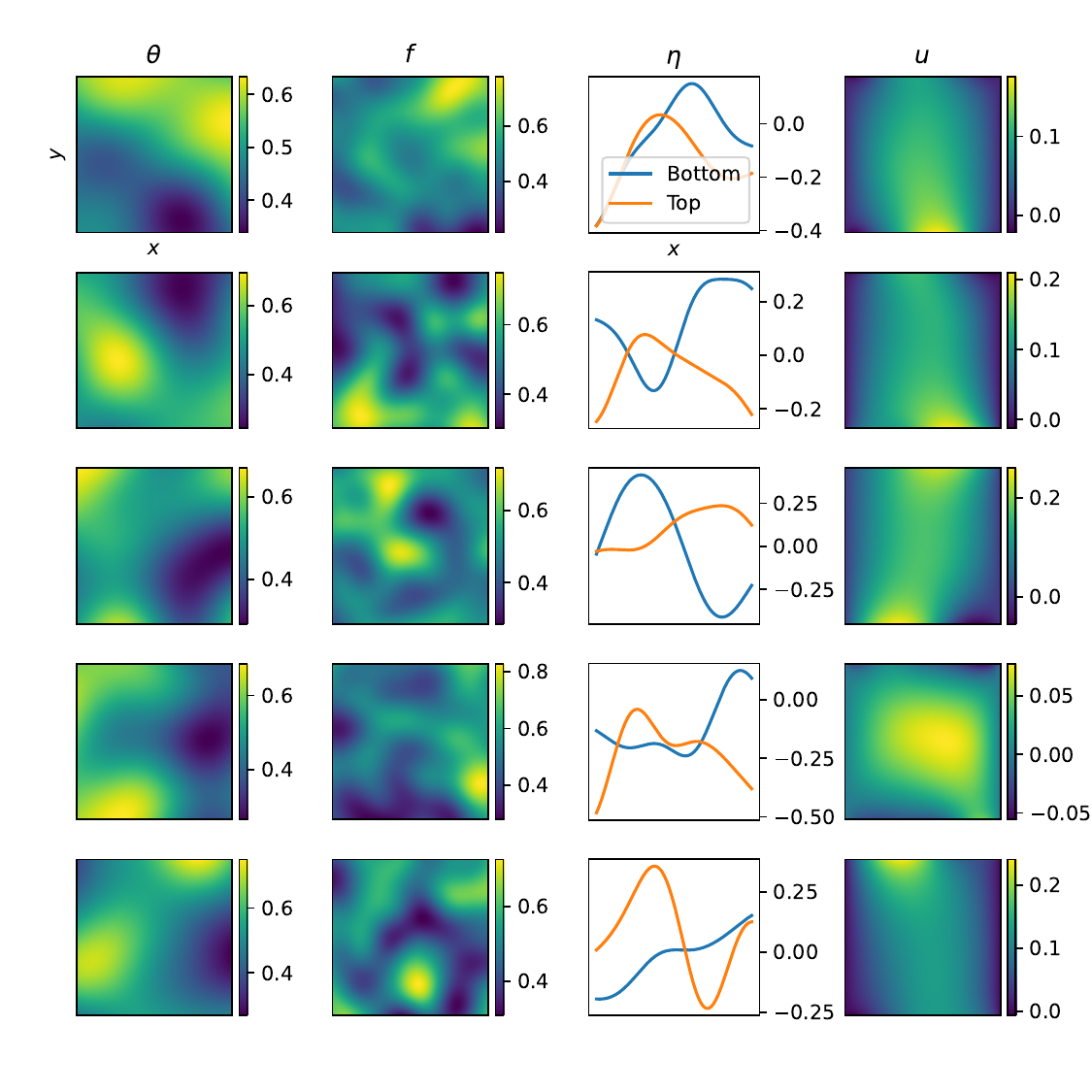}
        \caption{Finite element data set}
        \label{fig:example-solutions-nutils}
    \end{subfigure}
    \begin{subfigure}{0.49\textwidth}
        \includegraphics[width=\textwidth]{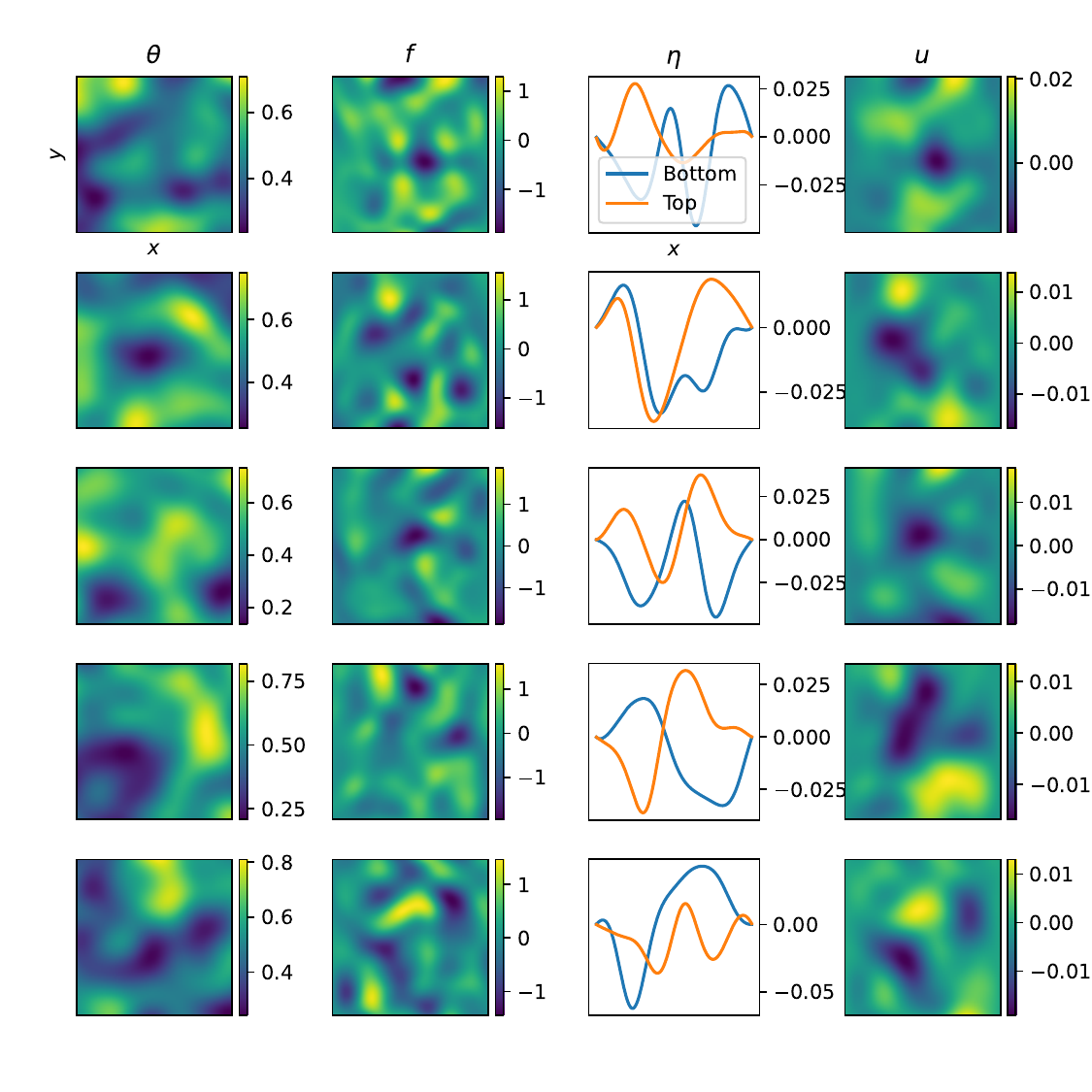}
        \caption{Manufactured data set}
        \label{fig:example-solutions-manufactured}
    \end{subfigure}
    \caption{Five random samples of solutions from the finite element data set (top) and the manufactured data set (bottom).}
\end{figure}

Table~\ref{tab:heat_eq_results} shows the error percentages achieved by the trained models on in-distribution and out-of-distribution testing data. Dataset A is used for training data and in-distribution testing data, while dataset B is used for out-of-distribution data (see \ref{sec:darcy-data-generation} for a description of the different data sets). The exact architectures of these models are given in Appendix~\ref{sec:appendix-model-architectures}. The details of the training procedure are given in Appendix~\ref{sec:appendix-training-procedure-darcy}. Figures~\ref{fig:example-solutions-nutils} and~\ref{fig:example-solutions-manufactured} show samples from both data sets.

The errors are given in terms of mean and standard deviation of the relative \(L^2\) error. We define the relative \(L^2\) error as \(e_i = \Vert \hat{u}^{(i)} - u^{(i)}\Vert_{L^2(\Omega)}/\Vert u^{(i)} \Vert_{L^2(\Omega)}\), where \(u^{(i)}, i = 1, \dots, N\) are the true solutions and \(\hat{u}^{(i)}\) are the model predictions.

As a reference, Table~\ref{tab:heat_eq_results} also includes the accuracy obtained by the Galerkin Finite Element Method (FEM) using the same \(12 \times 12\) cubic B-spline basis functions, as well as the error obtained by projecting the true solutions onto the span of the basis functions. The ``Projection'' entry is the lowest possible error achievable by any model using these basis functions. The final entry, ``POD-Galerkin'', refers to the Galerkin method using the first 144 POD modes of the training data as a function basis.

\begin{remark}
    We note that out of the models compared here, only the NGO and VarMiON architectures represent mappings that are linear in the source and boundary data of the PDE. In contrast, the DeepONet/RINO, FNO, CNO, and U-Net architectures can represent arbitrary (nonlinear) functions of \(\theta\), \(f\), \(\eta\), and \(g\). While this lack of structural constraint often results in higher out-of-distribution errors for these models, it also grants them the flexibility to represent a broader class of mappings. On the other hand, existing approaches that account for the linearity of the solution map in the source terms, such as~\cite{boullé2022data, teng2022learning}, focus on a single instance of the PDE coefficients rather than a parametric family, and therefore are not directly comparable to NGOs. Finally, although we employ a parametric linear PDE as a benchmark, such problems are standard for evaluating even fully nonlinear neural operators \cite{lu2021learning, li2020fourier, li2020multipole, raonic2024convolutional}. In Section~\ref{S nonlinear} we demonstrate that NGOs can also be effective for nonlinear PDEs via fixed-point iteration.
\end{remark}

\begin{table}
    \centering
    \caption{Accuracy of the tested models on different data sets. The `Galerkin' entry shows the solution errors of solving the PDEs with Galerkin finite elements on the B-spline basis used by the DeepONet, VarMiON, and NGOs. The `Projection' entry shows the error when projecting the true solution onto the B-spline basis, which is therefore the lowest error achievable by any model using this basis.}
    \begin{tabular}{l | r | r r}
        \toprule
        \multirow{2}*{Model} & \multirow{2}*{Parameters} & \multicolumn{2}{c}{Test error} \\
        & & In distribution & Out of distribution \\
        \midrule
        DeepONet              & 31224 & \(0.86\% \pm 0.73\%\) & \(\phantom{}80380.83\% \pm \phantom{}26549.77\%\) \\
        RINO                  & 31224 & \(0.96\% \pm 2.23\%\) & \(\phantom{}38483.79\% \pm \phantom{}22491.59\%\) \\
        VarMiON               & 31283 & \(0.27\% \pm 0.20\%\) & \(\phantom{000}20.53\% \pm \phantom{000}19.37\%\) \\
        FNO                   & 28975 & \(0.31\% \pm 0.16\%\) & \(\phantom{00}373.59\% \pm \phantom{00}149.42\%\) \\
        CNO                   & 28109 & \(1.41\% \pm 0.74\%\) & \(\phantom{00}774.59\% \pm \phantom{00}193.58\%\) \\
        U-Net                 & 30221 & \(0.82\% \pm 0.38\%\) & \(\phantom{00}340.80\% \pm \phantom{00}253.54\%\) \\
        \midrule
        Model NGO             & 28387 & \(0.24\% \pm 0.24\%\) & \(\phantom{000}12.94\% \pm \phantom{0000}3.51\%\) \\
        Preconditioned model NGO& 28387 & \(0.21\% \pm 0.17\%\) & \(\phantom{0000}6.68\% \pm \phantom{000}23.17\%\) \\
        Preconditioned data-free NGO & 28387 & \(2.82\% \pm 2.83\%\) & \(\phantom{0000}6.73\% \pm \phantom{0000}3.86\%\) \\
        Data NGO              & 27981 & \(0.26\% \pm 0.37\%\) & \(\phantom{000}12.07\% \pm \phantom{0000}3.73\%\) \\
        \midrule
        Galerkin              &       & \(0.10\% \pm 0.06\%\) & \(\phantom{0000}2.76\% \pm \phantom{0000}0.74\%\) \\
        Projection            &       & \(0.08\% \pm 0.05\%\) & \(\phantom{0000}2.63\% \pm \phantom{0000}0.70\%\) \\
        POD-Galerkin          &       & \(0.03\% \pm 0.02\%\) & \(\phantom{0000}4.38\% \pm \phantom{0000}1.46\%\) \\
        \bottomrule
    \end{tabular}
    \label{tab:heat_eq_results}
\end{table}

\begin{figure}[htb]
    \centering
    \begin{subfigure}{\linewidth}
        \centering
        \includegraphics[width=\textwidth]{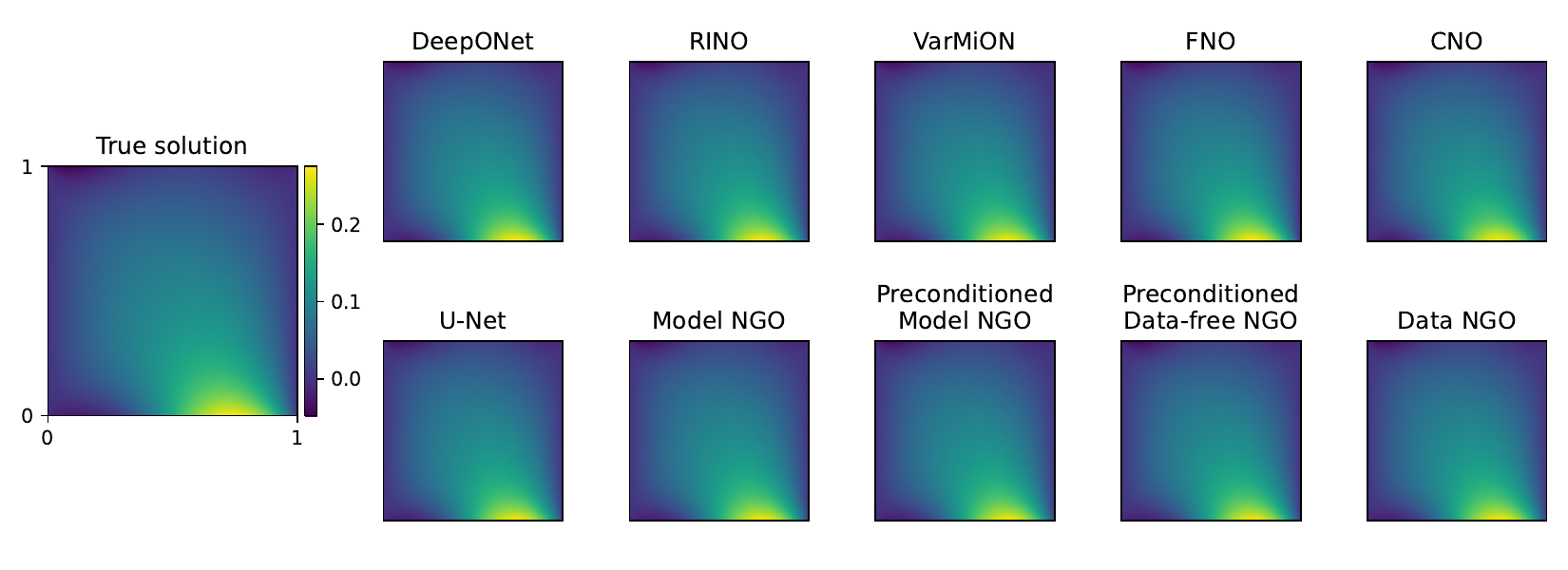}
        \caption{The true solution of one problem from the in-distribution data set, and model predictions of the different neural operators.}
        \label{subfig:diffusion-nutils-example-outputs}
    \end{subfigure}
    \begin{subfigure}{\linewidth}
        \centering
        \includegraphics[width=\textwidth]{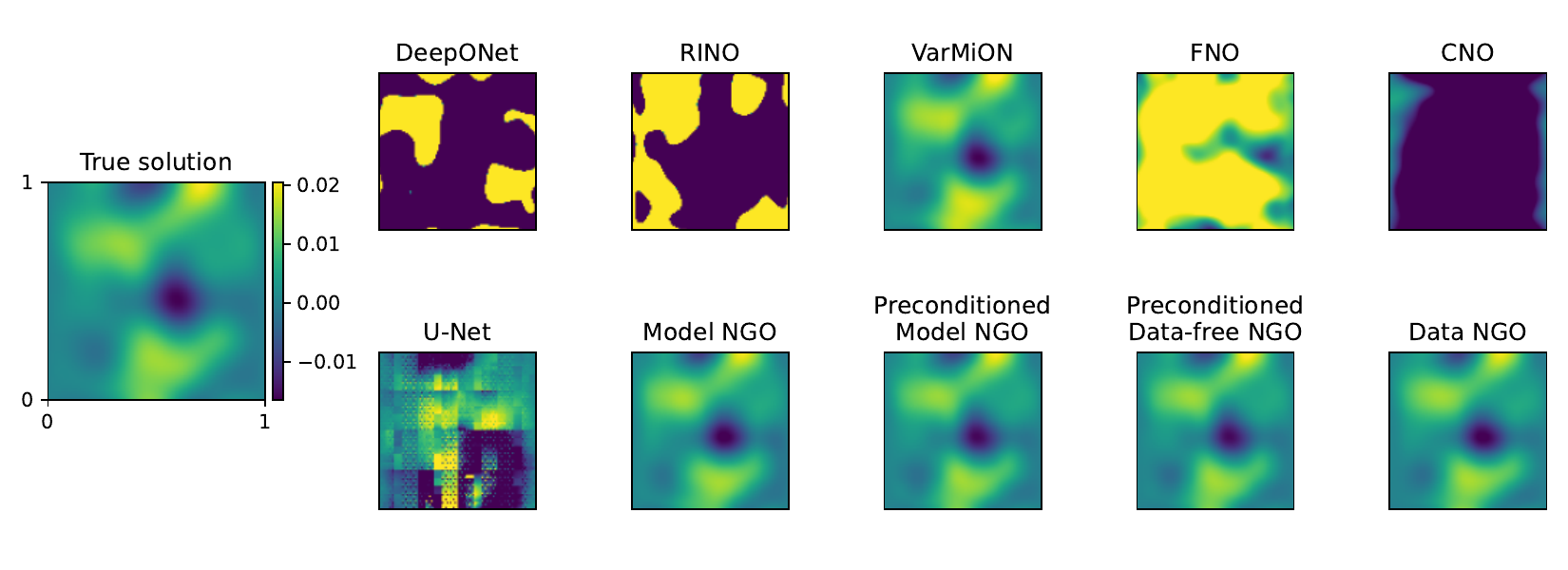}
        \caption{The true solution of one problem from the out-of-distribution data set, and model predictions of the different neural operators.}
        \label{subfig:diffusion-manufactured-example-outputs}
    \end{subfigure}
    \caption{The true solution and model predictions to one problem from the in-distribution dataset (Figure~\ref{subfig:diffusion-nutils-example-outputs}) and one problem from the out-of-distribution dataset (Figure~\ref{subfig:diffusion-manufactured-example-outputs}).}
    \label{fig:diffusion-example-outputs}
\end{figure}

Figure~\ref{fig:diffusion-example-outputs} shows the predictions made by the four NGO models and the five reference models, when tested on in-distribution data (Figure~\ref{subfig:diffusion-nutils-example-outputs}) and out-of-distribution data (Figure~\ref{subfig:diffusion-manufactured-example-outputs}). Note that all models perform well on the in-distribution data, producing approximate solutions that are visually indistinguishable from the true solution. However, with exception of the VarMiON, the reference models fail to generalize to out-of-distribution data, instead producing predictions with greater than \(100\%\) relative error. A noteworthy observation is that the U-Net, the only tested model that does not use a basis but instead acts on pointwise values only, fails to produce smooth functions as output.

\subsubsection{Generalization Errors Across Length Scales}\label{S Generalization Errors Across Length Scales}
In this section, we compare the NGOs against other models from literature in their ability to generalize across test data length scales. To elucidate the efficacy of the NGO construction that mimics the action of the Green's operator, depicted in Figure \ref{F NGO}, we compare the performance of NGOs against DeepONets and VarMiONs that use the same basis, NN architecture (for the learnable model component), number of trainable parameters and quadrature density. More specifically, we compare a DeepONet, VarMiON, data NGO, data-free NGO and model NGO that all use the same \(10 \times 10\) cubic B-spline basis, all use a U-Net with approximately $3\cdot10^4$ trainable model parameters, and all use a quadrature grid of approximately \(100 \times 100\) points. For completeness, we also report the performance of a standalone NN (U-Net) and an FNO with \(10 \times 10\) Fourier modes, with approximately the same numbers of trainable parameters and quadrature points. Details of the model architectures are highlighted in Appendix \ref{A Model definition details of Section}. The models have been trained on dataset C, described in Appendix \ref{sec:darcy-data-generation}, using the training procedure described in Appendix \ref{A training procedure steady diffusion Joost}.

\begin{figure}[htb!]
\centering
\includegraphics[width = 0.75\textwidth]{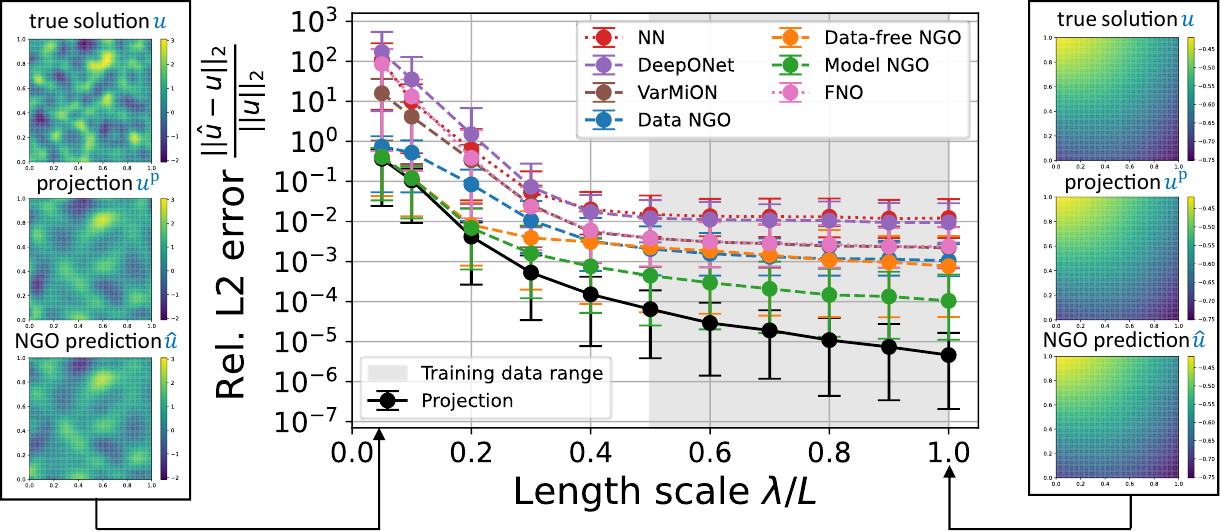}
\caption{\small{Relative \(L^2\) test error versus test data length scale $\lambda/L$ (where we used $L=1$) for a standalone NN (here a U-Net), and the same NN, when used in a DeepONet, VarMiON, data NGO, data-free NGO and model NGO.  All models have roughly \(3\cdot10^4\) trainable parameters, and all models (except the NN and FNO, also added as reference) use the same \(10 \times 10\) cubic B-spline basis, and are thus limited by the same \(L^2\) projection error lower bound, indicated in black. Points and error bars are, respectively, averages and 95\% confidence intervals on datasets of 1000 manufactured steady diffusion solutions, generated in the same way as dataset C.}}
\label{F neuraloperators}
\end{figure}

\begin{figure}[htb]
\centering
\includegraphics[width = 0.75\textwidth]{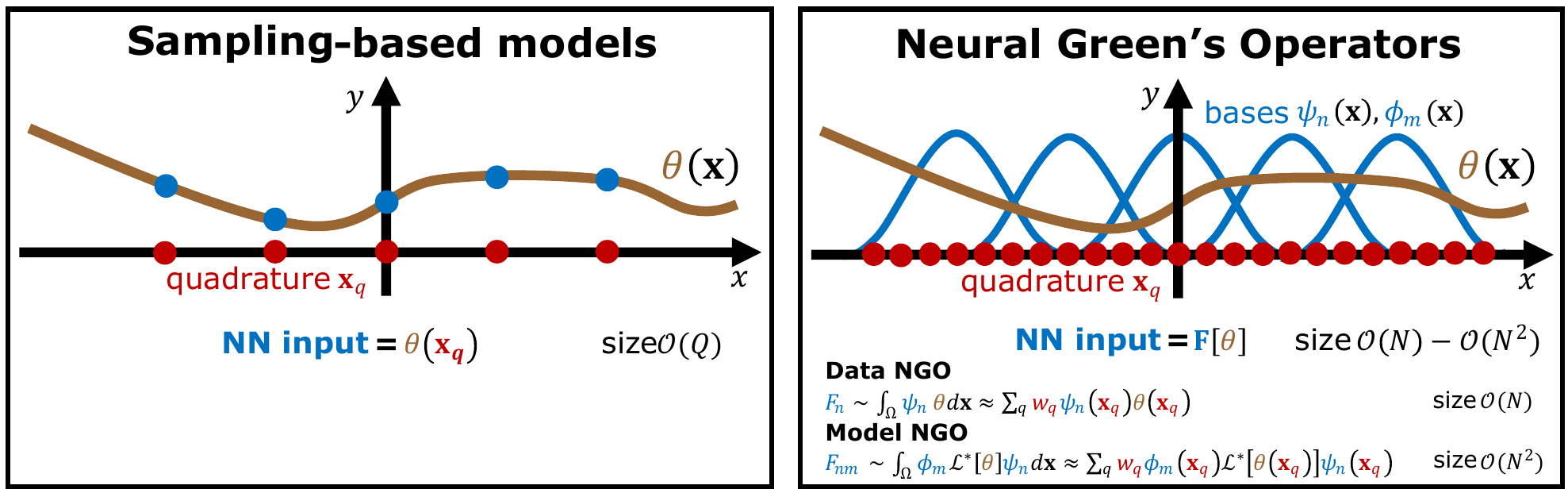}
\caption{\small{The input structure of an NGO compared to input structure of neural operators from literature like the DeepONet, FNO and VarMiON. Typically, the learnable components of canonical neural operators take as input the input function $\theta$ sampled on a quadrature grid $\vb{x}_q$, so the size of the input vector is determined by the number of quadrature points $Q$. On the other hand, the NGO system net takes as input inner products of the input function with a basis, which are computed using a quadrature rule with weights $w_q$ and points $\vb{x}_q$. The size of the NGO input is thereby determined by the size of the basis $N$, and independent of the number of used quadrature points $Q$. Therefore, for the same NN input size (in the example illustration, 5 numbers), an NGO can use (arbitrarily) finer quadrature, resulting in higher accuracy on problems that involve fine scale data, where dense quadrature is required.}}
\label{F NGOinput}
\end{figure}

The results are presented in Figure \ref{F neuraloperators}, where the accuracies of the different NOs are compared on test datasets with varying GRF length scales $\lambda$.
The models are tested on in-distribution length scales $0.5<\lambda/L<1$ (dataset C), as well as on out-of-distribution test data with finer length scales $0.05<\lambda/L<0.5$ (generated in the same way as dataset C). The black data in Figure \ref{F neuraloperators} are the \(L^2-\)projection errors, which result from projecting the true solutions onto the B-spline basis. The \(L^2-\)projection errors are a lower bound for the errors in the solutions of the NOs that use this B-spline basis (DeepONet, VarMiON, data NGO, data-free NGO and model NGO). We observe that while smooth solutions can be accurately represented on the B-spline basis, solutions with smaller length scales are underresolved, which is also visible in the example solution and its projection, shown in the left of Figure \ref{F neuraloperators}. The performance of the U-Net, a baseline neural network (NN) shown in red, is evaluated when used independently. This model achieves reasonable accuracy within the training data distribution; however, its error increases significantly outside this distribution. A similar trend is observed for the FNO and the DeepONet, with the FNO exhibiting a lower in-distribution error compared to the U-Net and DeepONet. VarMiONs, by contrast, preserve the linear dependence of the solution on $f$, $\eta$, and $g$, as well as the nonlinear dependence on $\theta$, as expressed in (\ref{E Green's operator Darcy}). The preservation of such structure explains their consistently higher accuracy compared with DeepONets and U-Nets. In contrast to VarMiONs, NGOs explicitly compute the right-hand side vector $d_n$ rather than learning it, significantly reducing the number of trainable parameters. For VarMiONs, the learnable matrices $\hat{F}_{nq}$, $\hat{H}_{nq}$, and $\hat{G}_{nq}$ scale with the quadrature density $Q$, leading to a parameter count of $\mathcal{O}(Q)$. With a fixed parameter budget of $3 \times 10^4$, VarMiONs are restricted to an input quadrature of $Q = 10 \times 10$, which may be insufficient to capture fine-scale features in the data. NGOs, however, take weighted averages of the material parameters $\theta$ against the basis functions as input. Consequently, their parameter count depends solely on the number of basis functions $N$ and is independent of the quadrature density $Q$. This independence enables NGOs to utilize arbitrarily fine quadrature at the input stage, leading to superior out-of-distribution accuracy on fine-scale data. The relationship between quadrature density and model input for DeepONets, VarMiONs, and NGOs is summarized in Figure~\ref{F NGOinput}.

\subsubsection{Comparison of Different Bases and System Network Architectures}\label{S Comparison of Different Bases and System Network Architectures}
\begin{figure}[htb]
\centering
\includegraphics[width = 0.75\textwidth]{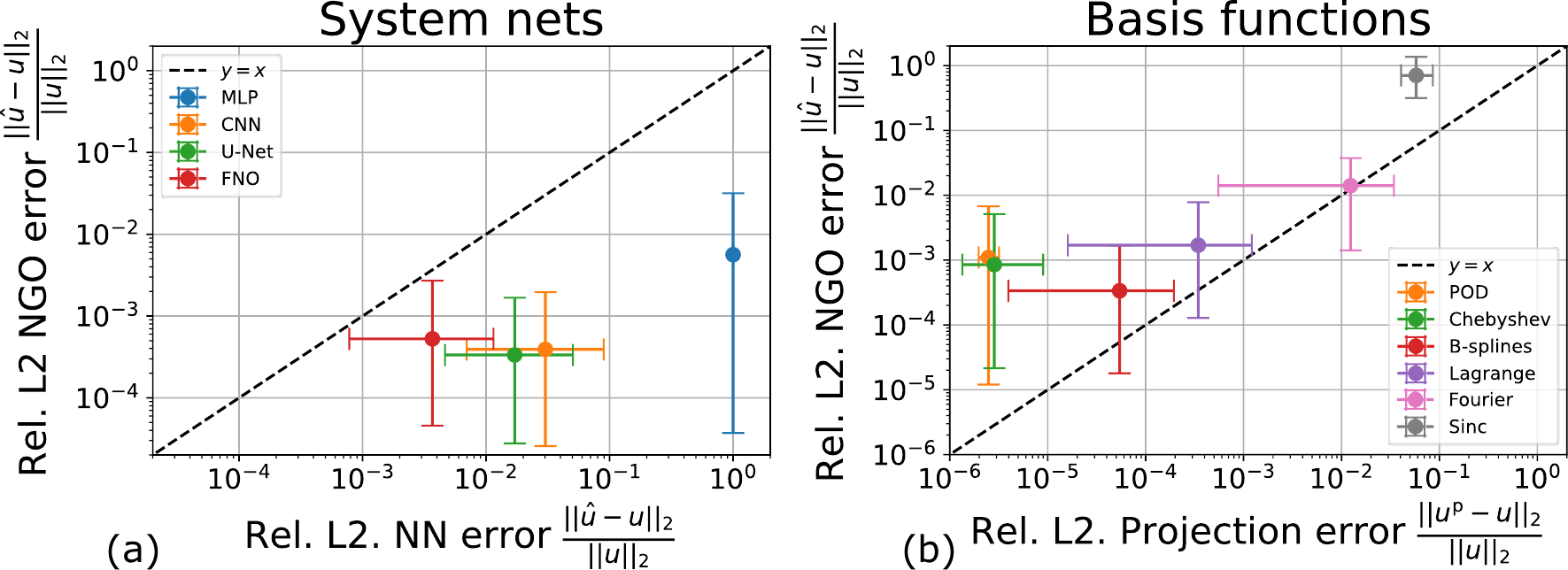}
\caption{\small{(a) Comparison of the error of a few canonical NN architectures when used as standalone solution approximators, versus the same NN when used as system net in a preconditioned model NGO with \(10 \times 10\) cubic B-spline bases. (b) Comparison of the \(L^2\) projection errors of solutions projected on a few canonical bases having 100 degrees of freedom, versus the error of a preconditioned model NGO with U-Net as a system network that uses the same bases. Points and error bars are, respectively, averages and 95\% confidence intervals on 1000 manufactured steady diffusion solutions, generated in the same way as dataset C.}}
\label{F systemnets}
\end{figure}

 In this section, we compare the approximation properties of NGOs that use different combinations of system net architectures and bases.  Details of the model architectures are highlighted in Appendix \ref{A Model definition details of Section}, and the system net architectural details can be found in Appendix \ref{A system net architectures Joost}. The models have been trained on dataset C, described in Appendix \ref{sec:darcy-data-generation}, using the training procedure described in Appendix \ref{A training procedure steady diffusion Joost}.
Figures \ref{F systemnets}(a) and (b) show the effect of using different system nets and bases to approximate solutions of (\ref{eq:darcy}) on in-distribution test datasets against the errors incurred by standalone architectures and \(L^2-\)projection errors onto the bases, respectively.
The results in Figure \ref{F systemnets}(a) convey that the use of NNs as system networks within an NGO consistently outperforms their standalone use as solution approximators. For example, the errors incurred by an MLP are reduced by more than two orders of magnitude when used within a preconditioned model NGO. While the FNO is the most accurate of the standalone network architectures, the U-Net performs the best when used as system network in our model NGO. The results in Figure \ref{F systemnets}(b) convey that while more efficient approximation spaces generally result in more accurate NGOs, we observe that the gap with respect to the \(L^2-\)projection errors widens for POD and Chebyshev bases. This widened gap may be due to the limited expressivity of the system network or the limitations of our training process. 

\subsubsection{Sensitivity Analysis}\label{S Sensitivity Analysis}
\begin{figure}[htb]
\centering
\includegraphics[width = 0.7\textwidth]{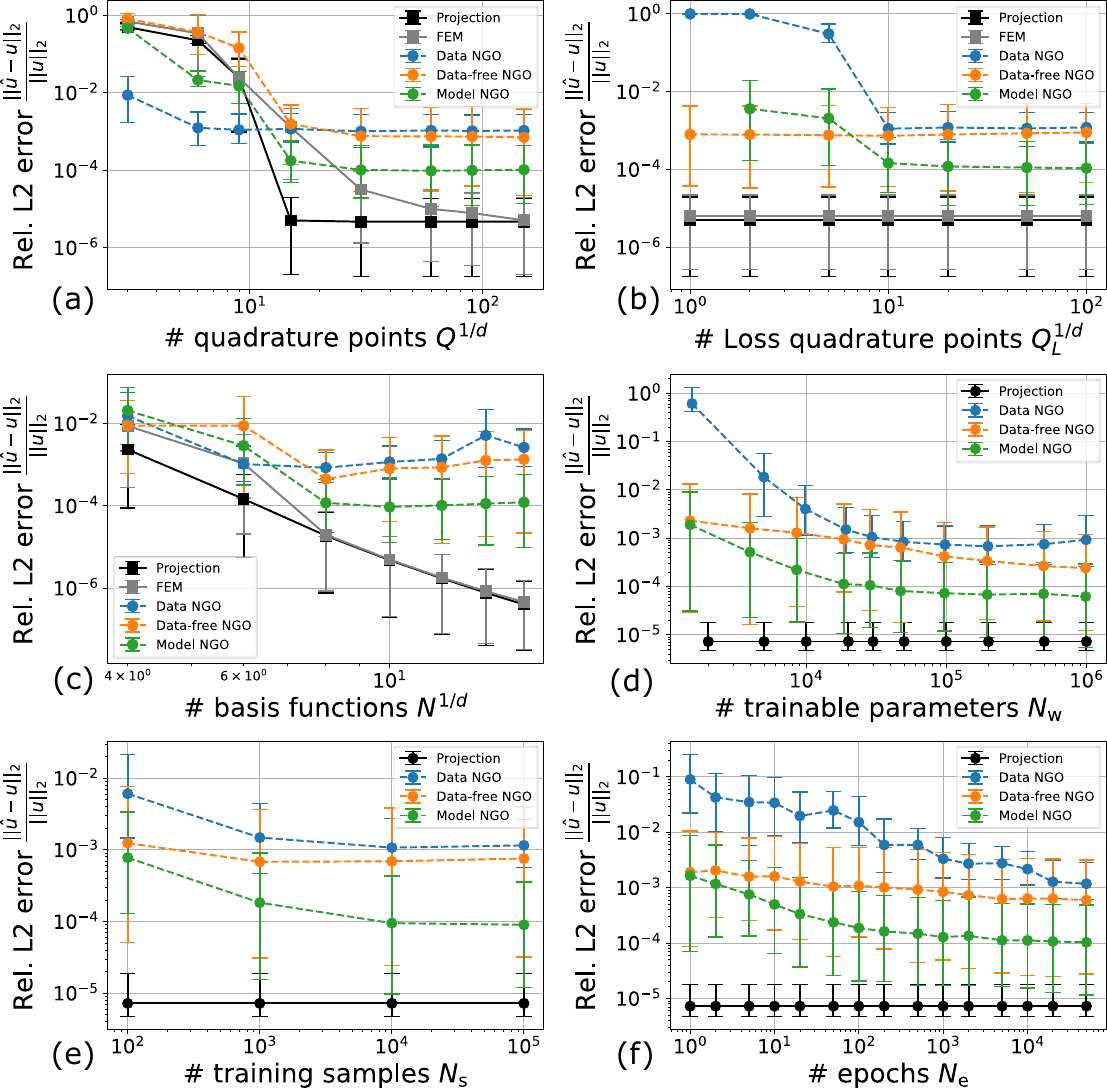}
\caption{\small{The relative \(L^2\) test error of a data NGO, Neumann-preconditioned data-free NGO, and a Neumann-preconditioned model NGO, in comparison to FEM and projection errors, versus the (a) number of assembly quadrature points per dimension $Q^{1/d}$, (b) number of loss quadrature points per dimension $Q_{\mathrm{L}}^{1/d}$, (c) number of basis functions $N$, (d) number of trainable parameters $N_w$, (e) number of training samples $N_\mathrm{s}$, and (f) number of training epochs $N_\mathrm{e}$. Points and error bars are, respectively, averages and 95\% confidence intervals on 1000 manufactured steady diffusion solutions, generated in the same way as dataset C.}}
\label{F erroranalysis}
\end{figure}

In this section, we numerically analyze the sensitivity of the NGO accuracy on important model and training parameters, namely, quadrature density, size of the basis, number of trainable parameters, size of the training dataset and the number of training epochs. Details of the model architectures are highlighted in Appendix \ref{A Model definition details of Section}. The models have been trained on dataset C, described in Appendix \ref{sec:darcy-data-generation}, using the training procedure described in Appendix \ref{A training procedure steady diffusion Joost}. Figure \ref{F erroranalysis} shows the resulting error curves of a data NGO, data-free NGO, model NGO, in comparison to FEM and $L^2$ projection. The main observation is that for all tests, the model NGO error stagnates at approximately \(10^{-4}\), so none of the parameters on the horizontal axes dominate the error. The discrepancy between this error and the projection error can be associated with the limited approximation capability of the used system net (here, a U-Net), or limitations of our training process. The design of a better system net and/or optimization strategy is left for future research. Another interesting observation from Figure \ref{F erroranalysis}(a) is that, as discussed in Section \ref{S Different types of NGOs}, the data-free NGO is bounded from below by the FEM error, which is not necessarily the case for data and model NGOs, and that data and model NGOs are less sensitive to assembly quadrature errors than FEM, and even projection. Final interesting observations are that for Neumann-preconditioned model NGOs, as few as \(10^4 - 10^5\) trainable parameters, \(10^4\) training samples, and \(5\cdot10^3\) training epochs are usually sufficient. A data NGO needs about \(2\cdot10^4\) epochs to converge.

\subsubsection{Training Speed}
\label{S Train Speed}
In addition to the accuracy of different models, we also compare the model architectures by their computational efficiency. To do this, the VarMiON, FNO, and NGO variants are benchmarked on a server with a 28-core 2.6GHz Intel Xeon Gold 6132 CPU and four NVIDIA Tesla V100 GPUs. The training performance is measured by evaluating the model on a batch of 100 samples, and also performing the backpropagation necessary for gradient-based optimization.
Note that NGOs first compute inner products of the PDE parameters with the basis functions or their derivatives. This step comes at a measurable computational cost, but does not depend on any trainable parameters of the models. As such, it is possible to precompute these vectors and matrices for a given data set, and train the NGO by passing precomputed arrays to it during training. The NGO performance is tested with this optimization implemented.

Table~\ref{tab:performance-comparison} shows the training and inference performance of FNOs, VarMiONs, as well as model NGOs and data NGOs. From this, it is clear that the DeepONet is the fastest model to train, followed by the VarMiON, the U-Net, and then the NGOs. The memory usage of the NGOs is higher than that of the DeepONet and VarMiON, but lower than that of the FNOs and comparable to the CNO.

\begin{table}[htb]
    \centering
    \caption{Performance comparison, in terms of time and memory usage, of NGOs compared to FNOs and VarMiONs. For the NGOs, the benchmark takes advantage of the ability to pre-assemble the system network inputs.}
    \begin{tabular}{l | r | r}
        \toprule
        Model & Computing time per batch [ms] & Peak memory usage [MB] \\
        \midrule
		Model NGO     &   9.2 &  498 \\
		Data free NGO &  10.7 &  565 \\
		Data NGO      &   5.5 &  345 \\
		DeepONet      &   1.1 &   32 \\
		VarMiON       &   3.7 &   81 \\
		FNO           & 106.7 & 4140 \\
		CNO           &  40.6 &  491 \\
		U-Net         &   5.2 &  201 \\
        \bottomrule
    \end{tabular}
    \label{tab:performance-comparison}
\end{table}

\subsubsection{NGO-based Preconditioners}
\label{S Precon}
Recently, neural operators have gained interest as methods to speed up the numerical solution of PDEs, using discretizations such as finite differences or finite elements. There are multiple ways to achieve this acceleration. One method, used by Zhang et al.~\cite{zhang2024blending}, Hu et al.~\cite{hu2025hybrid}, and Ackmann et al.~\cite{ackmann2020machine} among others, constructs a fixed-point iteration scheme for linear systems (derived from PDEs), that alternates between a ``standard'' smoothing step, such as Jacobi or Gauss-Seidel, and an application of the respective neural operator. Others, including Kopaničáková et al.~\cite{kopaničáková2025deeponet} and Xiang et al.~\cite{xiang2025unsupervised}, use the learned operator as a preconditioner directly, in combination with a linear solver such as Flexible GMRES~\cite{saad1993flexible} to allow nonlinear operators as preconditioners for linear systems. However, both approaches require specific choices for the solver or iteration scheme used, which limits their applicability to existing PDE solver software.

An advantage of NGOs is that the fact that they learn an approximation to the Green's function facilitates the construction of preconditioners for numerical discretizations of the same PDE.
Constructing preconditioners from Green's functions can be done qualitatively, as done by Loghin~\cite{loghin1999greens}, or numerically using learned Green's functions as done by Ichimura et al.~\cite{ichimura2020fast}.
In this section, we give an example of using a trained NGO to construct a preconditioner for a finite difference discretization for the diffusion equation. Specifically, a given discretization of a PDE yields a linear system with a matrix \(\mathbf{C}[\theta]\), for which we can construct a preconditioner \(\mathbf{M}[\theta]\) from the NGO, such that the condition number of \(\mathbf{P} = \mathbf{C}\mathbf{M}^{-1}\) is much smaller than that of \(\mathbf{C}\).

For this example, we take the steady diffusion problems from the testing data set, formulate a finite difference discretization of the PDE, and use the data NGO trained in Section~\ref{S Test problem: steady diffusion} to construct a preconditioner. For the finite difference discretization, we assume a uniform grid (i.e.~\(\Delta x = \Delta y \equiv h\)) and use a five-point difference scheme to approximate the differential operator:
\begin{subequations}
    \label{eq:preconditioning-diffusion-discretisation}
    \begin{align*}
        -\nabla\cdot\left(\theta\nabla u\right) &= -\partial_x\left(\theta\partial_xu\right) - \partial_y\left(\theta\partial_yu\right), \\
        \left[\partial_x (\theta \partial_xu)\right]_{i,j} &\approx \frac{1}{\Delta x}\left(\frac{\theta_{i,j} + \theta_{i+1,j}}{2}\frac{u_{i+1,j} - u_{i,j}}{\Delta x} - \frac{\theta_{i-1,j} + \theta_{i,j}}{2}\frac{u_{i,j} - u_{i-1,j}}{\Delta x}\right), \\
        \left[\partial_y (\theta \partial_yu)\right]_{i,j} &\approx \frac{1}{\Delta y}\left(\frac{\theta_{i,j} + \theta_{i,j+1}}{2}\frac{u_{i,j+1} - u_{i,j}}{\Delta y} - \frac{\theta_{i,j-1} + \theta_{i,j}}{2}\frac{u_{i,j} - u_{i,j-1}}{\Delta y}\right),
    \end{align*}
\end{subequations}
where \(\theta_{i,j} = \theta(x_i, y_j)\) and \(u_{i,j}\) is the finite-difference approximation to \(u(x_i, x_j)\). We take \(h = \frac{1}{99}\), which produces a discretization with \(100\) points in the \(y\)-direction and \(98\) points in the \(x\)-direction (the Dirichlet boundary conditions imply that no boundary points need to be included). This produces an \(9800 \times 9800\) linear system \(\mathbf{C}\vb{u} = \vb{b}\).

To precondition this linear system, we use the direct preconditioning (DP) approach described in Kopaničáková et al.~\cite{kopaničáková2025deeponet}: the linear system is preconditioned using a multiplicative combination of a smoother and a coarse-level preconditioner, with the latter represented by a neural network. In this work, block-Jacobi is used as a smoother, and the coarse-level preconditioner is given by either a DeepONet or an NGO. To apply a DeepONet or an NGO as a preconditioner for such an \(9800 \times 9800\) linear system \(\mathbf{Cu} = \mathbf{b}\), the model must be able to take as input any vector \(\vb{v} \in \mathbb{R}^{9800}\) and return an approximation for \(\mathbf{C}^{-1}\vb{v} \in \mathbb{R}^{9800}\). Since \(\mathbf{C}^{-1}\vb{v}\) corresponds to the solution of a diffusion equation where the right-hand-side terms correspond to the vector \(\vb{v}\), this can be done by converting the vector \(\vb{v}\) to a representation taken by the neural operator, and then applying said operator to obtain an approximation for the solution to the diffusion equation on the desired grid points.

\paragraph*{Preconditioning with NGOs}
To use an NGO as a coarse-level preconditioner, we note that the action of a Green's operator can be approximated on the points used by the discretization:
\begin{align}
    \label{eq:preconditioning-quadrature-approximation}
    \hat{u}(\vb{x}_i)
    &\approx \int_\Omega \hat{G}[\theta](\vb{x}_i, \vb{x}')\,f(\vb{x}')\,d\vb{x}'
    \approx \sum_{j=1}^{n} \hat{G}[\theta](\vb{x}_i, \vb{x}_j)\,f(\vb{x}_j)\,w_j,
\end{align}
where \(\vb{x}_j\) are the points used by the finite-difference discretization, and \(w_j\) are associated quadrature weights. NGOs approximate the Green's function as in \eqref{E Galerkin approximation Green's function}:
\begin{align}
    \label{eq:preconditioning-greens-function-approximation}
    \hat{G}[\theta](\vb{x}_i, \vb{x}_j) &\approx \sum_{m,n} \phi_m(\vb{x}_i)\mathbf{A}_{mn}[\theta]\phi_n(\vb{x}_j).
\end{align}
Since the right-hand-side vector \(\vb{v}\) contains pointwise evaluations of the source term, i.e.~\(\vb{v}_j = f(\vb{x}_j)\), we can write the discretized action of the NGO as
\begin{align} \label{eq:precon}
    \vb{C}^{-1}\vb{v} \approx \mathbf{P}_{\text{NGO}}(\vb{v}) := \vb{PA[\theta]R}\vb{v},
    \text{ where } \vb{P}_{ij} = \phi_j(\vb{x}_i),~
    \vb{R}_{ij} = \phi_i(\vb{x}_j)w_j.
\end{align}
Therefore, the action of the NGO as a preconditioner as represented by the sequence of operations on \(\vb{v}\) in (\ref{eq:precon}) may be conceived of as a coarse-level preconditioner in the following manner:
\begin{enumerate}
    \item Computing the inner product of \(\vb{v}\) with the NGO basis functions, denoted by the operator \(\vb{R}_{ij}\), represents a restriction operator.
    \item The matrix \(\vb{A}[\theta]\) computed by the NGO represents a coarse-scale solution operator.
    \item Expanding the resulting coefficients in terms of the basis \(\phi_m\), denoted by the operator \(\vb{P}_{ij}\), represents a prolongation operator.
\end{enumerate}
In this way, the application of preconditioners inferred from NGOs can be applied algebraically to any Krylov subspace method.  Furthermore, the multiplicative combination with a block-Jacobi preconditioner works as follows: after the application of the NGO-based preconditioner, the block-Jacobi preconditioner is applied as a correction:
\begin{subequations}
    \begin{align}
        \tilde{\vb{x}} &= \mathbf{P}_{\text{NGO}}(\vb{v}) \\
        \tilde{\vb{r}} &= \vb{v} - \mathbf{C}\tilde{\vb{x}} \\
        \mathbf{C}^{-1}\vb{v} &\approx \mathbf{P}_{\text{NGO + blk.Jac.}}(\vb{v}) := \tilde{\vb{x}} + \mathbf{P}_{\text{blk.Jac.}}(\tilde{\vb{r}}).
    \end{align}
    \label{eq:multiplicative-preconditioning}
\end{subequations}

\paragraph*{Preconditioning with DeepONets}
To use a DeepONet as a preconditioner, we follow the same procedure as described in Kopaničáková et al.~\cite{kopaničáková2025deeponet}, although our setup is slightly different since we apply the preconditioner to a finite-difference discretization, rather than a finite-element discretization. This means that in order to compute the right-hand side vector for the DeepONet, which consists of the source terms evaluated on the sensor nodes of the DeepONet, \(f\) is interpolated from the finite-difference discretization points onto the sensor nodes using nearest-neighbor interpolation (i.e.~the inputs to the DeepONet are taken to be the pointwise values on the discretization points \(\vb{x}_i\) closest to each of the \(12 \times 12\) sensor nodes of the DeepONet). While this step technically introduces an error in the interpolation, we believe this error to be negligible overall as the mesh size of \(h = \frac{1}{99}\) means that the difference between the DeepONet's sensor nodes and the actually sampled locations is small compared to the length scales of the input.

There are two more important details in the application of DeepONets as preconditioners:
\begin{itemize}
    \item The first is that while DeepONets are passed the individual source terms separately, in a discretization these terms are all combined into one right-hand-side vector. Since most of the entries in the right-hand-side vector follow from the volumetric source term \(f\) and not the boundary source term \(\eta\) (note that the Dirichlet boundary data \(g\) is always zero in the data set used here), the choice is made here to always set \(\eta \equiv 0\) and treat the right-hand-side vector \(\vb{v}\) as if it is entirely the result of the volumetric source term \(f\). While this may also introduce an error, there is no clear way to ``separate'' a right-hand side vector into contributions from the volumetric and boundary source terms and this limitation is not addressed in the original work.
    \item Secondly, since DeepONets are nonlinear functions, they may produce unexpected outputs when the right-hand-side vector \(\vb{v}\) is of a different order of magnitude than what is present in the training data. This is avoided by scaling the input vector so that its norm equals the average \(L^2\)-norm of the vectors \(\vb{f}\) in the training data, written here as \(r_0\). This enforces that the DeepONet-based preconditioner, while not linear, is positively homogeneous.
\end{itemize}
Algorithmically, applying the DeepONet-based preconditioner to a vector \(\vb{v} \in \mathbb{R}^{9800}\) can be expressed in pseudocode as follows:
\begin{subequations}
    \begin{align}
        \vb{f} &= \text{Interpolate}(\vb{v})
        & \text{nearest-neighbor interpolation} \\
        r &= \frac{\Vert \vb{f}\Vert}{r_0}
        & \text{normalization} \\
        \vb{c} &= \text{BranchNet}(\boldsymbol\theta, \tfrac{1}{r}\vb{f}, \boldsymbol\eta \equiv 0)
        & \text{apply branch net} \\
        \hat{\vb{u}}_i &= \sum_m \vb{c}_m\phi_m(\vb{x}_i)
        & \text{evaluate solution on grid points} \\
        \mathbf{C}^{-1}\vb{v} &\approx \mathbf{P}_{\text{DeepONet}}(\vb{v}) := r\,\hat{\vb{u}}.
        & \text{undo normalization}
    \end{align}
\end{subequations}
The multiplicative combination with the block-Jacobi preconditioner is done in the same way as for NGOs, i.e.~as in~\eqref{eq:multiplicative-preconditioning}.

\paragraph*{Results}
To test the efficacy of NGOs and DeepONets as preconditioners, we consider 1000 realizations of the diffusion problem. We solve the discretized equations with F-GMRES~\cite{saad1993flexible} restarted every 50 inner iterations, as well as unrestarted F-GMRES (written here as F-GMRES(50) and F-GMRES(\(\infty\)), respectively). However, F-GMRES(50) was found to diverge when used with a DeepONet preconditioner, and therefore the DeepONet preconditioning results are only available for unrestarted F-GMRES. Since the NGO preconditioner can also be applied to other linear solvers, the equations were also solved using the regular GMRES algorithm~\cite{saad1986gmres} (both restarted and unrestarted), as well as Bi-CGSTAB~\cite{vandervorst2024bicgstab}. The average number of iterations needed for convergence is shown in Table~\ref{tab:preconditioning-results}. The linear systems are solved with a relative tolerance of \(10^{-8}\). Right preconditioning is used in all cases.

\begin{table}
    \centering
    \caption{The average number of iterations that various iterative linear solvers needed to solve discretized diffusion equations. These averages are computed over 1000 realizations of diffusion problems.}
    \label{tab:preconditioning-results}
    \begin{tabular}{lrrrr}
        \toprule
        \multirow{2}*{Solver} & \multicolumn{4}{c}{Preconditioner} \\
        & None & Blk.~Jac. & Blk.~Jac.~+ NGO & Blk.~Jac.~+ DeepONet \\
        \midrule
        F-GMRES(\(\infty\)) &  482.9 & 293.7 & 50.0 & 274.9 \\
        F-GMRES(50)         & 1616.2 & 719.4 & 50.0 &  Fail \\
        GMRES(\(\infty\))   &  482.9 & 293.7 & 50.0 &     - \\
        GMRES(50)           & 1616.2 & 719.4 & 50.0 &     - \\
        Bi-CGSTAB           &  360.1 & 213.2 & 29.4 &     - \\
        \bottomrule
    \end{tabular}
\end{table}

\begin{figure}
    \centering
    \includegraphics[width=0.48\linewidth]{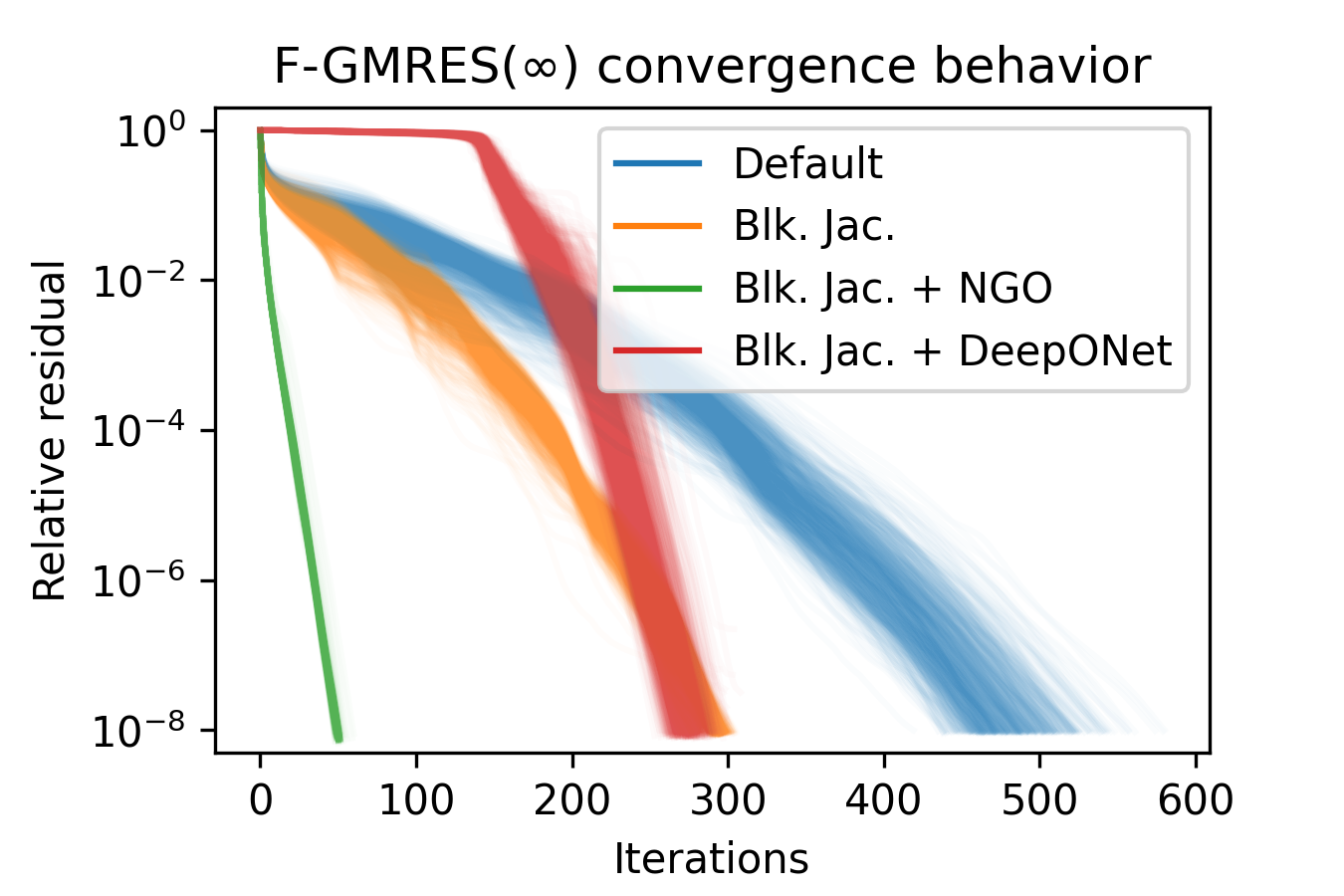}
    \hfill
    \includegraphics[width=0.48\linewidth]{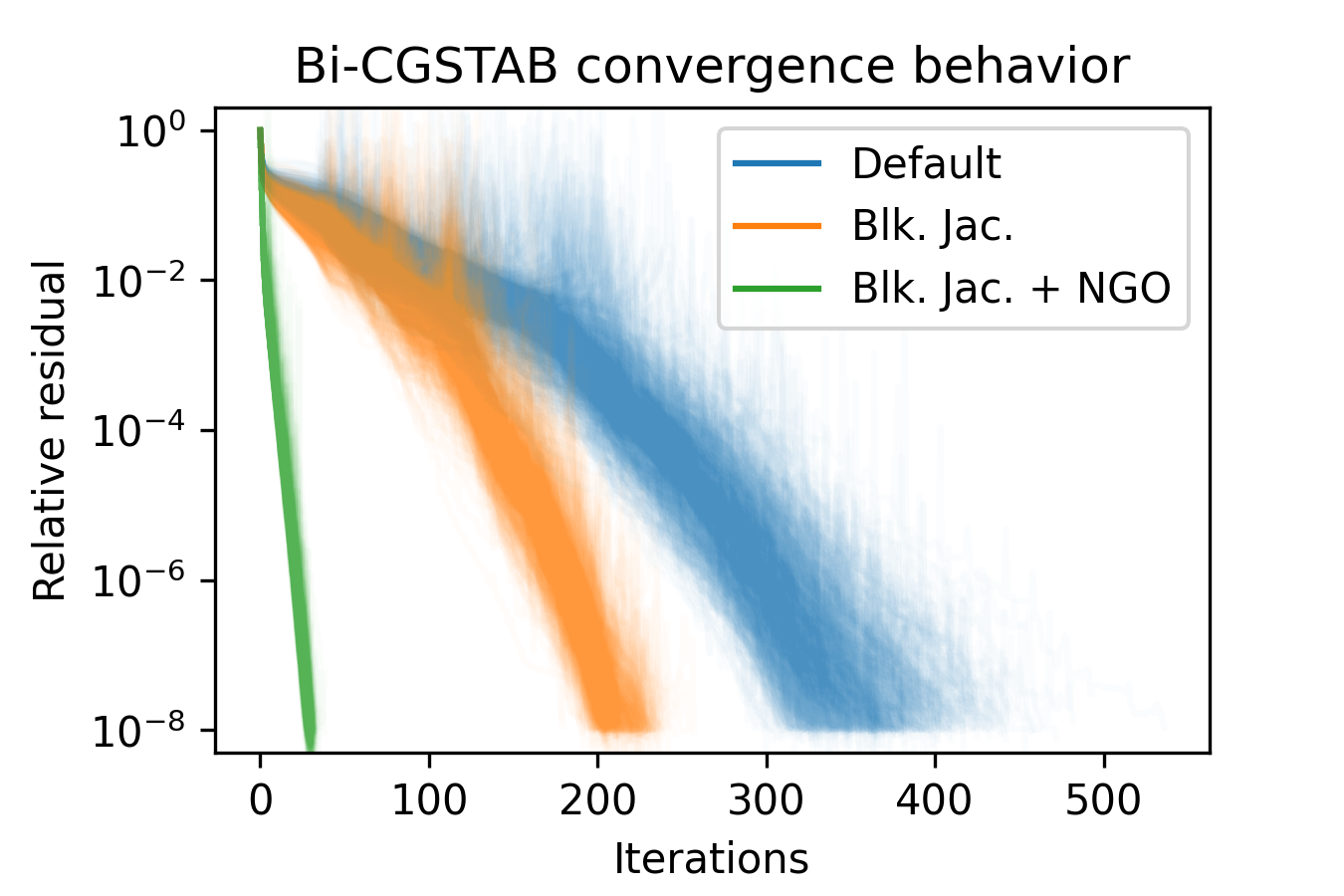}
    \caption{Convergence behavior of F-GMRES(\(\infty\)) and Bi-CGSTAB with various preconditioners.}
    \label{fig:preconditioning-convergence-plots}
\end{figure}

The convergence behavior of different solvers with various preconditioners is shown in Figure~\ref{fig:preconditioning-convergence-plots}. This figure shows that the NGO-based and DeepONet-based preconditioners accelerate F-GMRES in different ways: while the NGO-based preconditioner results in consistently fast convergence, the DeepONet causes F-GMRES to converge slowly at first but converge more quickly in later iterations.

Note that in Table~\ref{tab:preconditioning-results}, some combinations of solvers and preconditioners require the exact same number of iterations on average. This is due to two reasons:
\begin{itemize}
    \item Firstly, F-GMRES is equivalent to regular GMRES when used with a linear preconditioner, or with no preconditioner at all. Therefore, the convergence behavior of F-GMRES(\(\infty\)) and F-GMRES(50) are equivalent to that of GMRES(\(\infty\)) and GMRES(50), respectively (with the exception of the DeepONet-based preconditioner). 
    \item Secondly, with the block-Jacobi + NGO preconditioner, F-GMRES and GMRES very consistently converge in approximately 50 iterations, meaning that the restarted variants converge without ever restarting and therefore behave identically to their unrestarted counterparts. In particular, the ``50.0'' entries in Table~\ref{tab:preconditioning-results} are not erroneous and are confirmed by Figure~\ref{fig:preconditioning-convergence-plots} which shows unrestarted F-GMRES converge in 50 iterations with the NGO-based preconditioner. 
\end{itemize}

An important observation is that F-GMRES(\(\infty\)) with the block-Jacobi + DeepONet preconditioner needs almost as many iterations on average as F-GMRES(\(\infty\)) with only a block-Jacobi preconditioner. In other words, the inclusion of the DeepONet is not very productive. We suspect that this is due to the worse generalizability of DeepONets compared to NGOs: as the F-GMRES algorithm iterates, it may apply the preconditioner to vectors that don't correspond directly to right-hand side terms in the training data, i.e.~that lie out of the training distribution. As shown in Table~\ref{tab:heat_eq_results}, DeepONets are found not to generalize well beyond the training data set, and this may explain their poor performance as preconditioners in this test case. NGOs, on the other hand, were found to generalize relatively well, and indeed they are also highly effective preconditioners.

We must note that if generalization is indeed the cause of this, then the poor preconditioner performance of DeepONets could be avoided by training on a richer data set, so that the preconditioner performance relies less on the generalizability of the model. As such, the results found here do not contradict those of Kopaničáková et al.~\cite{kopaničáková2025deeponet}, who found DeepONets to make for effective preconditioners. Nevertheless, our results show that NGOs \emph{can} learn effective preconditioners from the same training data. Furthermore, their linearity allows NGOs to precondition linear solvers other than F-GMRES. This is relevant as Table~\ref{tab:preconditioning-results} shows Bi-CGSTAB to be a more efficient algorithm for the diffusion problem considered here, and Bi-CGSTAB cannot be used with a nonlinear preconditioner such as one derived from a DeepONet.

\begin{remark}
    In their work, Kopaničáková et al.~\cite{kopaničáková2025deeponet} also consider a second type of machine learning-based preconditioner, called the trunk-basis (TB) approach. There, the learned basis of the DeepONet is used to construct the restriction and prolongation operators \(\vb{R}\) and \(\vb{P}\), after which the preconditioner is constructed without using the branch network of the DeepONet. In this work, the DeepONet and NGO models both utilize the same basis, thereby making the trunk-basis approach trivially equivalent between both models. As such, the trunk-basis approach is omitted from the comparison.
\end{remark}
\begin{remark}
    Our construction of NGO-based preconditioners can be viewed as a coarse-level preconditioner within a classical two-level framework, in line with the approach of Nikolopoulos et al.~\cite{nikolopoulos2024ai} and, in the context of model order reduction, with the work of Pasetto et al.~\cite{pasetto2017reduced}.
\end{remark}

\FloatBarrier
\section{Extension to Other Problems}
\label{S Extend}
\subsection{Time-Dependent Diffusion}\label{S time dependent diffusion}
\begin{figure}[htb]
\centering
\includegraphics[width = 0.3\textwidth]{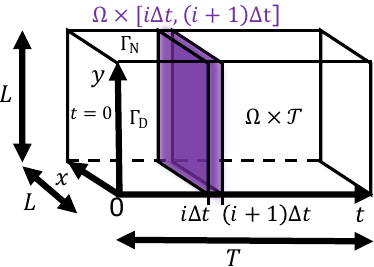}
\caption{\small{The space-time domain $\Omega\times\mathcal{T}$, with Neumann boundaries $\Gamma_{\mathrm{N}}$ on the top and bottom, Dirichlet boundaries $\Gamma_{\mathrm{D}}$ on the sides, and the initial condition boundary $t=0$ on the left side. We used $L=T=1$.}}
\label{F domain_dynamic_diffusion}
\end{figure}

In this section we consider the time-dependent diffusion problem:
\begin{equation}\label{E PDE time-dependent diffusion}
    \begin{aligned}
        \left(\pdv{}{t}-\nabla\cdot \theta\nabla \right) u &= f, &&\forall(\vb{x},t)\in \Omega\times \mathcal{T}, \\
        \vu{n} \cdot \theta \nabla u &= \eta,  &&\forall(\vb{x},t)\in \Gamma_\mathrm{N} \times \mathcal{T}, \\
        \theta u &= g,  &&\forall(\vb{x},t)\in \Gamma_\mathrm{D}\times \mathcal{T}, \\
        u &= u_0,
        &&\forall(\vb{x},t)\in \Omega\times \{0\},
    \end{aligned}
\end{equation}
where $\Omega=[0,1]^2$ and $\mathcal{T}=[0,T]$, which is a direct time-dependent generalization of the steady diffusion problem as treated in Section \ref{S Test problem: steady diffusion}. The differences with respect to the static problem are that the operator in (\ref{E PDE time-dependent diffusion}) contains a time derivative $\partial/\partial t$, the input functions $\theta$, $f$, $\eta$ and $g$ are functions of space and time $(\vb{x},t)$, and there is an initial condition $u=u_0$. Instead of resolving the Green's operator corresponding to \eqref{E PDE time-dependent diffusion} on the full time domain $\mathcal{T}$, we restrict ourselves to a time slab of length $\Delta t$, as shown in Figure \ref{F domain_dynamic_diffusion}.
The solution to problem \eqref{E PDE time-dependent diffusion} can be expressed in terms of a Green's operator as
\begin{equation}
\begin{aligned}
    \label{E G tdd}
        u^{(i)}(\vb{x},t) &= \int_\Omega G^{(i)}[\theta](i\Delta t) u^{(i-1)}(i\Delta t) d\vb{x}' +\int_{\Omega\times \Delta t_i} G^{(i)}[\theta] f d\vb{x}'dt'  
        + \int_{\Gamma_\mathrm{N}\times \Delta t_i} G^{(i)}[\theta] \eta d\vb{x}'dt' \\
        &-\int_{\Gamma_\mathrm{D}\times \Delta t_i} g \vu{n} \cdot \nabla G^{(i)}[\theta] d\vb{x}'dt',
\end{aligned}
\end{equation}
where $G^{(i)}$ is the Green's function corresponding to time slab $i$, and $u^{(0)}(0)=u_0$. The derivation of \eqref{E G tdd} is provided in Appendix \ref{A G tdd}. By now stepping forward in time $u^{(i)} \to u^{(i+1)}$ from $t=0$ to $t=T$, we resolve a piecewise continuous approximation of the true solution $u$ over the full time domain $\mathcal{T}$.

\subsubsection{NGO Architecture}
Substitution of \eqref{E Galerkin approximation Green's function}, where the basis functions $\psi^{(i)}_n$ and $\phi^{(i)}_m$ corresponding to time slab $i$ are now functions of space and time $(\vb{x},t)$, results in the NGO
\begin{equation}\label{E GGO HC2D 2}
    \hat{u}^{(i)}(\vb{x},t) = \hat{\mathcal{G}}^{(i)}[u^{(i-1)},\theta,f,\eta,g](\vb{x},t) = \hat{A}^{(i)}_{mn}(\vb{F}[\theta]) d^{(i)}_n[u^{(i-1)},f,\eta,g] \phi^{(i)}_m(\vb{x},t),
\end{equation}
where
\begin{equation}\label{E tdd rhs}
\begin{aligned}
    d^{(i)}_n[u^{(i-1)},f,\eta,g] &= \int_\Omega \psi^{(i)}_n(i\Delta t) u^{(i-1)}(i\Delta t) d\vb{x}' + \int_{\Omega\times \Delta t_i} \psi^{(i)}_n f d\vb{x}'dt' 
    + \int_{\Gamma_\mathrm{N}\times \Delta t_i} \psi^{(i)}_n \eta d\vb{x}'dt' 
    \\
    &- \int_{\Gamma_\mathrm{D}\times \Delta t_i} g \vu{n} \cdot \nabla \psi^{(i)}_n d\vb{x}'dt'.
\end{aligned}
\end{equation}
For model NGOs and data-free NGOs, the input $\vb{F}[\theta]$ is, as described in Section \ref{S Different types of NGOs}, the (unstabilized) system matrix
\begin{equation}\label{E F model tdd}
\begin{aligned}
    F_{nm}[\theta] &= \int_{\Omega\times \Delta t_i} \phi_m^{(i)} \mathcal{L}^*[\theta] \psi_n d\vb{x}'dt'
    - \int_{\Gamma\setminus \Gamma_{\mathrm{N}}\times \Delta t_i}\theta \psi_n \vu{n}\cdot \nabla \phi_m^{(i)} d\vb{x}'dt' 
    + \int_{\Gamma\setminus \Gamma_{\mathrm{D}}\times \Delta t_i} \theta \phi_m^{(i)} \vu{n}\cdot \nabla \psi_n d\vb{x}'dt' \\
    &+ \int_{\Omega}\psi_n((i+1)\Delta t)\phi_m^{(i)}((i+1)\Delta t) d\vb{x}',
\end{aligned}
\end{equation}
which is equivalent to Equation \ref{E bilinear form tdd}. For data NGOs, $\vb{F}[\theta]$ is the material parameter $\theta$ integrated against the test basis as
\begin{equation}
    F_n[\theta] = \int_{\Omega\times \Delta t_i} \psi_n \theta d\vb{x}'dt'.
\end{equation}
To data NGOs, we also provide the additional input $F_{t,n}\equiv \int_\Omega \psi_n\left( (i+1)\Delta t \right) d\vb{x}'$, for the reason explained in Section \ref{A Model architectures time-dependent diffusion}. We use basis functions $\boldsymbol{\psi}(\vb{x},t)\equiv\{ \psi_n(\vb{x},t)\}_{n=1}^{N}$ and $\boldsymbol{\phi}(\vb{x},t)\equiv\{ \phi_m(\vb{x},t)\}_{m=1}^{N}$, which are Kronecker outer products of 1D basis functions such as
\begin{equation}\label{E Kronecker product basis}
    \boldsymbol{\phi}(\vb{x},t) = \boldsymbol{\phi}^{(t)}(t) \otimes_{\mathrm{K}} \boldsymbol{\phi}^{(x)}(x) \otimes_{\mathrm{K}}  \boldsymbol{\phi}^{(y)}(y),
\end{equation}
where $\boldsymbol{\phi}^{x_i}(x_i)\equiv\{\phi_l^{x_i} (x_i) \}_{l=1}^{n_i}$, $n_i$ is the number of basis functions in dimension $i$, and $\otimes_{\mathrm{K}}$ is the Kronecker product. In Appendix \ref{A Kronecker product assembly of the system matrix}, we show how to exploit the Kronecker product structure of the basis to enable more time and memory efficient assembly of the system matrix for model NGOs and data-free NGOs.
If we now substitute
\begin{equation}
    u^{(i-1)}(i\Delta t) = u^{(i-1)}_m \phi^{(i-1)}_m(i\Delta t),
\end{equation}
into \eqref{E tdd rhs},  we end up with the time stepping scheme
\begin{equation}\label{E time stepping scheme}
    \hat{\vb{u}}^{(i)} = \hat{\vb{A}}^{(i)} \vb{d}^{(i)}= \hat{\vb{A}}^{(i)} \vb{M}_{\mathrm{lr}} \hat{\vb{u}}^{(i-1)} + \hat{\vb{A}}^{(i)} \vb{d}_{\vb{x}}^{(i)},
\end{equation}
where we define
\begin{equation}\label{E M_lr}
    M_{\mathrm{lr},nm} \equiv \int_\Omega \psi^{(i)}_n(i\Delta t) \phi^{(i-1)}_m(i\Delta t)d\vb{x}' = \int_\Omega \psi^{(i)}_n(i\Delta t) \phi^{(i)}_m((i+1)\Delta t)d\vb{x}',
\end{equation}
as the mass matrix corresponding to the basis functions evaluated at the left and right ends of the time step element, and
\begin{equation}
    d^{(i)}_{\vb{x},n}[f,\eta,g] \equiv d^{(i)}_n[u^{(i-1)},f,\eta,g] - \int_\Omega \psi^{(i)}_n(\Delta t_i) u^{(i-1)}(\Delta t_i) d\vb{x}'
\end{equation}
as the right-hand-side vector corresponding to the forcing and boundary condition terms. To summarize, the architectures of the NGOs that are used for the steady diffusion problem (treated in Section \ref{S The Neural Green's Operator}) can be used to predict a single timestep forward in time for the time-dependent diffusion problem, if one accounts for the additional dependence of the basis functions on time $t$, and the additional input $u^{i-1}$ (and a few other details, which are discussed in Appendix \ref{A Model architectures time-dependent diffusion}). The resulting model can then be used to march forward in time using Equation \ref{E time stepping scheme}. In practice, we define the basis $\boldsymbol{\phi}^{(t)}(t)$ only on a single element in time and shift this basis forward in time upon doing time steps, which is equivalent to a discontinuous basis in time. In the remainder of this section, we investigate methods to preserve key properties of solutions to Equation~(\ref{E PDE time-dependent diffusion}) that enable an NGO trained on a single time-step to predict solutions autoregressively over an arbitrary number of time steps.

\subsubsection{Inductive Bias: Time Stepping Stability}\label{S Inductive Bias: Time Stepping Stability}
To study the stability over time of the solution of the time-dependent diffusion problem, we assume a Galerkin discretization with $\boldsymbol{\psi}(\vb{x},t) = \boldsymbol{\phi}(\vb{x},t)$ and substitute the solution $u$ as test function $v$ into the weak form (Equation \ref{E unstabilized weak form}), which results in the energy equation
\begin{equation}\label{E continuous energy equation}
\begin{aligned}
    \Delta E^{(i)} &\equiv \frac{1}{2} \int_{\Omega} \left(u^{(i)} \right)^2((i+1)\Delta t)d\vb{x}' - \frac{1}{2} \int_{\Omega} \left(u^{(i-1)} \right)^2(i\Delta t)d\vb{x}' \\
    &= -D^{(i)}[\theta][u^{(i)},u^{(i)}] 
    + \int_{\Omega\times \Delta t_i} u^{(i)} f d\vb{x}'dt' 
    + \int_{\Gamma_{\mathrm{N}}\times \Delta t_i} u^{(i)} \eta d\vb{x}'dt' \\
    &+ \int_{\Gamma_{\mathrm{D}}\times \Delta t_i} g \left( C_{\mathrm{s}} - \vu{n}\cdot \nabla \right)u^{(i)} d\vb{x}'dt'
\end{aligned}
\end{equation}
where
\begin{equation}\label{E diffusive dissipation}
\begin{aligned}
    D^{(i)}[\theta][u^{(i)},u^{(i)}] &= \int_{\Omega\times \Delta t_i}\theta \norm{\nabla u^{(i)}}_2^2 d\vb{x}'dt'
    + \int_{\Gamma_{\mathrm{D}}\times \Delta t_i} \theta u^{(i)} \left(C_{\mathrm{s}}  - 2 \vu{n}\cdot \nabla \right) u^{(i)}  d\vb{x}'dt' \\
    &\geq \alpha^{(i)}[\theta] \norm{u^{(i)}}^2
\end{aligned}
\end{equation}
is the diffusive energy/entropy dissipation, the second right-hand-side term is the energy production, and the third and fourth terms are energy outflow on the Neumann and Dirichlet boundaries, respectively. $C_{\mathrm{s}} $ is a Nitsche stabilization constant \cite{Bazilevs_Hughes_2005}, which is required to ensure non-negativity of $D$. Substituting $u^{(i)}=u^{(i)}_m \phi^{(i)}_m$ in (\ref{E diffusive dissipation}) results in the corresponding discrete energy dissipation relation
\begin{equation}\label{E discrete energy equation}
    \frac{1}{2} \norm{\vb{u}^{(i)}}_{\vb{M}_{\mathrm{rr}}}^2 - \frac{1}{2} \norm{\vb{u}^{(i-1)}}_{\vb{M}_{\mathrm{rr}}}^2 
    \leq \left(\vb{u}^{(i)}\right)^{T} \vb{h}^{(i)}_x,
\end{equation}
where $\vb{u}^{(i)} = \{u_m^{(i)}\}_{m=1}^{N}$,
\begin{equation}
    \vb{h}_x^{(i)} =  \int_{\Omega\times \Delta t_i} \boldsymbol{\phi}^{(i)} f d\vb{x}'dt' 
    + \int_{\Gamma_{\mathrm{N}}\times \Delta t_i} \boldsymbol{\phi}^{(i)} \eta d\vb{x}'dt' 
    + \int_{\Gamma_{\mathrm{D}}\times \Delta t_i} g \left( C_{\mathrm{s}}  - \vu{n}\cdot \nabla \right)\boldsymbol{\phi}^{(i)} d\vb{x}'dt'
\end{equation}
and $\norm{.}_{\vb{M}_{\mathrm{rr}}}$ is the vector norm induced by the mass matrix
\begin{equation}\label{E M_rr}
    M_{\mathrm{rr},nm} \equiv M^{(i)}_{\mathrm{rr},nm} = \int_\Omega \phi^{(i)}_n((i+1)\Delta t) \phi^{(i)}_m((i+1)\Delta t)d\vb{x}'
\end{equation}
corresponding to the basis functions evaluated at the end of the time step. The discrete energy equation \eqref{E discrete energy equation} describes the evolution of the $L^2$ norm (in space) of the discrete solution $u^{(i)}=u^{(i)}_m \phi^{(i)}_m$ over time.

\paragraph{Operator Stability}
To derive a stability criterion for the output of the system network from \eqref{E discrete energy equation}, we consider, without loss of generality, the homogeneous counterpart of \eqref{E PDE time-dependent diffusion} where $f=\eta=g=0$. In the homogeneous setting, the discrete energy inequality \eqref{E discrete energy equation} simplifies to
\begin{equation}\label{E discrete energy equation homogeneous}
    \frac{1}{2} \norm{\vb{u}^{(i)}}_{\vb{M}_{\mathrm{rr}}}^2 - \frac{1}{2} \norm{\vb{u}^{(i-1)}}_{\vb{M}_{\mathrm{rr}}}^2 \leq 0.
\end{equation}
Substitution of the time stepping scheme \eqref{E time stepping scheme} (with $\vb{d}^{(i)}_{\vb{x}}=0$) into Equation (\ref{E discrete energy equation homogeneous}) (with $\vb{h}^{(i)}_{\vb{x}}=0$) and reordering gives
\begin{equation}\label{E u i-1 inequality}
    \frac{\norm{\hat{\vb{A}}^{(i)}[\theta] \vb{M}_{\mathrm{lr}}\vb{u}^{(i-1)}}_{\vb{M}_{\mathrm{rr}}}}{\norm{\vb{u}^{(i-1)}}_{\vb{M}_{\mathrm{rr}}}} \leq 1.
\end{equation}
Since \eqref{E u i-1 inequality} is true for all $\vb{u}^{(i-1)}$,
it implies the stability criterion
\begin{equation}\label{E stability criterion}
    \sup_{\vb{u}^{(i-1)}} \frac{\norm{\hat{\vb{A}}^{(i)}[\theta] \vb{M}_{\mathrm{lr}}\vb{u}^{(i-1)}}_{\vb{M}_{\mathrm{rr}}}}{\norm{\vb{u}^{(i-1)}}_{\vb{M}_{\mathrm{rr}}}} \equiv \norm{\hat{\vb{A}}^{(i)}[\theta] \vb{M}_{\mathrm{lr}}}_{\vb{M}_{\mathrm{rr}}} \leq 1
\end{equation}
To guarantee stability of the NGOs, we therefore need to ensure that NGOs produce matrices $\hat{\vb{A}}^{(i)}[\theta]$ that satisfy the criterion \eqref{E stability criterion}. Note that \eqref{E stability criterion} is a property of the matrix $\hat{\vb{A}}^{(i)}[\theta]$ that is independent of $f$, $\eta$ and $g$, so \eqref{E stability criterion} is also true for the inhomogeneous problem. 

To ensure that  output of the system network in model and data-free NGOs are dissipative in the sense of \eqref{E stability criterion}, we rely on a preconditioned system network, as described in Section \ref{S Preconditioning the System Network}, where $\vb{F}_0$ incorporates stabilization using Nitsche's method \cite{Bazilevs_Hughes_2005} guaranteeing that $\vb{F}_0^{-1}$ satisfies (\ref{E stability criterion}). For data NGOs, however, we assume the PDE information to be unknown and, thus, the system matrix to be unavailable. To ensure that a data NGO satisfies \eqref{E stability criterion} we propose a matrix norm scaling layer:
\begin{equation}\label{E spectral radius scaling}
\hat{\vb{A}}'^{(i)} = 
    \begin{cases}
        \begin{aligned}
            &\hat{\vb{A}}^{(i)} &\norm{\hat{\vb{A}}^{(i)}\vb{M}_{\mathrm{lr}}}_{\vb{M}_{\mathrm{rr}}} \leq S, \\
            &\frac{S}{\norm{\hat{\vb{A}}^{(i)}\vb{M}_{\mathrm{lr}}}_{\vb{M}_{\mathrm{rr}}}} \hat{\vb{A}}^{(i)}
            &\norm{\hat{\vb{A}}^{(i)}\vb{M}_{\mathrm{lr}}}_{\vb{M}_{\mathrm{rr}}} > S,
        \end{aligned}
    \end{cases}
\end{equation}
which guarantees that
\begin{equation}
    \norm{\hat{\vb{A}}^{(i)}\vb{M}_{\mathrm{lr}}}_{\vb{M}_{\mathrm{rr}}} \leq S \leq 1.
\end{equation}
Here, $S \leq 1$ is a hyperparameter that determines the maximum norm that the data NGO is allowed to predict. That is, if the output of the system network $\hat{\vb{A}}^{(i)}$ is such that it violates \eqref{E stability criterion}, we scale it back, such that the resulting scaled matrix $\hat{\vb{A}}'^{(i)}$ does satisfy \eqref{E stability criterion}. For computational reasons, we do not explicitly calculate $\|\hat{\vb{A}}^{(i)}\vb{M}_{\mathrm{lr}}\|_{\vb{M}_{\mathrm{rr}}}$, but we estimate it up to relative error tolerance $\epsilon=10^{-3}$ using a power iteration algorithm \cite{golub2013matrix}, of which implementation details can be found in Appendix \ref{A Training Procedure Time-Dependent Diffusion}.

\begin{figure}[htb]
\centering
\includegraphics[width = 0.75
\textwidth]{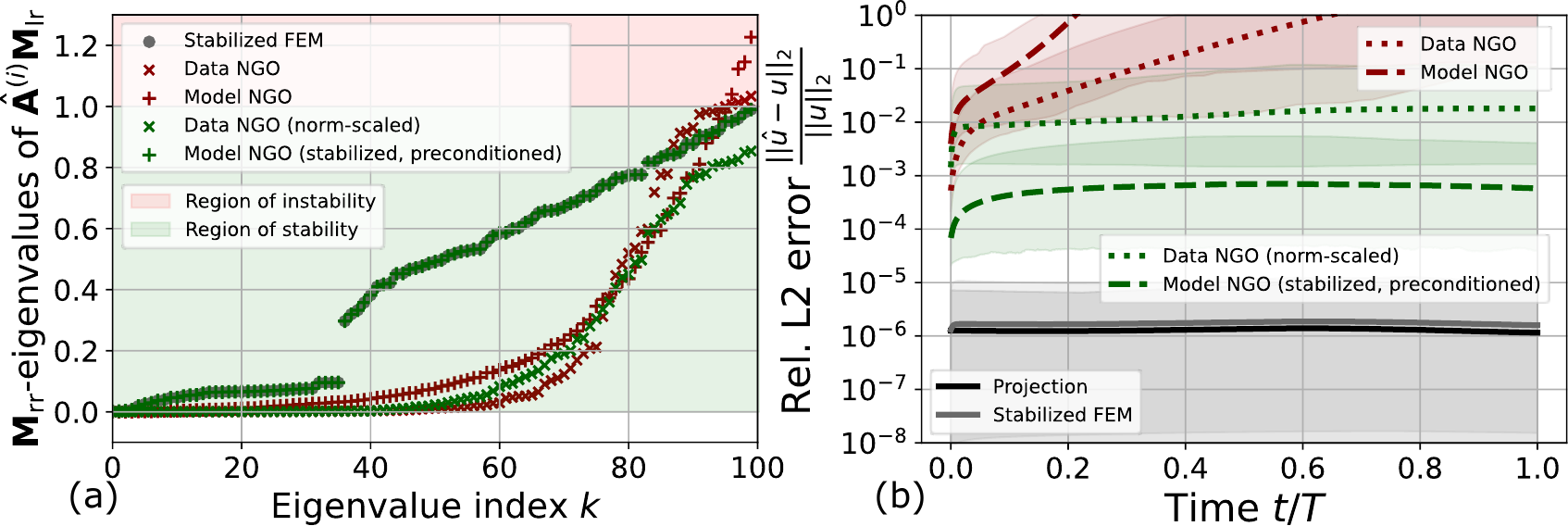}
\caption{\small{Effect of adding the Nitsche-stabilized Neumann series preconditioning to a model NGO, and the norm scaling layer \eqref{E spectral radius scaling} to a data NGO, on their (a) mass matrix $\vb{M}_{\mathrm{rr}}$-generalized eigenvalue spectrum and (b) relative time step \(L^2\) test error over time. 
Projection and stabilized FEM are included as reference in black and gray, respectively. In (a), points correspond to the spectrum of a single matrix corresponding to a manufactured $\theta$ (manufactured as in dataset D). In (b), lines and shaded areas are, respectively, 95\% confidence intervals on 100 manufactured solutions generated in the same way as dataset D, but with $\lambda/L=\tau/T=1$ (where we used $L=T=1$)}.}\label{F spectralradiusscaling}
\end{figure}

In Figure \ref{F spectralradiusscaling}, we show the effect of the preconditioning strategy and the norm scaling layer on the mass-matrix-generalized eigenvalue spectrum and long term stability of a model NGO and data NGO, respectively. Details of the model architectures are highlighted in Appendix \ref{A Model architectures time-dependent diffusion}. The models have been trained on dataset D, described in Appendix \ref{A data generation time-dependent diffusion}, using the training procedure described in Appendix \ref{A Training Procedure Time-Dependent Diffusion}. In Figure \ref{F spectralradiusscaling}(a), we show that the preconditioning strategy for the model NGO and the norm scaling layer \eqref{E spectral radius scaling} for the data NGO ensure that both models satisfy \eqref{E stability criterion}. The $\vb{M}_{\mathrm{rr}}$-generalized eigenvalue spectrum of the model NGO also visually overlaps with the stabilized FEM spectrum. In Figure \ref{F spectralradiusscaling}(b), we show that the included inductive biases ensure that both models have time integration errors that remain stable over time for 1000 time steps of size $\Delta t/T = 10^{-3}$, even though both NGOs have been trained on a single time step of size $\Delta t=10^{-3}$ only.
We observed that choosing a conservative $S < 1$ (we chose 0.8) results in lower error accumulation.

\paragraph{Conservation and Dissipation Laws}
In the remainder of this section we explore the effect of controlling the errors in the laws of mass conservation (derived by setting $v=1$ in \eqref{E unstabilized weak form} and $u^{(i)}=u^{(i)}_m \phi^{(i)}_m$),
\begin{equation}\label{E massconservation}
\begin{aligned}
    \Delta m^{(i)} &\equiv 
    u^{(i)}_m  \int_{\Omega} \phi^{(i)}_m((i+1)\Delta t) d\vb{x}' 
    - u^{(i-1)}_m \cdot \int_{\Omega}\phi^{(i-1)}_m(i\Delta t) d\vb{x}' \\
    &= \int_{\Omega\times \Delta t_i} f d\vb{x}'dt' + \int_{\Gamma_{\mathrm{N}}\times \Delta t_i} \eta d\vb{x}'dt' 
    + 
    u^{(i)}_m  \int_{\Gamma\setminus \Gamma_{\mathrm{N}}\times \Delta t_i}\theta \vu{n}\cdot \nabla \phi^{(i)}_m d\vb{x}'dt',
\end{aligned}
\end{equation}
and energy dissipation (\ref{E discrete energy equation}) on the accuracy of the predictions of NGOs of solutions (\ref{E PDE time-dependent diffusion}) over time. 
Collecting \eqref{E massconservation} and \eqref{E discrete energy equation} we obtain a system of equations for the vector of unknowns $\vb{u}^{(i)}$ given by
\begin{equation}\label{E discrete conservation equations}
\begin{cases}
\begin{aligned}
    \left(\vb{u}^{(i)}\right)^T\vb{c}^{(i)} &= C^{(i)} &\mathrm{(mass),} \\
    \frac{1}{2} \norm{\vb{u}^{(i)}}_{\vb{M}_{\mathrm{rr}}}^2 - \frac{1}{2} \norm{\vb{u}^{(i-1)}}_{\vb{M}_{\mathrm{rr}}}^2 &\leq \left(\vb{u}^{(i)}\right)^{T} \vb{h}^{(i)}_x
        &\mathrm{(energy),}
\end{aligned}
\end{cases}
\end{equation}
where we defined 
\begin{equation}
    c^{(i)}_m[\theta] = \int_{\Omega} \phi_m^{(i)}((i+1)\Delta t)d\vb{x}' - \int_{\Gamma \setminus \Gamma_{\mathrm{N}}} \theta \vu{n}\cdot \nabla \phi_m^{(i)} d\vb{x}'dt'
\end{equation}
and 
\begin{equation}
    C^{(i)}[u^{(i-1)},f,\eta] = \int_{\Omega}u_m^{(i-1)} \phi_m^{(i-1)} (i\Delta t) d\vb{x}' + \int_{\Omega\times \Delta t_i} f d\vb{x}'dt' + \int_{\Gamma_{\mathrm{N}}\times \Delta t_i} \eta d\vb{x}'dt'. 
\end{equation}
To control the error in the output $\hat{\vb{u}}^{(i)}$ of an NGO with respect to the relations in \eqref{E discrete conservation equations} we apply the following correction
\begin{equation}\label{E conservation ansatz}
    \hat{\vb{u}}'^{(i)} = a\hat{\vb{u}}^{(i)} + b \vb{c}^{(i)},
\end{equation}
where $a$ and $b$ are scalars. Substitution of  \eqref{E conservation ansatz} into \eqref{E discrete conservation equations} results in a quadratic system of equations that can be solved for $a$ and $b$. Algorithmically, we start with $a=1$ and correct only the mass conservation equation by solving for $b$. We then check for violation of the energy inequality, and if violated, we do another correction on top of the first correction, in which we solve for $a<1$ and $b$ simultaneously. 

\begin{figure}[htb]
\centering
\includegraphics[width = 0.75\textwidth]{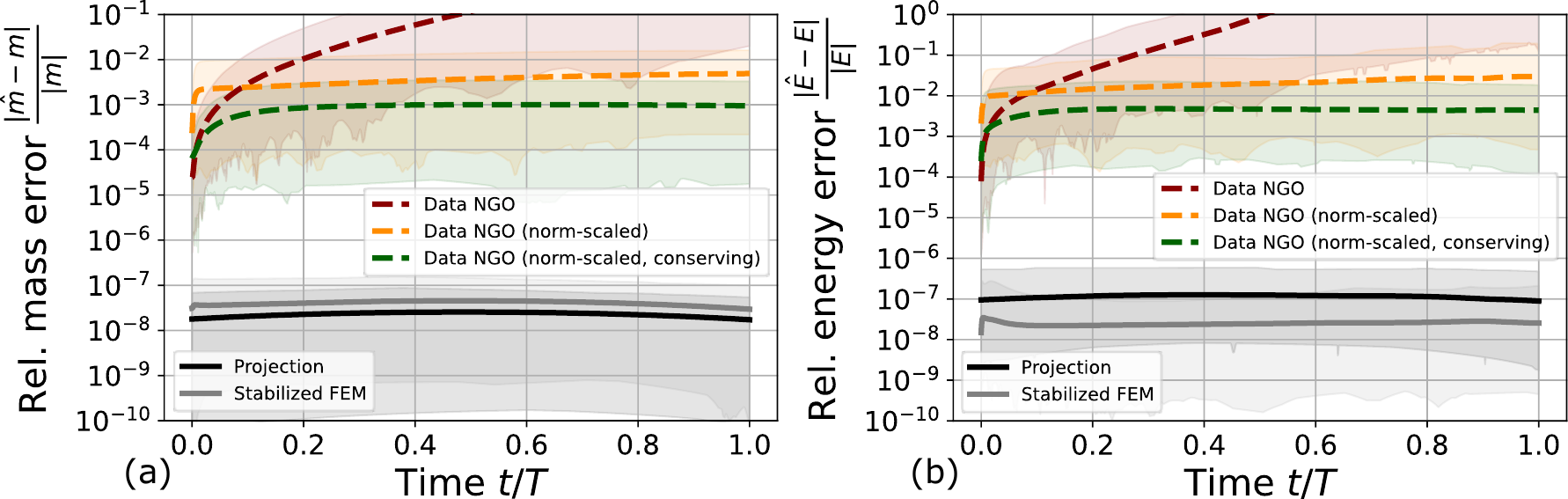}
\caption{\small{Effect of adding the conservation correction \eqref{E conservation ansatz} to a data NGO on the errors of its predicted (a) $m(t) =  
\int_\Omega u(\vb{x},t) d
\vb{x}$ and (b) energy $E(t) = 1/2
\int_\Omega u^2(\vb{x},t) d
\vb{x}$ over time. Projection and stabilized FEM are included as reference in black and gray, respectively. We note that the apparent mass conservation error in the projected solution arises from quadrature inaccuracies and would otherwise vanish.
All models are trained on dataset D (Appendix \ref{A data generation}). Lines and shaded areas are, respectively, averages and 95\% confidence intervals on 100 manufactured solutions generated in the same way as dataset D, but with $\lambda/L=\tau/T=1$ (where we used $L=T=1$)}.}\label{F conservationlaws}
\end{figure}

In Figure (\ref{F conservationlaws}), we show the effect of adding the correction in \eqref{E conservation ansatz} to a data NGO on its errors in the predicted mass $m(t) =  
\int_\Omega u(\vb{x},t) d\vb{x}$ and energy $E(t) = 1/2
\int_\Omega u^2(\vb{x},t) d\vb{x}$ over time.
Details of the model architectures are highlighted in Appendix \ref{A Model architectures time-dependent diffusion}. The models have been trained on dataset D, described in Appendix \ref{A data generation time-dependent diffusion}, using the training procedure described in Appendix \ref{A Training Procedure Time-Dependent Diffusion}.
While it can be observed in Figures (\ref{F conservationlaws}) (a) and (b) that the correction (\ref{E conservation ansatz}) gives an improvement in the error of the predicted mass $\hat{m}(t)$ and energy $\hat{E}(t)$ over time, with respect to $m(t)$ and $E(t)$, respectively, it does not eliminate such errors. Such a discrepancy is due to the fact the correction in \eqref{E conservation ansatz} only gives us control over the equality in the mass conservation equation of \eqref{E discrete conservation equations}, but not over the magnitudes of the different contributions, and only ensures the satisfaction of an inequality in the energy dissipation relation of \eqref{E discrete conservation equations} which may be inexact.

\begin{figure}[htb]
\centering
\includegraphics[width = 0.4\textwidth]{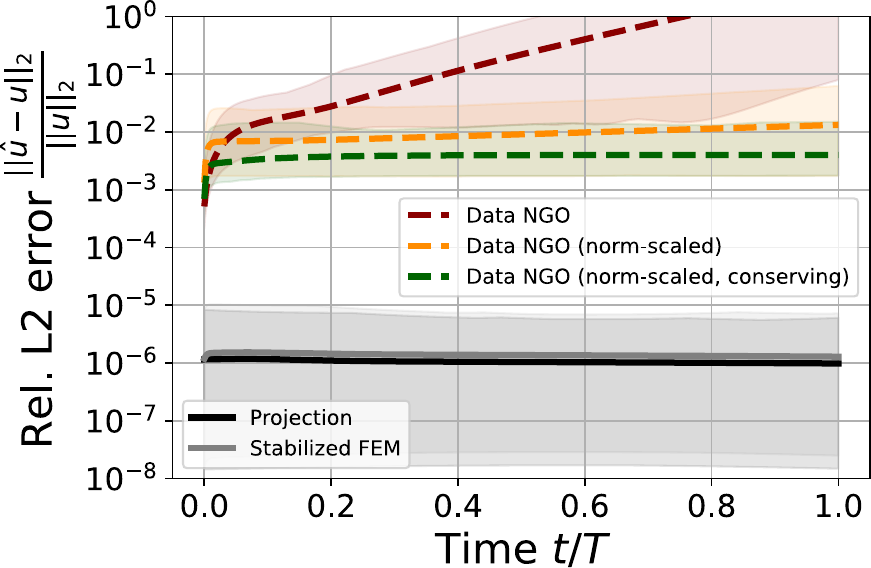}
\caption{\small{Effect of adding the conservation correction \eqref{E conservation ansatz} to a data NGO. Projection and stabilized FEM are included as reference in black and gray, respectively.
All models are trained on dataset D (Appendix \ref{A data generation}). Lines and shaded areas are, respectively, averages and 95\% confidence intervals on 100 manufactured solutions generated in the same way as dataset D, but with $\lambda/L=\tau/T=1$ (where we used $L=T=1$)}.}\label{F errorsovertime_C_D}
\end{figure}

Finally, Figure (\ref{F errorsovertime_C_D}) shows that incorporating the mass and energy relations in \eqref{E discrete conservation equations} through the correction \eqref{E conservation ansatz} to a data NGO results in an overall lower solution errors over time.

\subsubsection{Results}
In this section, we perform the same test as considered in Figure \ref{F neuraloperators}, except that here, we compare NGOs to other neural operators in their generalization across test data length as well as time scales. Additionally, we compare the long term stability of NGOs against the same models from literature. The architectural details of the used models are essentially identical to those corresponding to Figure \ref{F neuraloperators}, except that the models now also use quadrature and basis functions in space-time. Specifically, the DeepONet, VarMiON, data NGO, data-free NGO and model NGO now use a 2 linear $\times$ 10 cubic $\times$ 10 cubic B-spline basis in space-time  $(t,x,y)$ and all models use a quadrature grid of approximately $2\times 100 \times 100$ points. The FNO uses \(2 \times 10 \times 10\) Fourier modes, and the standalone NN (U-Net) uses 3-dimensional convolutions. The VarMiON is now limited to using only $2\times 5 \times 5$ uniform quadrature points, because of the limited parameter budget of $3\cdot 10^4$ trainable parameters. The inductive bias presented in Section \ref{S Inductive Bias: Time Stepping Stability} is incorporated in the data NGOs, data-free NGOs and model NGOs. Details of the model architectures are highlighted in Appendix \ref{A Model architectures time-dependent diffusion}. The models have been trained on dataset D, described in Appendix \ref{A data generation time-dependent diffusion}, using the training procedure described in Appendix \ref{A Training Procedure Time-Dependent Diffusion}.

\begin{figure}[htb]
\centering
\includegraphics[width = 0.75\textwidth]{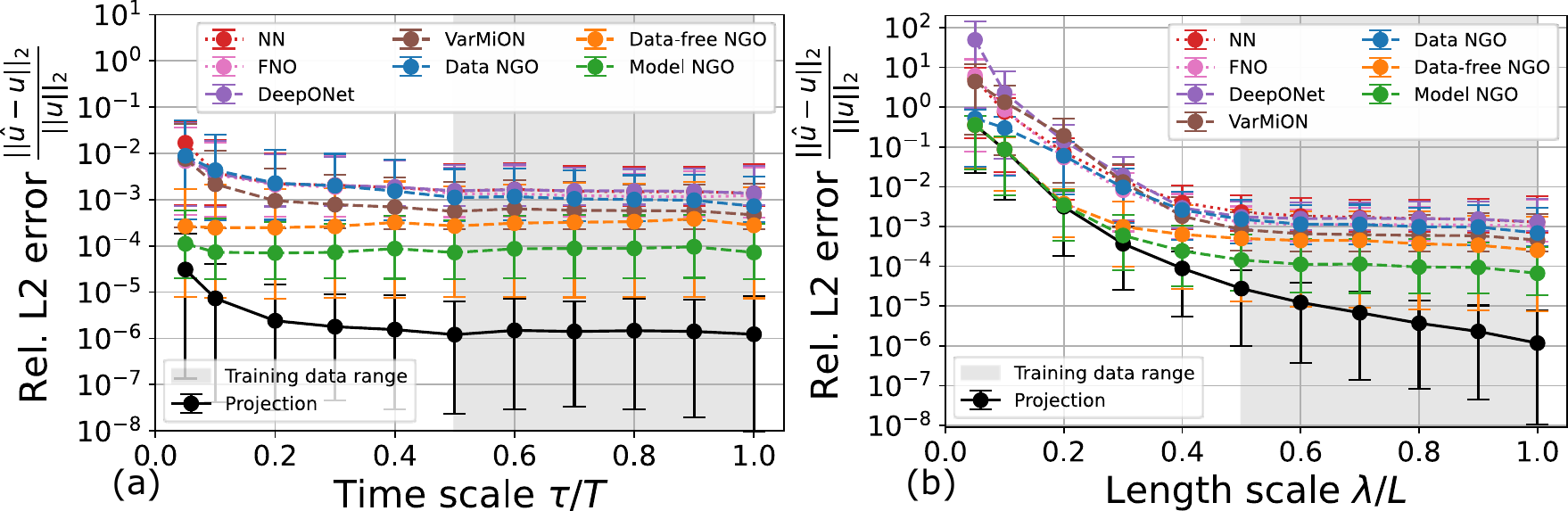}
\caption{\small{Single time step relative \(L^2\) test error versus (a) time scale $\tau/T$ and (b) length scale $\lambda/L$ for a standalone NN (here a U-Net), and the same NN, when used in a DeepONet, VarMiON, data NGO, data-free NGO and model NGO. All models have roughly $3 \cdot 10^4$ trainable parameters, and all models (except the NN and FNO) use the same 2 linear $\times$ 10 cubic $\times$ 10 cubic B-spline basis in space-time $(t,x,y)$. Points and error bars are, respectively, averages and 95\% confidence intervals on 1000 manufactured solutions generated in the same way as dataset D, but with $\lambda/L=1$ (in (a)) and with $\tau/T=1$ (in (b)) (where we used $L=T=1$)}.}\label{F dynamicdiffusion_scales}
\end{figure}

\paragraph{Generalization Error Across Length and Time Scales}
In Figure \ref{F dynamicdiffusion_scales}, the single timestep accuracies of the different NOs are compared on test datasets with varying GRF time and length scales $\tau$ and $\lambda$. The models are tested on in-distribution time and length scales $0.5<\tau/T<1$, $0.5<\lambda/L<1$, (dataset D), as well as on out-of-distribution test data with finer scales $0.05<\tau/T<0.5$ $0.05<\lambda/L<0.5$ (generated in the same way as dataset D). 
Observations are similar to those in Figure \ref{F neuraloperators} for the steady diffusion problem: the model NGO is the most accurate, followed by the data-free NGO. This is the case in the test with varying length scale as well as in the test with varying time scale, on large scale in-distribution data, as well as on fine scale out-of-distribution data. A difference with respect to the steady diffusion results presented in Figure \ref{F neuraloperators} is that the data NGO has slightly higher errors. The explanation is that we needed to compromise on single timestep accuracy, in order to achieve long-term stability. We observed that choosing a smaller, more conservative norm clipping value $S<1$ resulted in less error accumulation, and therefore, we chose $S=0.8$. However, $S=0.8$ is far off from the value $S \approx 0.99$ that agrees with the true data, which explains the higher single timestep error of the data NGO.

\begin{figure}[htb!]
\centering
\includegraphics[width=0.75\textwidth]{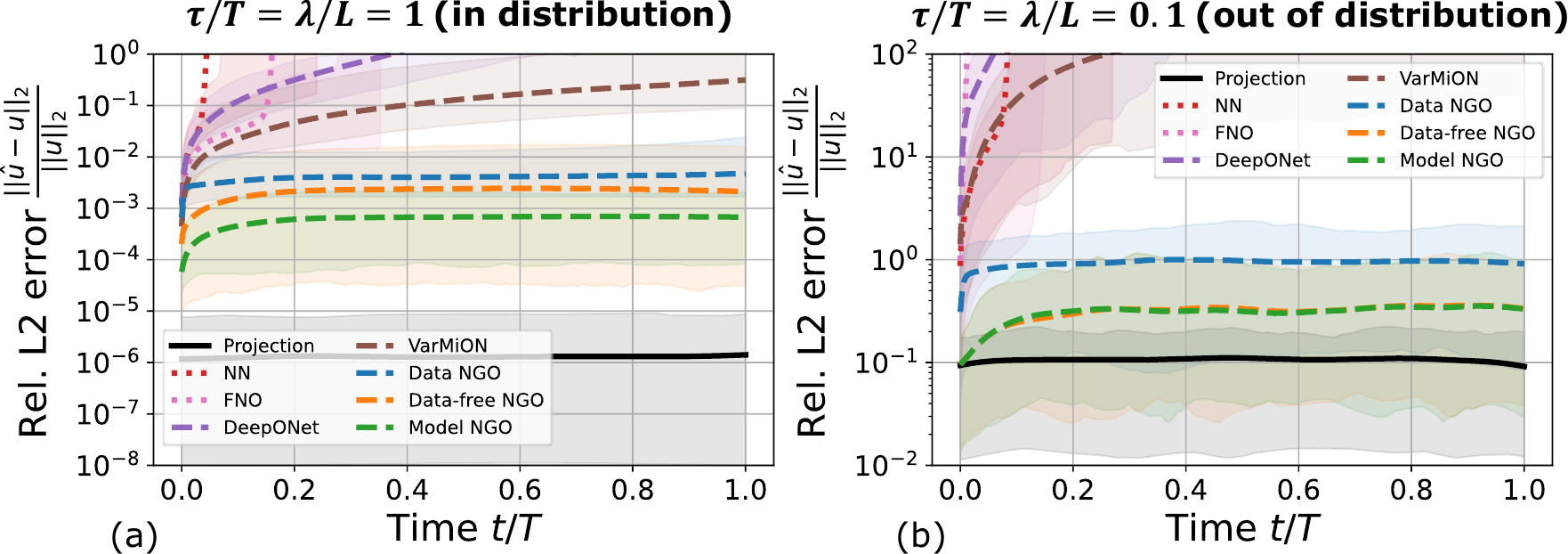}
\caption{\small{Relative time step $L^2$ test error over time on (a) in distribution data with time and length scale $\tau/T=\lambda/L=1$ and (b) out of distribution data with $\tau/T=\lambda/L=0.1$, for a standalone NN (here a U-Net), and the same NN, when used in a DeepONet, VarMiON, data NGO, data-free NGO and model NGO (an FNO is also added for comparison). All models have roughly \(3 \cdot 10^4\) trainable parameters, and all models (except the NN and FNO) use the same 2 linear $\times$ 10 cubic $\times$ 10 cubic B-spline basis, and are thus limited by the same \(L^2\) projection error lower bound, indicated in black. Points and shaded areas are, respectively, averages and 95\% confidence intervals on 100 manufactured solutions generated in the same way as dataset D, but with $\tau/T=\lambda/L=1$ (in (a)) and with $\tau/T=\lambda/L=0.1$ (in (b)) (where we used $L=T=1$)}.}\label{F errorsovertime}
\end{figure}

\paragraph{Stability of Different NOs Over Time}
Figure \ref{F errorsovertime} shows a comparison of the errors over time of the same set of models as presented in Figure \ref{F dynamicdiffusion_scales}. 
In Figure \ref{F errorsovertime}(a), the models are tested on a set of 100 different in-distribution problems with large time and length scales $\tau/T=\lambda/L=1$. Note that the single timestep error (Figure \ref{F dynamicdiffusion_scales}) only determines the model error after the first timestep: a lower single timestep error does not necessarily imply less accumulation of error over time. Only the NGOs manage to keep their errors finite over time, which demonstrates the effectiveness of the additional inductive bias that the Green's formulation of the NGO allows us to introduce. In Figure \ref{F errorsovertime}(b), we test the same models on 100 far out-of-distribution problems with time and length scales $\tau/T=\lambda/L=0.1$. While the best approximation in the space deteriorates in the presence of such fine scales in the data, the results do demonstrate the reliability of the NGOs, provided by the additional inductive bias, in far out-of-distribution conditions.

\FloatBarrier
\subsection{Advection-Diffusion}
The next test case we consider is the advection-diffusion equation with inhomogeneous Dirichlet and Neumann boundary conditions:
\begin{align}
    \begin{cases}
        \hfill -\nabla\cdot\left( \theta\nabla u \right) + \vb{c}\cdot\nabla u = f &\text{on } \Omega = (0, 1)^2, \\
        \hfill \theta \nabla u \cdot \vb{n} - \left( \vb{c}\cdot\vb{n} \right)_-u = \eta &\text{on } \Gamma_{\mathrm{N}} = (0, 1) \times \left\{ 0, 1 \right\}, \\
        \hfill \theta u = g &\text{on } \Gamma_{\mathrm{D}} = \left\{ 0, 1 \right\} \times (0, 1),
    \end{cases}
    \label{eq:advection-diffusion}
\end{align}
where \(\left( \vb{c}\cdot\vb{n} \right)_- = \min\left( \vb{c}\cdot\vb{n}, 0 \right)\). Compared to the pure diffusion equation, the advection-diffusion equation presents a more challenging problem due to a number of factors. First, the solution now depends nonlinearly on both the diffusion \(\theta\) and the velocity field \(\vb{c}\). Second, for large P\'eclet numbers (the ratio between the advection \(\vb{c}\) and the diffusion \(\theta\)) the true solution will feature boundary layers near the Dirichlet boundaries, which may be difficult to resolve. Note that the boundary condition \(\theta \nabla u \cdot \vb{n} - \left( \vb{c}\cdot\vb{n} \right)_-u = \eta\) differs from the Neumann boundary condition on the inflow part of the top and bottom boundaries (i.e.~where \(\vb{c}\cdot\vb{n} < 0\)). This change was made to facilitate stable finite element discretizations.

\begin{figure}[htb]
    \centering
    \includegraphics[width=0.8\linewidth]{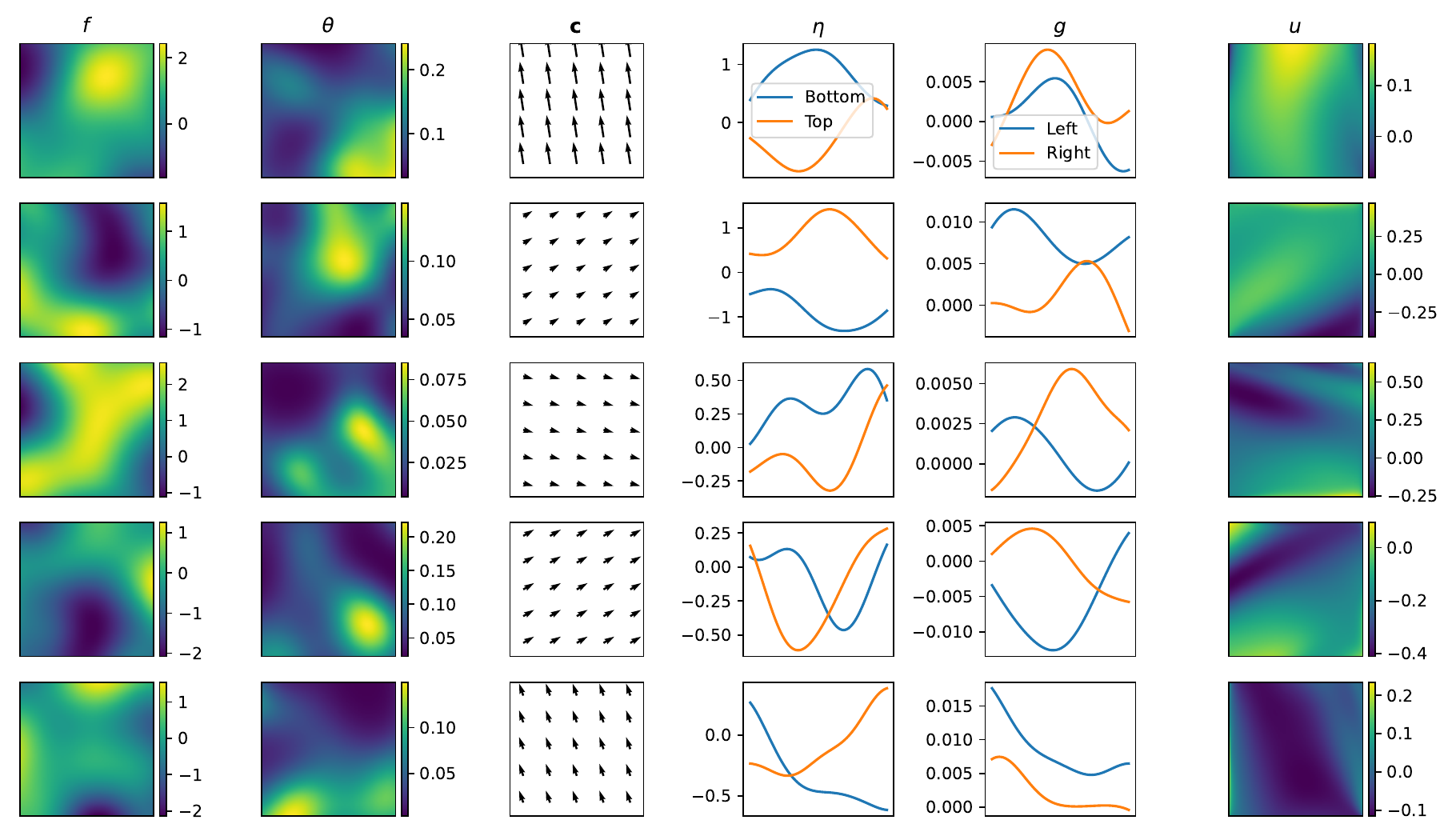}
    \caption{Five example problems and solutions from the advection-diffusion data set.}
    \label{fig:advection-diffusion-data-sample}
\end{figure}

Figure~\ref{fig:advection-diffusion-data-sample} shows five problems from the advection-diffusion data set. Note that the varying direction and magnitude of the velocity field means that different problems lead to boundary layers of varying strengths and near different parts of the boundary.

\subsubsection{NGO architectures}
The architectures of the NGOs are largely unchanged from their architectures in Section~\ref{S Generalization Errors of Different NOs}. The differences are listed in Appendix~\ref{S model architectures advection-diffusion}. Regarding the training procedure, all relevant parameters are still as given in Table~\ref{tab:training_parameters}, with the exception of the out-of-distribution training data, which is now absent.

\subsubsection{Results}
\newcommand{\x}{\phantom{0}}
\begin{table}
    \centering
    \caption{Errors (in terms of mean and standard deviation) of different model architectures on the advection diffusion data set.}
    \begin{tabular}{l | l | l | l}
        \toprule
        \multirow{2}*{Model} & \multirow{2}*{Parameters} & \multicolumn{2}{c}{Test error} \\
        & & Training set & Testing set \\
        \midrule
        DeepONet   & 36819 & \( 32.52\% \pm  15.46\%\) & \( 33.30\% \pm  19.52\%\) \\
        VarMiON    & 32233 & \( 12.21\% \pm \x7.16\%\) & \( 12.30\% \pm \x6.89\%\) \\
        FNO        & 31817 & \(\x8.95\% \pm \x2.72\%\) & \( 11.49\% \pm \x5.80\%\) \\
        CNO        & 29837 & \( 15.43\% \pm \x6.46\%\) & \( 21.23\% \pm  10.30\%\) \\
        U-Net      & 31721 & \( 17.88\% \pm \x6.85\%\) & \( 18.02\% \pm \x7.29\%\) \\
        \midrule
        Data NGO   & 28021 & \(\x6.82\% \pm \x3.26\%\) & \(\x7.66\% \pm \x3.99\%\) \\
        Model NGO  & 28387 & \(\x7.06\% \pm \x3.34\%\) & \(\x7.98\% \pm \x4.11\%\) \\
        \midrule
        Galerkin   &       & \( 14.13\% \pm  12.46\%\) & \( 14.57\% \pm  14.10\%\) \\
        SUPG       &       & \(\x6.93\% \pm \x5.92\%\) & \(\x7.00\% \pm \x6.21\%\) \\
        Projection &       & \(\x3.99\% \pm \x3.43\%\) & \(\x4.06\% \pm \x3.66\%\) \\
        POD-Galerkin &     & \( 30.42\% \pm  19.25\%\) & \( 29.16\% \pm  16.52\%\) \\
        \bottomrule
    \end{tabular}
    \label{tab:advection-diffusion-results}
\end{table}

Table~\ref{tab:advection-diffusion-results} shows the training and testing error of the data NGO, model NGO, and FNO on the advection-diffusion data. For reference, the table also includes the SUPG, Galerkin and \(L^2\) projection errors, both using the same cubic B-spline basis as the NGOs. The table also lists the accuracy of POD-Galerkin, a simple reduced basis method that produces excellent results for the pure diffusion equation (see Table~\ref{tab:heat_eq_results}). For advection-diffusion, POD-Galerkin produces worse results due to the fact that Galerkin methods are generally unsuitable for advection-dominated problems.

\begin{figure}
    \centering
    \begin{subfigure}{0.9\linewidth}
        \centering
        \includegraphics[width=0.8\linewidth]{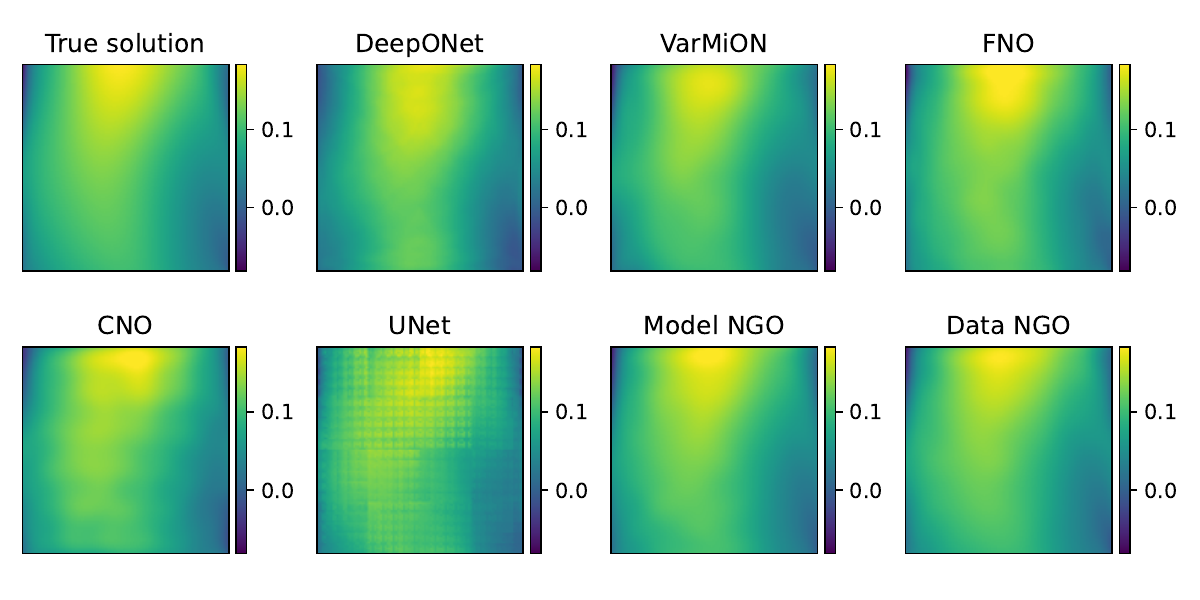}
        \caption{The true solution of one problem from the training data set, and model predictions of the different neural operators.}
        \label{fig:advection-diffusion-example-1}
    \end{subfigure}
    \begin{subfigure}{0.9\linewidth}
        \centering
        \includegraphics[width=0.8\linewidth]{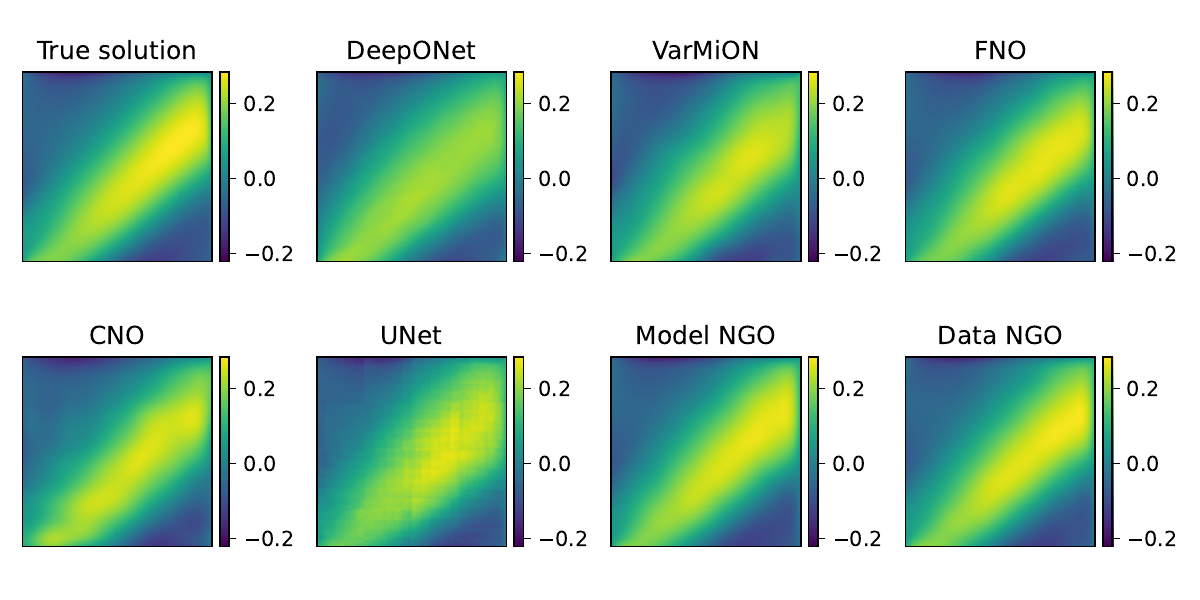}
        \caption{The true solution of one problem from the testing data set, and model predictions of the different neural operators.}
        \label{fig:advection-diffusion-example-2}
    \end{subfigure}
    \caption{The true solution and model predictions to one problem from the training data set (Figure~\ref{fig:advection-diffusion-example-1}) and one problem from the testing data set (Figure~\ref{fig:advection-diffusion-example-2}).}
    \label{fig:advection-diffusion-example}
\end{figure}

Figure~\ref{fig:advection-diffusion-example} shows the true solution alongside the solution predictions made by the different model architectures for one problem from the training data, and one from the testing data.

To highlight the computational efficiency of the different models we show the time and memory requirements of CNOs, FNOs, and NGOs per batch of 100 inputs in Table~\ref{tab:advection-diffusion-performance}. From these results, it is clear that NGOs require more time and less memory to train than U-Nets and VarMiONs, but less than CNOs and FNOs.
\begin{table}
    \centering
    \caption{Performance comparison, in terms of time and memory usage, of NGOs compared to other models.}
    \begin{tabular}{l | r | r}
        \toprule 
        Model & Computing time per batch [ms] & Peak memory usage [MB] \\ 
        \midrule
        Model NGO &   9.3 &  522 \\
        Data NGO  &   6.1 &  347 \\
        VarMiON   &   3.6 &   87 \\
        FNO       & 107.3 & 4201 \\
        CNO       &  43.2 &  880 \\
        U-Net     &   5.6 &  260 \\
        \bottomrule
    \end{tabular}
    \label{tab:advection-diffusion-performance}
\end{table}

\begin{figure}[htb]
    \centering
    \includegraphics[width=0.4\linewidth]{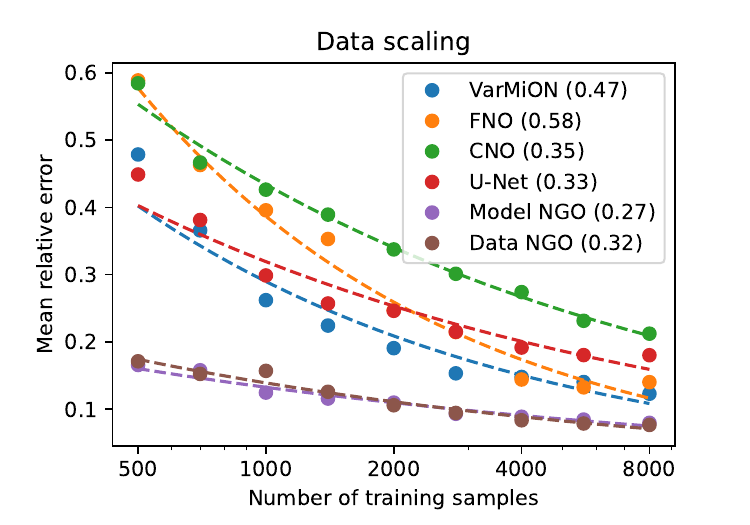}
    \quad
    \includegraphics[width=0.4\linewidth]{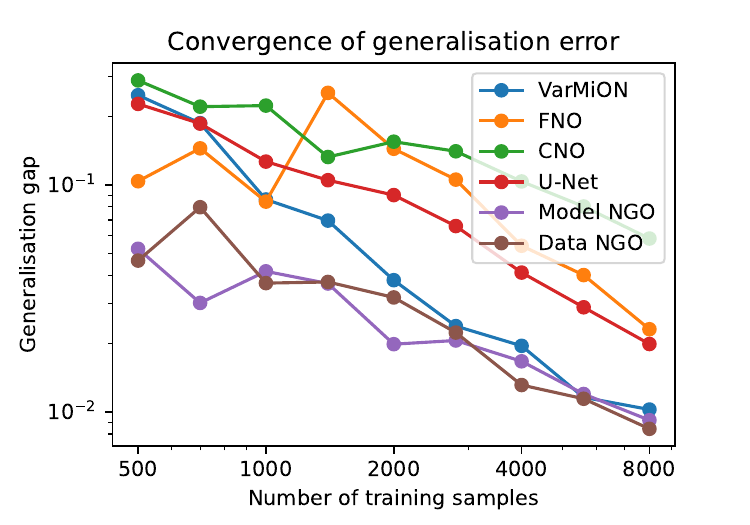}
    \caption{Left: mean relative error of VarMiON, FNO, CNO, U-Net, model NGO, and data NGO, each for varying amounts of training data. The fitted curves are of the form \(\text{err} = a \cdot N^{-r}\). The exponent \(r\) is listed in parentheses for each model. Right: the generalization gap, i.e.~the difference between the testing error and the training error, for varying amounts of training data.}
    \label{fig:advection-diffusion-data-scaling}
\end{figure}

To illustrate the scaling of the error with the data-set size we show in Figure~\ref{fig:advection-diffusion-data-scaling}  the effect that the amount of training data has on the model accuracy for FNOs, CNOs, data NGOs, and model NGOs. As the plot shows, both NGO models outperform the other models for each tested data set size, although the difference becomes small for large data sets.  Figure~\ref{fig:advection-diffusion-data-scaling} also shows the generalization gap, i.e.~the difference between the training and testing error, which shows that NGOs consistently generalize better than FNOs, CNOs, and U-Nets. The VarMiON generalizes approximately as well as NGOs for relatively large data sets, but shows similar generalization to other models when the training data set is small.

\subsection{Nonlinear Diffusion}
\label{S nonlinear}
The final test case we consider is the nonlinear diffusion problem
\begin{align}
    \begin{cases}
        \hfill -\nabla\cdot\left(\theta\left[u\right]\nabla u \right) = f &\text{on } \Omega = (0, 1)^2, \\
        \hfill \theta\left[u\right] \nabla u \cdot \vb{n} = \eta &\text{on } \Gamma_{\mathrm{N}} = (0, 1) \times \left\{ 0, 1 \right\}, \\
        \hfill \theta\left[u\right] u = g &\text{on } \Gamma_{\mathrm{D}} = \left\{ 0, 1 \right\} \times (0, 1),
    \end{cases}
    \label{E nonlinear diffusion}
\end{align}
where 
\begin{equation}\label{E nonlinear theta}
    \theta\left[u\right](\vb{x}) = \theta_0(\vb{x}) + \alpha u(\vb{x})
\end{equation}
is a diffusion coefficient that depends linearly on the solution $u$. Here, $\theta_0(\vb{x})$ is a spatially varying, solution-independent part of the diffusion coefficient $\theta$, and $\alpha=0.1$ is a constant. Like in the steady diffusion problem \ref{S Test problem: steady diffusion}, the domain $\Omega$ is the unit square, shown in Figure \ref{F unitsquare}, Neumann boundary conditions $\eta(\vb{x})$ are applied on the top and bottom boundaries, and Dirichlet boundary conditions $g(\vb{x})$ are applied on the left and right boundaries.

Classical solution strategies to solve nonlinear PDEs like (\ref{E nonlinear diffusion}) often rely on iterative methods that formulate the solution to the nonlinear problem as a sequence of successive approximations to an associated linear parametric subproblem \cite{quarteroni2006numerical}. NGOs are well suited to accelerate such methods by directly replacing the expensive step of computing the solution to the parametrized linear subproblem within an iterative loop with an operator evaluation. To demonstrate the effectiveness of embedding NGOs into nonlinear iterative solvers, we consider the Picard iteration method \cite{kelley1995iterative, quarteroni2006numerical} for approximating solutions of \eqref{E nonlinear diffusion}. Starting from an initial guess $u^{(0)}=0$, the Picard iterates $\{u^{(k)}\}_{k\ge1}$, where $k \in \{1,2,\ldots\}$, are defined by solving:
\begin{align}
    \begin{cases}
        \hfill -\nabla \cdot \bigl(\theta[u^{(k-1)}]\nabla u^{(k)}\bigr) = f
        & \text{in } \Omega = (0,1)^2, \\[4pt]
        \hfill \theta[u^{(k-1)}]\nabla u^{(k)} \cdot \mathbf{n} = \eta
        & \text{on } \Gamma_{\mathrm{N}} = (0,1)\times\{0,1\}, \\[4pt]
        \hfill \theta[u^{(k-1)}]\,u^{(k)} = g
        & \text{on } \Gamma_{\mathrm{D}} = \{0,1\}\times(0,1).
    \end{cases}
    \label{E picard}
\end{align}
In this framework, a sequence of solutions $u^{(k)}$ of the linearized subproblems provides successive approximations of the solution $u$ of nonlinear PDE, given the previous iterate $u^{(k-1)}$.

Note that for each $k \in \{1,2,\ldots\}$, the system \eqref{E picard} is an instance of the parametric steady diffusion problem \eqref{eq:darcy}, obtained by setting $u \to u^{(k)}$ and $\theta \to \theta[u^{(k-1)}]$, and therefore the solution to (\ref{E picard}) can be written in terms of a parametric Green's operator as shown in (\ref{E Green's operator Darcy}). This observation enables the direct reuse of an NGO trained for the parametric linear problem, described in section \ref{S Generalization Errors Across Length Scales}, as a surrogate solver within each Picard iteration. However, we observed in our FEM tests for this problem that stabilization of our weak form, as described in remark \ref{rem:nitsche}, is necessary for the convergence of the Picard iterations. While the data NGOs pretrained on the linear problem described in \ref{S Generalization Errors Across Length Scales} can be reused directly, the preconditioned data-free and model NGOs needed to account for such stabilization by adding the Nitsche stabilization terms \eqref{E Nitsche F} and \eqref{E Nitsche d} to the input matrix $\vb{F}$, the Neumann preconditioner $\vb{F}_0$ (described in section \ref{S Preconditioning the System Network}), and to the right-hand-side vector $\vb{d}$, respectively.

\subsubsection{Results}
In this section, we assess the effectiveness of embedding NGOs, trained on solutions of the parametric linear diffusion problem described in Section~\ref{S Test problem: steady diffusion}, within a Picard iteration (\ref{E picard}) to solve the nonlinear diffusion problem \eqref{E nonlinear diffusion} parametrized by $(\theta_0, f, \eta, g)$. We compare the resulting performance against NO architectures from the literature that are capable of directly approximating solutions to nonlinear PDEs without resorting to iterative schemes. Specifically, we consider a U-Net (denoted NN), a DeepONet, and a FNO. For completeness, we also evaluate the performance of the U-Nets, DeepONets, and FNOs under two training and deployment strategies: 
\begin{itemize}
    \item[(i)] Direct approximation of the solution operator for the parametric nonlinear problem. For this strategy the architectures and training procedure of the NNs, FNOs, DeepONets, are also identical to those in Section \ref{S Generalization Errors Across Length Scales} (Appendices \ref{A Model definition details of Section} and \ref{A training procedure steady diffusion Joost}, respectively), however, these have been trained on nonlinear diffusion dataset F, described in Appendix \ref{A data generation nonlinear diffusion}.
    \item[(ii)] Training on the corresponding linear parametric problem and embedding the resulting models within Picard iterations, analogous to our approach with NGOs. For this strategy we reuse the trained models presented in Section \ref{S Generalization Errors Across Length Scales} for the linear diffusion problem: their architectures, training procedure and dataset (dataset C) are highlighted in Appendices \ref{A Model definition details of Section}, \ref{A training procedure steady diffusion Joost} and \ref{sec:darcy-data-generation}, respectively.
\end{itemize}
This comparison allows us to isolate the benefits of NGOs from those attributable solely to iterative deployment.  

\begin{figure}[htb!]
\centering
\includegraphics[width=\textwidth]{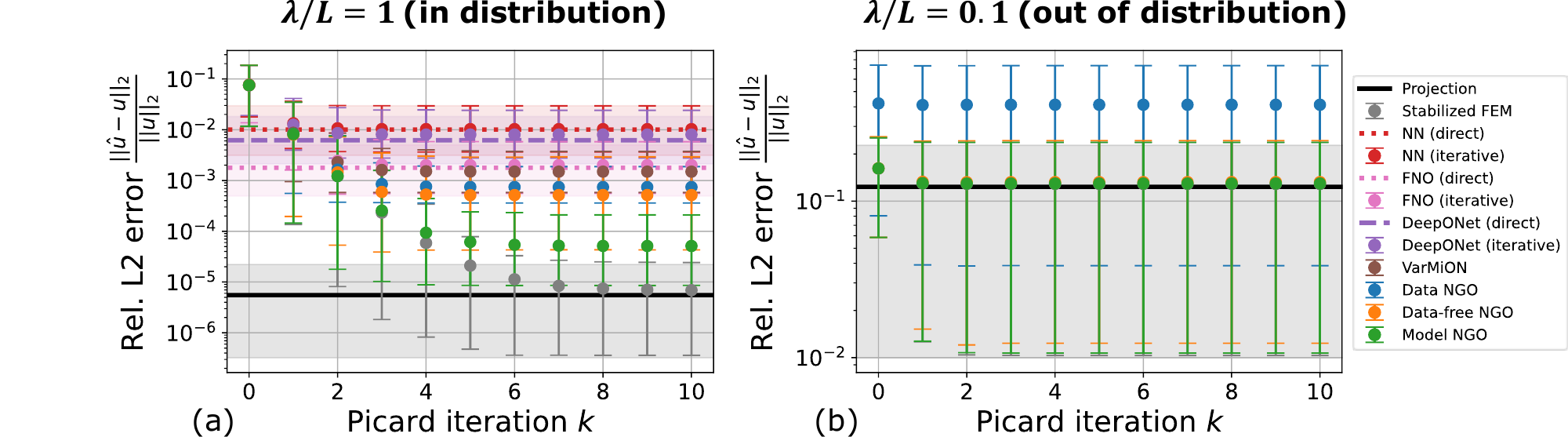}
\caption{\small{Relative $L^2$ test error over fixed point iteration number $k$ on (a) in distribution data with length scale $\lambda/L=1$ and (b) out of distribution data with $\lambda/L=0.1$ (where we used $L=1$), for a standalone NN (here a U-Net), and the same NN, when used in a DeepONet, VarMiON, data NGO, data-free NGO and model NGO (an FNO is also added for comparison). For the NN, DeepONet and FNO, results are included for a direct as well as for an iterative solution approach. All models have roughly \(3 \cdot 10^4\) trainable parameters, and all models (except the NN and FNO) use the same 10 $\times$ 10 cubic B-spline basis, and are thus limited by the same \(L^2\) projection error lower bound, indicated in black. Points and lines are averages, and error bars and shaded areas are 95\% confidence intervals, over datasets of 1000 manufactured solutions generated in the same way as Dataset F (Appendix \ref{A data generation nonlinear diffusion}).}}\label{F fixedpoint}
\end{figure}

Figure~\ref{F fixedpoint} reports the relative error of the different network architectures as a function of the Picard iteration index $k$. These curves are overlaid with horizontal lines indicating the errors achieved by networks that directly approximate the solution operator of the nonlinear problem.
In Figure~\ref{F fixedpoint}(a) we present the results of in-distribution testing for a length scale $\lambda/L = 1$. We observe that all networks deployed within Picard iterations initially mirror the convergence trajectory of the stabilized FEM; however, they eventually stagnate at higher errors. Among all models considered, the model NGO most closely tracks the convergence behavior of the stabilized FEM and achieves the lowest final error, followed by the data-free NGO and the data NGO. Furthermore, we also observe that the iterative use of the NN, FNO, and DeepONet converges to error levels comparable to those obtained by their direct nonlinear counterparts, indicating that iterative deployment does not confer a clear advantage over direct approximation strategy for these architectures in our tests.

\begin{figure}[htb!]
\centering
\includegraphics[width = 0.75\textwidth]{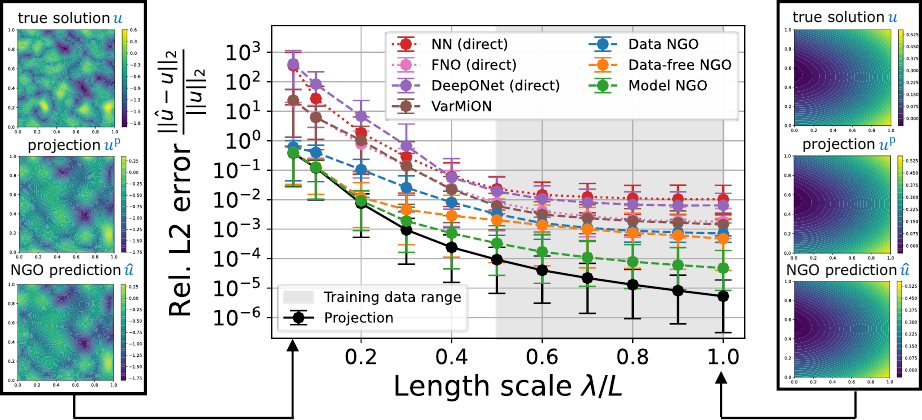}
\caption{\small{Relative \(L^2\) test error versus test data length scale $\lambda/L$ (where we used $L=1$) for a standalone NN (here a U-Net), and the same NN, when used in a DeepONet, VarMiON, data NGO, data-free NGO and model NGO. 
All models have roughly \(3\cdot10^4\) trainable parameters, and all models (except the NN and FNO, also added as reference) use the same \(10 \times 10\) cubic B-spline basis, and are thus limited by the same \(L^2\) projection error lower bound, indicated in black. Points and error bars are, respectively, averages and 95\% confidence intervals on datasets of 1000 manufactured nonlinear diffusion solutions, generated in the same way as dataset F (Appendix \ref{A data generation nonlinear diffusion}).}}
\label{F lengthscales_NL}
\end{figure}

Figure~\ref{F fixedpoint}(b) reports the results of out-of-distribution testing for a finer length scale $\lambda/L = 0.1$. Here we observe that all NGO variants converge in a single Picard iteration, with the errors of the data-free and model NGOs visually overlapping with the projection error. For clarity, Figure~\ref{F fixedpoint}(b) omits the results for the NN, FNO, DeepONet, and VarMiON, as their relative errors are significantly larger. This trend is further corroborated in Figure~\ref{F lengthscales_NL}, which summarizes the approximation errors of all considered neural operators across a range of test length scales $\lambda/L$.

The results presented in Figure \ref{F lengthscales_NL} closely mirror the trends observed for the linear diffusion problem in Figure \ref{F neuraloperators}. Specifically, NGOs maintain their superior performance when applied to nonlinear diffusion, exhibiting both higher in-distribution accuracy and significantly more robust generalization to fine-scale, out-of-distribution data. However, as NGOs function as structural components of an iterative solver rather than end-to-end mappings, they inherit the requirements of conventional numerical methods: their convergence relies on the availability of a suitable initial guess, and they require multiple operator evaluations per solution. This stands in contrast to other NO architectures, which provide a direct, albeit less generalizable, single-pass approximation.

\begin{remark}
We observe that, for out-of-distribution fine-scale test data, the performance of the data-free NGO exhibits variability: among multiple independently and equivalently trained models, stable convergence was not consistently achieved. Consequently, we report results for the best-performing model selected from five such training runs. Developing inductive biases that explicitly promote stability of fixed-point iterations, analogous to the inductive biases for time-stepping stability introduced in Section~\ref{S Inductive Bias: Time Stepping Stability}, is an important direction for future work.
\end{remark}

\section{Conclusion}\label{S Conclusion}
In this work, we have introduced the Neural Green's Operator, a machine learning architecture for learning the solution operator to parametric families of linear PDEs. Our construction of NGOs is based on a finite-dimensional representation of Green's functions. Such a representation enables the preservation of the linear action of Green's operators on the inhomogeneity fields, while approximating the nonlinear dependence of the Green's function on the coefficients of the PDE using neural networks that take weighted averages of such coefficients as input. This construction reduces the complexity of the problem from learning the entire solution operator and its dependence on all parameters to only learning the Green's function and its dependence on the PDE coefficients. Moreover, taking weighted averages, rather than point samples, of input functions decouples the network size from the number of sampling points, enabling efficient resolution of multiple scales in the input fields. We distinguished three use cases of NGOs: the model NGO, which is used when the underlying PDE is known and training data is available; the data-free NGO, which is used when the underlying PDE is known, but training data is unavailable; and the data NGO, which is used when training data is available, but the model that underlies the data is unknown. For the steady diffusion problem, the accuracy of NGOs exceeds that of VarMiONs, DeepONets, and FNOs, particularly when testing models on finer-scale data that do not lie in the training distribution. Similar trends were observed for the time-dependent diffusion problem, the advection-diffusion problem and the nonlinear diffusion problem.

We show that an explicit representation of the Green’s function enables the incorporation of inductive biases directly into the network architecture. In particular, preconditioning the output of the system network leads to improved solution accuracy, while for time-dependent problems, embedding operator stability together with conservation and dissipation laws enables stable long-term autoregressive rollouts and, consequently, pointwise accuracy in time. Furthermore, we demonstrated that Green's functions inferred by NGOs can be used to construct effective matrix preconditioners for accelerating iterative solvers of large-scale problems. Preconditioners based on NGOs mimic popular two-level preconditioning strategies, making them applicable to a broad range of equations and linear solvers. Furthermore, NGO-based preconditioners were found to be significantly more effective than a similar strategy using DeepONets.

As a demonstration of the extension of NGOs to parametric nonlinear problems, we show that NGOs trained exclusively on solutions of associated linear PDEs can be effectively embedded within iterative nonlinear solvers. When supplied with a suitable initial guess, this approach yields accurate approximations to solutions of nonlinear parametric PDEs without retraining the model on nonlinear data. Moreover, the accuracy trends observed for the steady linear diffusion problem carry over to the nonlinear setting: NGOs consistently outperform VarMiONs, DeepONets, and FNOs, particularly in out-of-distribution regimes involving finer spatial scales than those seen during training. These results underscore the potential of NGOs to serve as reusable, modular components within classical numerical algorithms, effectively bridging linear training and nonlinear deployment.

While we provide a proof of concept for the use of NGOs as solution operators for PDEs, several research directions emerge from this work. One research direction includes enriching the space of parameters that are encoded in NGOs to solve PDEs in more general settings. Such settings can include varying geometries of the problem domain, as well as varying the type of boundary condition applied on the boundary segments. Another direction of research involves the customized use of NGOs to accelerate other algorithms. Of particular interest is the extension of the matrix preconditioning strategies for accelerating iterative solvers for large-scale nonlinear systems, and time integration strategies for accelerating transient solvers. Lastly, we are also interested in exploring training strategies, such as multi-stage approaches \cite{wang2024multi}, that would lead to even more accurate models.

Code corresponding to the steady diffusion problem is available at \url{https://github.com/JoostHMPrins/ngo}.

\section*{Acknowledgments}
This project has received funding from the Eindhoven Artificial Intelligence Institute under the Exploratory Multidisciplinary AI Research program for the NGO-PDE project. The authors thank Barry Koren of Eindhoven University of Technology for his valuable insights and helpful comments.

\bibliography{citations}

\appendix
\section{Derivation of Green's Operators}
\label{sec:greens-function-derivation}
\subsection{Steady Diffusion}
\label{sec:appendix-darcy-greens-function}
To derive the Green's operator for the steady diffusion equation \eqref{E d Darcy} as given by \eqref{E Green's operator Darcy}, we follow the approach of Section \ref{S Green's operators}. We start with testing the governing equation against a test function $v$ as
\begin{equation}\label{E Darcy 1}
    -\int_\Omega v \nabla \cdot \theta \nabla u d\vb{x}' = \int_\Omega v f d\vb{x}',
\end{equation}
and perform integration by parts twice on the left-hand side of \eqref{E Darcy 1} to get
\begin{equation}
    -\int_\Omega u \nabla \cdot \theta \nabla v d\vb{x}' + \int_{\Gamma} v \vu{n}\cdot \theta \nabla u d\vb{x}' - \int_{\Gamma} \theta u \vu{n}\cdot \nabla v d\vb{x}' = \int_\Omega v f d\vb{x}',
\end{equation}
where the first term on the left-hand side is the adjoint term. If we now split the integral in the second and third term over $\Gamma_{\mathrm{N}}$ and $\Gamma \setminus \Gamma_{\mathrm{N}}$, and over $\Gamma_{\mathrm{D}}$ and $\Gamma \setminus \Gamma_{\mathrm{D}}$, respectively, and substitute the boundary data $\eta$ and $g$, respectively, and rearrange terms such that all unknown terms appear on the left-hand side, we get
\begin{equation}\label{E weak form x}
    B[v,u] = L[v],
\end{equation}
where
\begin{equation}\label{E Darcy 2}
B[v,u] =-\int_\Omega u \nabla \cdot \theta \nabla v d\vb{x}' 
    - \int_{\Gamma\setminus \Gamma_{\mathrm{N}}} \theta v \vu{n}\cdot \nabla u d\vb{x}' 
    + \int_{\Gamma\setminus \Gamma_{\mathrm{D}}} \theta u \vu{n}\cdot \nabla v d\vb{x}' 
\end{equation}
and
\begin{equation}
    L[v] = \int_\Omega v f d\vb{x}' 
    + \int_{\Gamma_{\mathrm{N}}} v \eta d\vb{x}' 
    - \int_{\Gamma_{\mathrm{D}}} g \vu{n}\cdot \nabla v d\vb{x}'.
\end{equation}
If we now define the test function $v$ to be a Green's function $G$ that satisfies
\begin{equation}\label{E definition G Darcy}
    \begin{cases}
    \begin{aligned}
        -\nabla \cdot \theta(\vb{x}') \nabla G(\vb{x},\vb{x}') &= \delta(\vb{x} - \vb{x}') &&\forall\vb{x}'\in \Omega, \\
        \theta(\vb{x}') G(\vb{x},\vb{x}') &= 0 &&\forall \vb{x}'\in \Gamma \setminus \Gamma_{\mathrm{N}}, \\
        \vu{n} \cdot \theta(\vb{x}') \nabla G(\vb{x},\vb{x}') &= 0 &&\forall \vb{x}'\in \Gamma \setminus \Gamma_{\mathrm{D}},
    \end{aligned}
    \end{cases}
\end{equation}
$B[G,u]$ reduces to $u(\vb{x})$, and thereby, the weak form \ref{E weak form x} reduces to the Green's operator \ref{E Green's operator Darcy}.

\subsection{Time-Dependent Diffusion}\label{A G tdd}
The derivation of the Green's operator for the time-dependent diffusion problem as given by \eqref{E G tdd} is analogous to the Green's function of the steady problem, except that now all functions and basis functions are functions of space and time $(\vb{x},t)$, and the integrals extend over the space and time domains $\Omega=[0,L]^2$ and $\Delta t_i=[\Delta t,(i+1)\Delta t]$. Additionally, the time derivative in the operator gives rise to a few additional terms in the weak form. If we test the term of the time derivative $\partial u^{(i)}/\partial t$ against the test function $G$ and integrate by parts, we get
\begin{equation}
    \int_{\Omega\times \Delta t_i} v \pdv{u^{(i)}}{t} d\vb{x}'dt' = \int_{\Omega\times \Delta t_i} \pdv{}{t}\left(vu^{(i)}\right) d\vb{x}'dt' - \int_{\Omega\times \Delta t_i} u^{(i)} \pdv{v}{t} d\vb{x}'dt'.
\end{equation}
Integrating the first term on the right-hand side gives
\begin{equation}
    \int_{\Omega\times \Delta t_i} v \pdv{u^{(i)}}{t} d\vb{x}'dt' = \int_{\Omega}\left[ vu^{(i)}\right]_{t'=i\Delta t}^{t'=(i+i)\Delta t}d\vb{x}' - \int_{\Omega\times \Delta t_i} u^{(i)} \pdv{v}{t} d\vb{x}'dt',
\end{equation}
and substitution of the initial condition $u^{(i)}(i\Delta t)=u^{(i-1)}(i\Delta t)$ gives
\begin{equation}
\begin{aligned}
    \int_{\Omega\times \Delta t_i} v \pdv{u^{(i)}}{t} d\vb{x}'dt' &= \int_{\Omega}v((i+1)\Delta t)u^{(i)}((i+1)\Delta t) d\vb{x}' -\int_{\Omega}v(i\Delta t)u^{(i-1)}(i\Delta t) d\vb{x}' \\
    &- \int_{\Omega\times \Delta t_i} u^{(i)} \pdv{v}{t} d\vb{x}'dt'.
\end{aligned}
\end{equation}
Adding all terms of \eqref{E Darcy 2}, which belong to the spatial part $-\nabla \cdot \theta \nabla$ of the operator, we get the weak form
\begin{equation}\label{E unstabilized weak form}
    B[v,u^{(i)}]=L[v].
\end{equation}
Here,
\begin{equation}\label{E bilinear form tdd}
\begin{aligned}
    B[v,u^{(i)}] &= \int_{\Omega\times \Delta t_i} u^{(i)} \mathcal{L}^*[\theta] v d\vb{x}'dt'
    - \int_{\Gamma\setminus \Gamma_{\mathrm{N}}\times \Delta t_i}\theta v \vu{n}\cdot \nabla u^{(i)} d\vb{x}'dt'
    + \int_{\Gamma\setminus \Gamma_{\mathrm{D}}\times \Delta t_i} \theta u^{(i)} \vu{n}\cdot \nabla v d\vb{x}'dt' \\
    &+ \int_{\Omega}v((i+1)\Delta t)u^{(i)}((i+1)\Delta t) d\vb{x}'
\end{aligned}
\end{equation}
and
\begin{equation}
    L[v] = \int_{\Omega}v(i\Delta t)u^{(i-1)}(i\Delta t) d\vb{x}'
    +\int_{\Omega\times \Delta t_i} v f d\vb{x}'dt' 
    + \int_{\Gamma_{\mathrm{N}}\times \Delta t_i} v \eta d\vb{x}'dt' 
    - \int_{\Gamma_{\mathrm{D}}\times \Delta t_i} g \vu{n}\cdot \nabla v d\vb{x}'dt',
\end{equation}
where $\mathcal{L}^*[\theta]=-\partial/\partial t-\nabla \cdot \theta \nabla$. If we now define the test function $v$ to be a Green's function $G$ that satisfies
\begin{equation}\label{E definition G Darcy 2}
    \begin{cases}
    \begin{aligned}
        \mathcal{L}^*[\theta] G^{(i)} &= \delta &&\forall(\vb{x}',t')\in \Omega\times \Delta t_i, \\
        \theta G^{(i)} &= 0 &&\forall (\vb{x}',t') \in \Gamma \setminus \Gamma_{\mathrm{N}}\times \Delta t_i, \\
        \vu{n} \cdot \theta \nabla G^{(i)} &= 0 &&\forall (\vb{x}',t')\in \Gamma \setminus \Gamma_{\mathrm{D}}\times\Delta t_i, \\
        G^{(i)} &=0 &&\forall (\vb{x}',t')\in \Omega\times\mathcal\{0\},
    \end{aligned}
    \end{cases}
\end{equation}
$B[G,u]$ reduces to $u(\vb{x},t)$, which leaves us with the Green's operator \eqref{E G tdd}.

\subsection{Advection-diffusion}
For the advection-diffusion tests, the NGO formulation is not based directly on Green's function formulation. Instead, we use the weak form derived by Bazilevs and Hughes~\cite{Bazilevs_Hughes_2005}, modified to handle the Neumann/Robin boundary conditions:
\begin{align*}  
    &(-\nabla v, \vb{c}u - \theta\nabla u)_\Omega
    + (\mathbb{L}v\tau, \mathcal{L}u)_\Omega
    + (v, -\theta\nabla u\cdot\vb{n} + \vb{c}\cdot\vb{n}u)_{\Gamma_D}
    + (-\gamma\theta\nabla v\cdot\vb{n} - \vb{c}\cdot\vb{n}v, u)_{\Gamma_{D,\text{in}}} \\
    & + (-\gamma\theta\nabla v\cdot\vb{n}, u)_{\Gamma_{D,\text{out}}}
    + \frac{C}{h}(v\theta, u)_{\Gamma_D}
    + \left(v, (\vb{c}\cdot\vb{n})_+u\right)_{\Gamma_N} = (v, f)_\Omega
    + (v, \eta)_{\Gamma_N}
    + (\mathbb{L}v\tau, f)_{\Omega} \\
    & + \left(-\gamma\nabla v\cdot\vb{n} - \frac{1}{\theta}\vb{c}\cdot\vb{n}v, g\right)_{\Gamma_{D, \text{in}}}
    + (-\gamma\nabla v\cdot\vb{n}, g)_{\Gamma_{D,\text{out}}} + \frac{C}{h}(v, g)_{\Gamma_D}.
\end{align*}

Here, \(\mathcal{L}u = -\nabla\cdot(\theta\nabla u) + \vb{c}\cdot\nabla u\), \(\mathbb{L}v = \vb{c}\cdot\nabla v\), and \(\gamma\) is chosen as \(-1\). In particular, for both data NGOs and model NGOs the right-hand-side vector is computed as
\begin{align*}
    L[v] &= (v, f)_\Omega + (v, \eta)_{\Gamma_N} + \left(-\gamma\nabla v\cdot\vb{n} - \frac{1}{\theta}\vb{c}\cdot\vb{n}v, g\right)_{\Gamma_{D, \text{in}}} + (-\gamma\nabla v\cdot\vb{n}, g)_{\Gamma_{D,\text{out}}}.
\end{align*}

\section{Data generation}\label{A data generation}
\subsection{Steady Diffusion}
\label{sec:darcy-data-generation}
To test the efficacy of different model architectures, three different data sets are generated. Note that these are the data sets used to obtain the results presented in Section~\ref{S Generalization Errors of Different NOs} and elsewhere, unless otherwise specified.
\begin{itemize}
    \item For \textbf{dataset A}, \(N = 10000\) random triples \(\left( \theta, f, \eta \right)\) are sampled, and the resulting PDE is solved to obtain \(u\). The functions \(\theta\) are obtained by sampling \(10000\) independent functions \(\hat{\theta}\) from a Gaussian Random Field (GRF) with correlation length \(\lambda = 0.4\), after which \(\theta\) is defined as \(\theta_i(\vb{x}) = a\hat{\theta}(\vb{x}) + b\), where the scalars \(a, b\) are chosen such that \(\theta\) lies in the interval \([0.02, 0.99]\), i.e.~\(\min_i \min_{\vb{x}} \theta_i(\vb{x}) = 0.02\) and \(\max_i \max_{\vb{x}} \theta_i(\vb{x}) = 0.99\). The source terms \(f\) are generated using the same process, except with \(\lambda = 0.2\). Finally \(\eta\) is generated in the same way with \(\lambda = 0.3\) and the resulting functions are scaled and shifted to lie in \([-1, 1]\). The Dirichlet boundary data \(g\) are identically zero.

    To solve the PDEs, we make use of the Nutils~\cite{van_zwieten_2023_10068507} finite element solver. The domain is discretized on a \(20\times20\) grid of cubic B-splines (resulting in \(23^2\) basis functions). Comparing the solutions to those obtained on a \(40 \times 40\) grid, we find that the numerical solutions have a relative error of about \(3 \cdot 10^{-4}\).

    \item For \textbf{dataset B}, solutions are obtained using manufactured solutions \cite{hasani2024generating}. In this approach, the solution \(u\) and the parameters \(\theta\) are randomly generated, after which the source and boundary terms are computed by evaluating the relevant differential operator acting on \(u\). Firstly, the functions \(\theta\) are generated using the same method as is used in the previous data set, but with smaller characteristic length scale \(\lambda = 0.2\). Then, random functions \(\hat{u}\) are generated from a GRF with \(\lambda = 0.2\), and the solutions are given as \(u_i(\vb{x}) = x_1\left( 1 - x_1 \right)\hat{u}_i(\vb{x})\), so that the solutions satisfy homogeneous Dirichlet boundary conditions. Then, \(f\) and \(\eta\) are simply computed by evaluating their corresponding operators in \eqref{eq:darcy} through analytical expressions for the relevant derivatives. Finally, for each sample we scale \(f_i\), \(\eta_i\), and \(u_i\) uniformly so that \(\Vert f_i \Vert_{L^2} = \frac{1}{2}\). This normalization is done to ensure that the inputs \(\eta, \theta, f\) are of a similar magnitude to what they are in the training data. For the VarMiON and NGO, the normalization does not affect the model accuracy since these models are linear in \(f\) and \(\eta\). However, DeepONets and FNOs, which do not have this linearity, are expected to perform best when the inputs are of a similar magnitude as the training data.
\end{itemize}
The first data set will be used for training, validation, and testing, while the second data set will be used only for testing. The reason for this distinction is to test how well models trained on one data set can generalize to differently distributed data. When using models as preconditioners in numerical methods, the expectation is that the model will be used on problems that do not lie in the training data distribution. As such, generalization to such data is considered an important factor for models used as preconditioners. 
\paragraph{Dataset C} Dataset C consists of 10000 manufactured GRF solutions, defined as  
\begin{equation}\label{E MS}
u(\vb{x})=b+ c \cdot \mathrm{GRF}({\lambda};\vb{x}),   
\end{equation}
where GRF is a Gaussian random field with correlation length scale $\lambda$, scaling $c$ and offset $b$. We manufacture $\theta$ and $u$ by drawing the numbers $b$, $c$ and $\lambda$ from the distributions indicated in Table \ref{T Dataset C} and substituting them in Equation \ref{E MS}. The corresponding $f$, $\eta$ and $g$ were obtained by passing the manufactured $\theta$ and $u$ through the respective PDE and boundary operators of \ref{eq:darcy}.
\begin{table}[htb!]
\centering
\caption{Distributions used for the Gaussian random field offset $b$, scaling $c$ and length scale $\lambda/L$ for the solution $u$ and material parameter $\theta$ for dataset C, manufactured using Equation \ref{E MS}. $U$ is the uniform distribution.}
\begin{tabular}{|l|l|l|l|}
\hline
         & \textbf{Offset} $b$ & \textbf{Scaling} $c$ & \textbf{Length scale} $\lambda/L$ \\ \hline \textbf{Material parameter}
$\theta$ & 1            & $U(0,0.2)$   & $U(0.5,1)$         \\ \hline
\textbf{Solution} $u$      & $U(-1,1)$      & $U(0,1)$     & $U(0.5,1)$         \\ \hline
\end{tabular}
\label{T Dataset C}
\end{table}

\subsection{Time-Dependent Diffusion}\label{A data generation time-dependent diffusion}
\paragraph{Dataset D}
Dataset D consists of 10000 manufactured GRF solutions, defined as
\begin{equation}\label{E MS t}
u(\vb{x},t)=b+c\cdot  \mathrm{GRF}({\lambda},\tau;\vb{x},t),
\end{equation}
where $\tau$ is the timescale of the GRF, that we vary independently of the length scale $\lambda$. We manufacture $\theta$ and $u$ by drawing the numbers $b$, $c$, $\lambda$ and $\tau$ from the distributions indicated in Table \ref{T Dataset D} and substituting them in Equation \ref{E MS t}. The corresponding $u_0$, $f$, $\eta$ and $g$ were obtained by passing the manufactured $\theta$ and $u$ through the respective PDE and boundary operators of \ref{E PDE time-dependent diffusion}.
\begin{table}[htb!]
\centering
\caption{Distributions used for the Gaussian random field offset $b$, scaling $c$, length scale $\lambda/L$ and time scale $\tau/T$ for the solution $u$ and material parameter $\theta$ for dataset C, manufactured using Equation \ref{E MS t}. $U$ is the uniform distribution.}
\begin{tabular}{|l|l|l|l|l|}
\hline
         & \textbf{Offset} $b$ & \textbf{Scaling} $c$ & \textbf{Length scale} $\lambda/L$ & \textbf{Time scale} $\tau/T$ \\ \hline
\textbf{Material parameter} $\theta$ & 1            & $U(0,0.2)$   & $U(0.5,1)$         & $U(0.5,1)$      \\ \hline
\textbf{Solution} $u$      & $U(-1,1)$     & $U(0,1)$     & $U(0.5,1)$         & $U(0.5,1)$      \\ \hline
\end{tabular}
\label{T Dataset D}
\end{table}

\subsection{Advection-Diffusion}
\paragraph{Dataset E} Solutions to the advection-diffusion equation \eqref{eq:advection-diffusion} are obtained by drawing random parameters \(\theta, \vb{c}, f, g\), and \(\eta\), and subsequently solving for \(u\). The parameters are distributed as follows:
\begin{itemize}
    \item \(\vb{c}\) is chosen to be spatially constant, with \(c_x\) and \(c_y\) both distributed as \(\mathcal{N}(\mu=0, \sigma^2 = 25)\). This produces velocity fields with a uniformly distributed direction. In particular, some sample problems have a velocity field that points towards the left or right boundary, producing a boundary layer there due to the Dirichlet boundary conditions. Other velocity fields point towards the top or bottom boundaries, in which case no boundary layer appears.
    \item The distribution of \(\theta\) is chosen as \(\theta(\vb{x}) = \frac{1}{10}\log\left(e^{0.01} + \exp\left(\hat{\theta}(\vb{x})\right)\right)\) where \(\hat{\theta} \sim \text{GRF}(\lambda = 0.4)\). The reason for this distribution is to ensure positivity of \(\theta\) (in this case \(\theta > 0.001\)), while also allowing the ratio \(\frac{\max_{\vb{x}} \theta(\vb{x})}{\min_{\vb{x}} \theta(\vb{x})}\) to be high, i.e.~ensuring that \(\theta\) is not approximately constant.
    \item The other fields are chosen as \(f \sim \text{GRF}(\lambda = 0.4)\), \(g \sim 0.005\text{GRF}(\lambda = 0.4)\) and \(\eta \sim 0.5\text{GRF}(\lambda=0.4)\). For the boundary data \(g\) and \(\eta\), the left and right (resp.~top and bottom) segments are generated independently. The scaling factors for \(g\) and \(\eta\) are chosen such that all three terms of \(f, g, \eta\) contribute significantly to the solution.
\end{itemize}

Given these parameters, the PDEs are solved on a \(53 \times 53\) cubic B-spline basis (i.e.~with a mesh size of \(h = \frac{1}{50}\)), using the stabilized weak form of Bazilevs and Hughes~\cite{Bazilevs_Hughes_2005}.

\subsection{nonlinear Diffusion}\label{A data generation nonlinear diffusion}
\paragraph{Dataset F} Dataset F consists of 10000 manufactured GRF solutions, defined as in Equation \ref{E MS}. We manufacture $\theta_0$ and $u$ by drawing the numbers $b$, $c$ and $\lambda$ from the distributions indicated in Table \ref{T Dataset F} and substituting them in Equation \ref{E MS}. $\theta[u]$ is obtained by substitution of the manufactured $\theta_0$ and $u$ into Equation \ref{E nonlinear theta}, where $\alpha=0.1$. The corresponding $f$, $\eta$ and $g$ were obtained by passing the manufactured $\theta[u]$ and $u$ through the respective PDE and boundary operators of \ref{E nonlinear diffusion}.
\begin{table}[htb!]
\centering
\caption{Distributions used for the Gaussian random field offset $b$, scaling $c$ and length scale $\lambda/L$ for the solution $u$ and material parameter $\theta_0$ for dataset C, manufactured using Equation \ref{E MS}. $U$ is the uniform distribution.}
\begin{tabular}{|l|l|l|l|}
\hline
         & \textbf{Offset} $b$ & \textbf{Scaling} $c$ & \textbf{Length scale} $\lambda/L$ \\ \hline \textbf{Material parameter}
$\theta_0$ & 1            & $U(0,0.1)$   & $U(0.5,1)$         \\ \hline
\textbf{Solution} $u$      & $U(-1,1)$      & $U(0,1)$     & $U(0.5,1)$         \\ \hline
\end{tabular}
\label{T Dataset F}
\end{table}

\section{Model architectures}
\subsection{Steady Diffusion}
\subsubsection{Architectures of Section \ref{S Generalization Errors of Different NOs}}\label{A architectures Hugo}
\label{sec:appendix-model-architectures}
\begin{figure}[htb]
    \centering
    \includegraphics[scale=0.6]{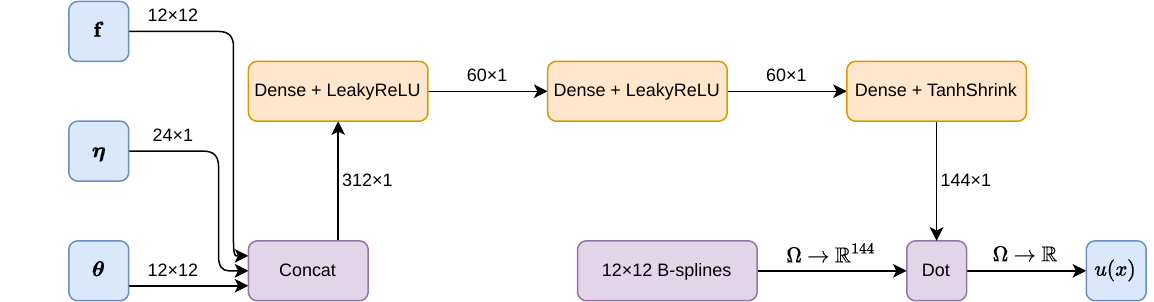} \\
    \vspace{2em}
    \includegraphics[scale=0.6]{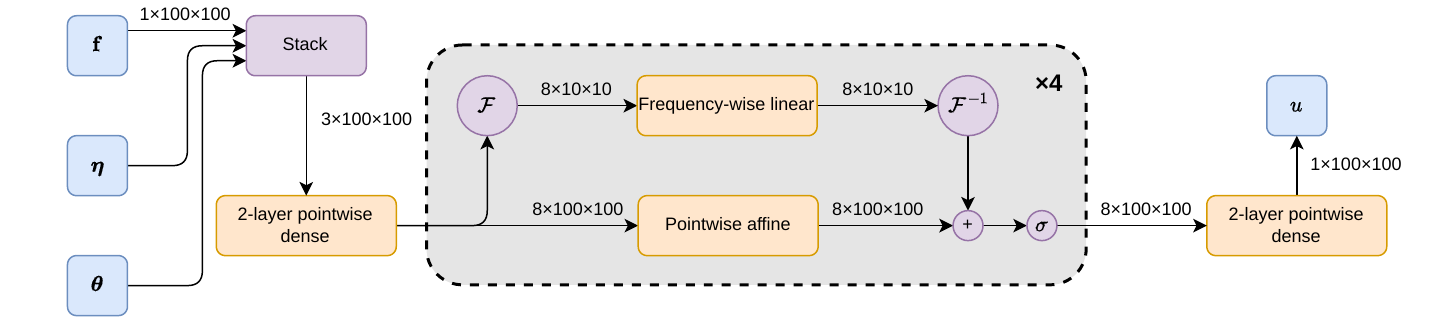} \\
    \vspace{2em}
    \includegraphics[scale=0.6]{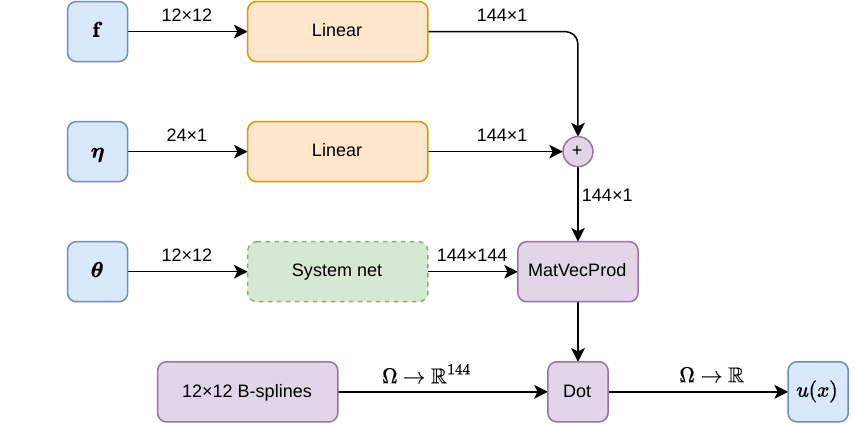}
    \caption{From top to bottom: architectures of the DeepONet/RINO, FNO, and VarMiON.}
    \label{fig:architectures-reference-models-1}
\end{figure}

\begin{figure}[htb]
    \centering
    \includegraphics[scale=0.6]{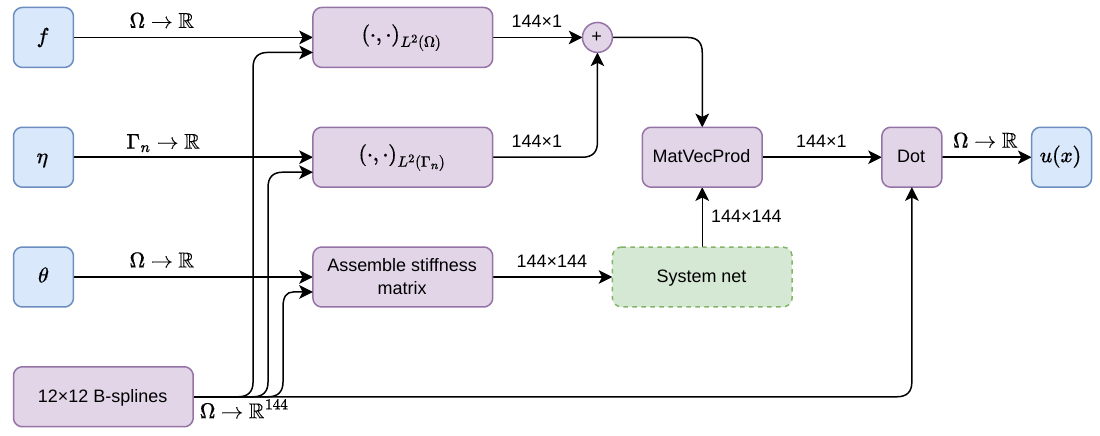} \\
    \vspace{2em}
    \includegraphics[scale=0.6]{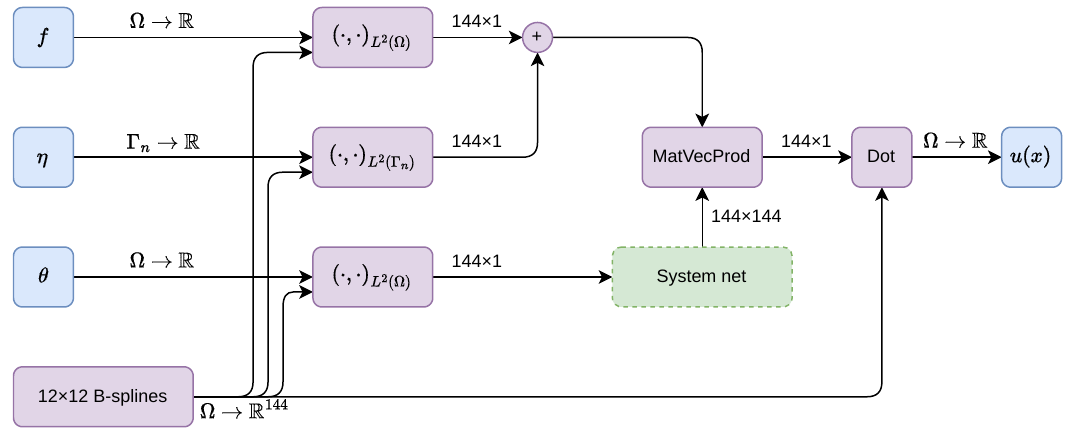} \\
    \caption{Architecture of the model NGO (top) and data NGO (bottom).}
    \label{fig:architectures-reference-models-2}
\end{figure}

\begin{figure}[htb]
    \centering
    \includegraphics[width=0.48\textwidth]{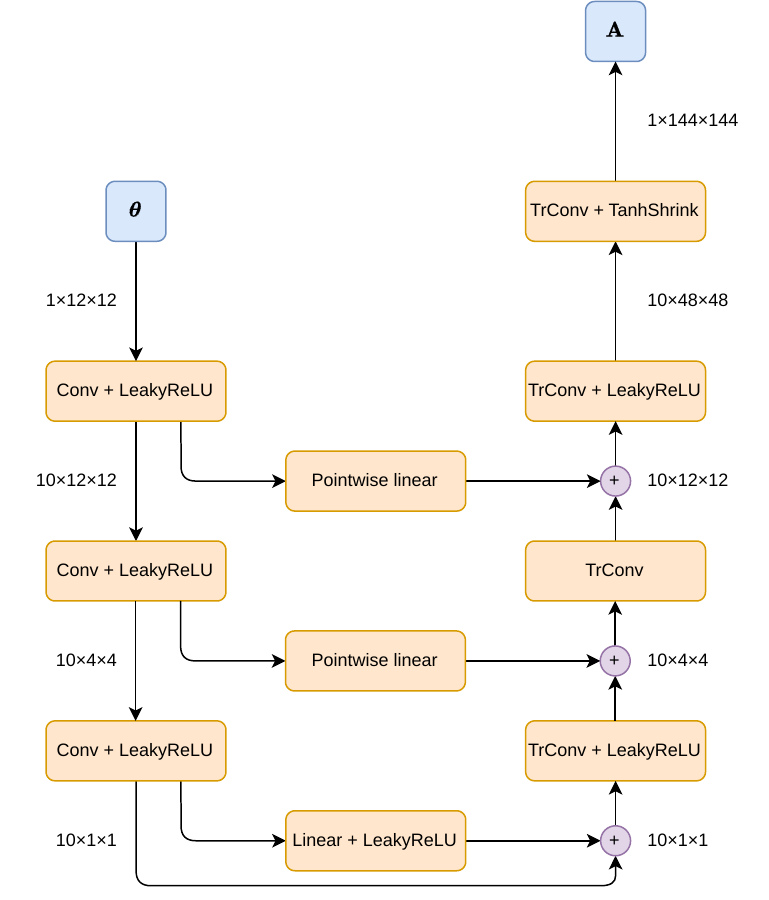}
    \hfill
    \includegraphics[width=0.48\textwidth]{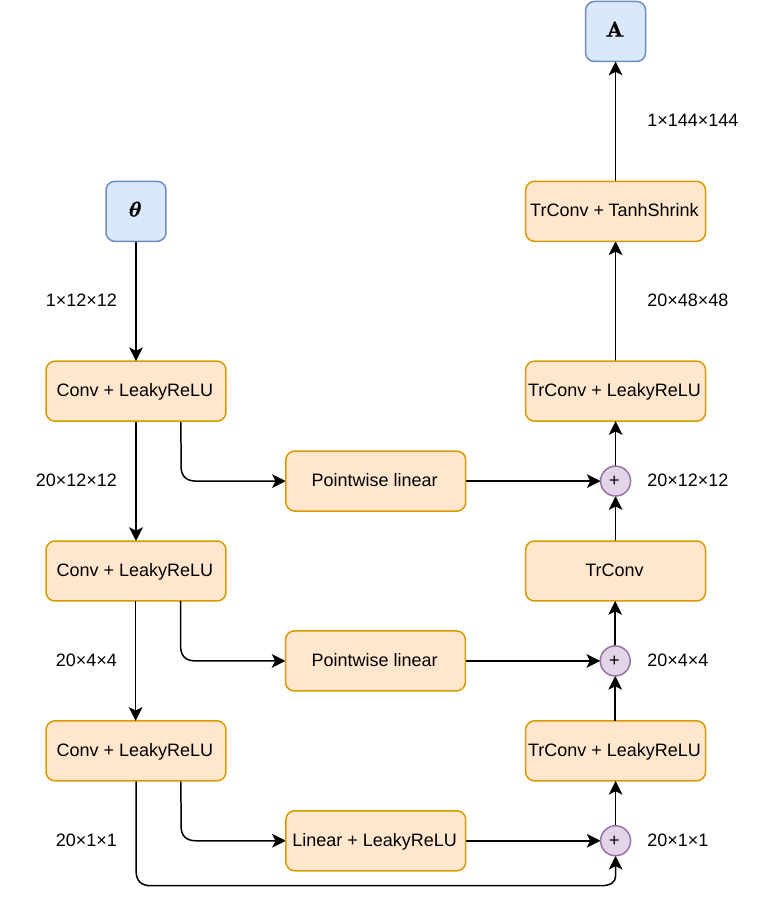} \\
    \vspace{2em}
    \includegraphics[width=0.48\textwidth]{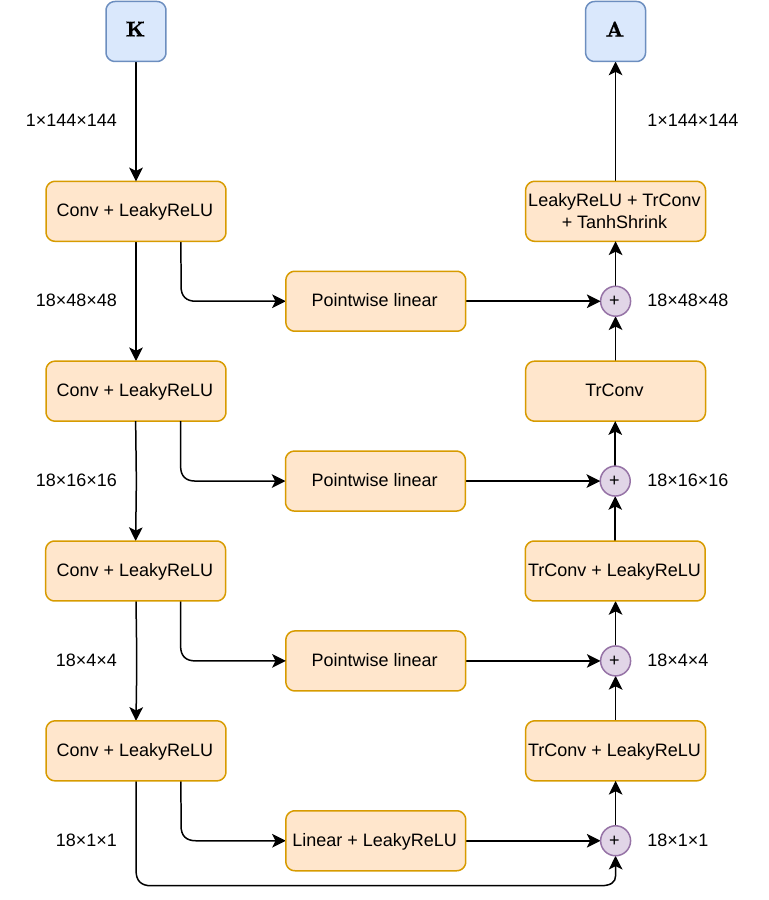}
    \caption{
        U-Net (system) architectures of the VarMiON (top left), data NGO (top right), and model NGO (bottom).
    }
    \label{fig:unets}
\end{figure}

Figure~\ref{fig:architectures-reference-models-1} shows the architectures of the DeepONet/RINO, FNO, and VarMiON. Figure~\ref{fig:architectures-reference-models-2} shows the architectures of the NGOs. The architectures of the system networks for the VarMiON and NGOs are shown in Figure~\ref{fig:unets}. The U-Net architectures for the data NGO and VarMiON are identical with exception of the number of channels in the hidden layers. This is  due to the extra \(144 \times 144\) linear layer in the VarMiON, which means that the U-Net must have fewer parameters in order to keep the total parameter count approximately the same.

\subsubsection{Architectures of Section \ref{S Generalization Errors Across Length Scales}}\label{A Model definition details of Section}
The architectures of the different models used in Section \ref{S Generalization Errors Across Length Scales} follow the structure of those presented in Section \ref{A architectures Hugo}, except that the learnable component in the DeepONet is a U-Net instead of an MLP, and that the models accept the Dirichlet boundary condition $g$ as an additional input function. Furthermore, the models use bases and quadrature grids of slightly different sizes, and therefore, the used NN architectures have different input and output shapes. The structure of the input and output shapes of the learnable components are clarified in Table \ref{T model definitions and shapes Joost}, together with the used mathematical model definitions. Table \ref{T NO settings sd Joost} contains the specific settings used regarding the basis, quadrature and learnable model component. For reproducibility, the specific system net settings used are included in Table \ref{T system net settings sd Joost}. The number of channels of the NN architectures have been used to equalize their trainable parameter counts to approximately 30000.

\begin{table}[htb!]
\centering
\caption{The mathematical definitions of the models used to produce the results in Section \ref{S Generalization Errors Across Length Scales}, together with the input and output shapes of the learnable components (or system nets) of these models. Hats indicate quantities learned by NNs. $\boldsymbol{\theta}:=\{\theta_i\}_{i=1}^{n_{\boldsymbol{\theta}}}$ and $\boldsymbol{f}:=\{f_i\}_{i=1}^{n_{\boldsymbol{f}}}$ denote the set of all $n_{\boldsymbol{\theta}}$ PDE coefficient fields and $n_{\boldsymbol{f}}$ inhomogeneity fields (forcings and boundary conditions), respectively. $Q_x$ and $Q_y$ denote the number of quadrature points $\vb{x}_q$ in directions $x$ and $y$, respectively. $N_x$ and $N_y$ denote the number of basis functions $\phi_m$ in directions $x$ and $y$, respectively, and $N=N_x N_y$ is the total number of basis degrees of freedom. $\vb{F}[\theta]$ is defined by Equation \ref{E F DNGO sd} for the data NGO and by Equation \ref{E F MNGO sd} for the data-free and model NGO, respectively. The right-hand side vector $d_n$ is defined by Equation \ref{E d Darcy}. A VarMiON learns $\hat{d}_n$ using a single layer MLP $\hat{F}_{i,nq}$ for every discretized inhomogeneity field $f_i(\mathbf{x}_{i,q})$.}
\begin{tabular}{|l|l|l|l|}
\hline
                       & \textbf{Definition}                                                                                                                                                                                                                                                                                         & \textbf{NN input shape}                                & \textbf{NN output shape} \\ \hline
\textbf{NN (U-Net)}    & $\hat{u}(\mathbf{x}_q) = \hat{u}\left(\boldsymbol{\theta}(\mathbf{x}_q),\boldsymbol{f}(\mathbf{x}_q)\right)$                                                                                                                                                                                                & $(n_{\boldsymbol{\theta}}+n_{\boldsymbol{f}},Q_x,Q_y)$ & $(Q_x,Q_y)$              \\ \hline
\textbf{DeepONet}      & $\hat{u}(\mathbf{x}) = \hat{u}_m\left(\boldsymbol{\theta}(\mathbf{x}_q),\boldsymbol{f}(\mathbf{x}_q)\right)\phi_m(\mathbf{x})$                                                                                                                                                                              & $(n_{\boldsymbol{\theta}}+n_{\boldsymbol{f}},Q_x,Q_y)$ & $(N_x,N_y)$              \\ \hline
\textbf{VarMiON}       & \begin{tabular}[c]{@{}l@{}}$\hat{u}(\mathbf{x}) = \hat{A}_{mn}\left(\boldsymbol{\theta}(\mathbf{x}_q)\right)\hat{d}_n\left(\boldsymbol{f}(\mathbf{x}_q)\right)\phi_m(\mathbf{x})$\\ where $\hat{d}_n \left( \boldsymbol{f} (\mathbf{x}_q) \right) = \sum_i \hat{F}_{i,nq} f_i(\mathbf{x}_{i,q})$\end{tabular} & $(n_{\boldsymbol{\theta}},Q_x,Q_y)$                    & $(N,N)$                  \\ \hline
\textbf{Data NGO}      & $\hat{u}(\mathbf{x}) = \hat{A}_{mn}\left(\mathbf{F}\left[\boldsymbol{\theta}\right]\right)d_n\left[\boldsymbol{f}\right]\phi_m(\mathbf{x})$                                                                                                                                                                 & $(n_{\boldsymbol{\theta}},N_x,N_y)$                    & $(N,N)$                  \\ \hline
\textbf{Data-free NGO} & $\hat{u}(\mathbf{x}) = \hat{A}_{mn}\left(\mathbf{F}\left[\boldsymbol{\theta}\right]\right)d_n\left[\boldsymbol{f}\right]\phi_m(\mathbf{x})$                                                                                                                                                                 & $(N,N)$                                                & $(N,N)$                  \\ \hline
\textbf{Model NGO}     & $\hat{u}(\mathbf{x}) = \hat{A}_{mn}\left(\mathbf{F}\left[\boldsymbol{\theta}\right]\right)d_n\left[\boldsymbol{f}\right]\phi_m(\mathbf{x})$                                                                                                                                                                 & $(N,N)$                                                & $(N,N)$                  \\ \hline
\textbf{FNO}           & $\hat{u}(\mathbf{x}_q) = \hat{u}\left(\boldsymbol{\theta}(\mathbf{x}_q),\boldsymbol{f}(\mathbf{x}_q)\right)$                                                                                                                                                                                                & $(n_{\boldsymbol{\theta}}+n_{\boldsymbol{f}},Q_x,Q_y)$   & $(Q_x,Q_y)$              \\ \hline
\end{tabular}
\label{T model definitions and shapes Joost}
\end{table}

\begin{table}[htb!]
\centering
\caption{Specific settings used for the different models of Section \ref{S Generalization Errors Across Length Scales} regarding basis, quadrature and system net. $N_x$, $N_y$, $Q_x$ and $Q_y$ are the number of basis functions and quadrature points, respectively, in directions $x$ and $y$, respectively. The U-Nets and FNO follow the structure of Figures \ref{fig:architectures-reference-models-1} and \ref{fig:unets}.}
\begin{tabular}{|l|ll|ll|ll|}
\hline
                       & \textbf{Basis}                     & \textbf{}            & \textbf{Quadrature}                & \textbf{}            & \textbf{System Net}                &                   \\ \cline{2-7} 
                       & \multicolumn{1}{l|}{\textbf{Type}} & \textbf{$(N_x,N_y)$} & \multicolumn{1}{l|}{\textbf{Type}} & \textbf{$(Q_x,Q_y)$} & \multicolumn{1}{l|}{\textbf{Type}} & \textbf{\#params} \\ \hline
\textbf{NN (U-Net)}    & \multicolumn{1}{l|}{-}             & -                    & \multicolumn{1}{l|}{Uniform}       & $(100,100)$          & \multicolumn{1}{l|}{U-Net}         & 28770             \\ \hline
\textbf{DeepONet}      & \multicolumn{1}{l|}{B-splines}     & $(10,10)$            & \multicolumn{1}{l|}{Uniform}       & $(100,100)$          & \multicolumn{1}{l|}{U-Net}         & 28878             \\ \hline
\textbf{VarMiON}       & \multicolumn{1}{l|}{B-splines}     & $(10,10)$            & \multicolumn{1}{l|}{Uniform}       & $(10,10)$            & \multicolumn{1}{l|}{U-Net}         & 29072             \\ \hline
\textbf{Data NGO}      & \multicolumn{1}{l|}{B-splines}     & $(10,10)$            & \multicolumn{1}{l|}{Gaussian}      & $(99,99)$            & \multicolumn{1}{l|}{U-Net}         & 29482             \\ \hline
\textbf{Data-free NGO} & \multicolumn{1}{l|}{B-splines}     & $(10,10)$            & \multicolumn{1}{l|}{Gaussian}      & $(99,99)$            & \multicolumn{1}{l|}{U-Net}         & 28590             \\ \hline
\textbf{Model NGO}     & \multicolumn{1}{l|}{B-splines}     & $(10,10)$            & \multicolumn{1}{l|}{Gaussian}      & $(99,99)$            & \multicolumn{1}{l|}{U-Net}         & 28590             \\ \hline
\textbf{FNO}           & \multicolumn{1}{l|}{Fourier modes} & $(10,10)$            & \multicolumn{1}{l|}{Uniform}       & $(100,100)$          & \multicolumn{1}{l|}{FNO}           & 29128             \\ \hline
\end{tabular}
\label{T NO settings sd Joost}
\end{table}

\begin{table}[htb!]
\centering
\caption{Specific settings of the system nets used by the different models of Section \ref{S Generalization Errors Across Length Scales}.}
\begin{tabular}{|l|l|l|l|l|}
\hline
                       & \textbf{Channels} & \textbf{Kernel sizes (= strides)}                   \\ \hline
\textbf{NN (U-Net)}               & 15                & (2,2), (2,2), (5,5), (5,5), (5,5), (5,5), (2,2), (2,2)   \\ \hline
\textbf{DeepONet}      & 18                & (10,10), (1,1), (2,2), (5,5), (5,5), (2,2), (1,1), (1,1) \\ \hline
\textbf{VarMiON} & 12                & (1,1),(1,1), (2,2), (5,5), (5,5), (2,2), (1,1), (10,10) \\ \hline
\textbf{Data NGO} & 22                & (1,1), (1,1), (2,2), (5,5), (5,5), (2,2), (1,1), (10,10) \\ \hline
\textbf{Data-free NGO} & 15                & (2,2), (2,2), (5,5), (5,5), (5,5), (5,5), (2,2), (2,2)   \\ \hline
\textbf{Model NGO}     & 15                & (2,2), (2,2), (5,5), (5,5), (5,5), (5,5), (2,2), (2,2)   \\ \hline
\textbf{FNO}           & 7                 & -                                                 \\ \hline
\end{tabular}
\label{T system net settings sd Joost}
\end{table}

\subsubsection{System net architectures of Section \ref{S Comparison of Different Bases and System Network Architectures}}\label{A system net architectures Joost}
The four system nets used in Figure \ref{F systemnets} are defined as follows:
\begin{itemize}
    \item MLP (multilayer perceptron): a 4 layer multilayer perceptron with ReLU activations and zero biases. The input and output shapes have been properly adjusted to match the data, and the (equal) width of the hidden layers has been used to adjust the number of trainable parameters to approximately $3 \cdot 10^4$.
    \item CNN (convolutional neural network): the CNN follows the structure of the bottom model of Figure \ref{fig:unets}, without skip connections. The kernel, input and output shapes are mentioned in Tables \ref{T model definitions and shapes Joost}, \ref{T NO settings sd Joost} and \ref{T system net settings sd Joost}.
    \item U-Net: the U-Net follows the structure of the bottom model of Figure \ref{fig:unets}. The kernel, input and output shapes are mentioned in Tables \ref{T model definitions and shapes Joost}, \ref{T NO settings sd Joost} and \ref{T system net settings sd Joost}.
    \item FNO (fourier neural operator): the FNO follows the structure shown in Figure \ref{fig:architectures-reference-models-1}. The kernel, input and output shapes are mentioned in Tables \ref{T model definitions and shapes Joost}, \ref{T NO settings sd Joost} and \ref{T system net settings sd Joost}.
\end{itemize}

\subsection{Time-Dependent Diffusion}\label{A Model architectures time-dependent diffusion}
The model architectures of Section \ref{S time dependent diffusion} follow the definitions presented in Table \ref{T model definitions and shapes Joost}, except that the bases $\phi_m$ now depend on space-time $(\vb{x},t)$ and that the quadrature is now defined on the space-time domain shown in Figure $\ref{F domain_dynamic_diffusion}$. The settings of the different neural operators regarding the basis, system net, quadrature and number of trainable parameters are presented in Table \ref{T NO settings tdd Joost}. For reproducibility, the specific system net settings used are included in Table \ref{T system net settings tdd Joost}.
\begin{table}[htb]
\centering
\caption{Specific settings used for the different models of Section \ref{S time dependent diffusion} regarding basis, quadrature and system net. $N_t$, $N_x$, $N_y$, $Q_t$, $Q_x$ and $Q_y$ are the number of basis functions and quadrature points, respectively, in directions $t$, $x$ and $y$, respectively. The U-Nets and FNO follow the structure of Figures \ref{fig:architectures-reference-models-1} and \ref{fig:unets}.}
\begin{tabular}{|l|ll|ll|ll|}
\hline
                       & \textbf{Basis}                     & \textbf{}                & \textbf{Quadrature}                & \textbf{}                & \textbf{System Net}                &                   \\ \cline{2-7} 
                       & \multicolumn{1}{l|}{\textbf{Type}} & \textbf{$(N_t,N_x,N_y)$} & \multicolumn{1}{l|}{\textbf{Type}} & \textbf{$(Q_t,Q_x,Q_y)$} & \multicolumn{1}{l|}{\textbf{Type}} & \textbf{\#params} \\ \hline
\textbf{NN (U-Net)}    & \multicolumn{1}{l|}{-}             & -                        & \multicolumn{1}{l|}{Uniform}       & $(2,100,100)$            & \multicolumn{1}{l|}{U-Net}         & 29702             \\ \hline
\textbf{DeepONet}      & \multicolumn{1}{l|}{B-splines}     & $(2,10,10)$              & \multicolumn{1}{l|}{Uniform}       & $(2,100,100)$            & \multicolumn{1}{l|}{U-Net}         & 29141             \\ \hline
\textbf{VarMiON}       & \multicolumn{1}{l|}{B-splines}     & $(2,10,10)$              & \multicolumn{1}{l|}{Uniform}       & $(2,5,5)$                & \multicolumn{1}{l|}{U-Net}         & 29638             \\ \hline
\textbf{Data NGO}      & \multicolumn{1}{l|}{B-splines}     & $(2,10,10)$              & \multicolumn{1}{l|}{Gaussian}      & $(2,99,99)$              & \multicolumn{1}{l|}{U-Net}         & 29008             \\ \hline
\textbf{Data-free NGO} & \multicolumn{1}{l|}{B-splines}     & $(2,10,10)$              & \multicolumn{1}{l|}{Gaussian}      & $(2,99,99)$              & \multicolumn{1}{l|}{U-Net}         & 26556             \\ \hline
\textbf{Model NGO}     & \multicolumn{1}{l|}{B-splines}     & $(2,10,10)$              & \multicolumn{1}{l|}{Gaussian}      & $(2,99,99)$              & \multicolumn{1}{l|}{U-Net}         & 26556             \\ \hline
\textbf{FNO}           & \multicolumn{1}{l|}{Fourier modes} & $(2,10,10)$              & \multicolumn{1}{l|}{Uniform}       & $(2,100,100)$            & \multicolumn{1}{l|}{FNO}           & 28734             \\ \hline
\end{tabular}
\label{T NO settings tdd Joost}
\end{table}
\begin{table}[htb]
\centering
\caption{Specific settings of the system nets used by the different models of Section \ref{S time dependent diffusion}.}
\begin{tabular}{|l|l|l|l|l|}
\hline
                       & \textbf{Channels} & \textbf{Kernel sizes (= strides)}                                                                              \\ \hline
\textbf{NN (U-Net)}    & 11                &  (1,2,2), (1,2,2), (1,5,5), (2,5,5), (2,5,5), (1,5,5), (1,2,2), (1,2,2) \\ \hline
\textbf{DeepONet} & 11                & (1,10,10), (1,1,1), (1,2,2), (2,5,5), (2,5,5), (1,2,2), (1,1,1), (1,1,1) \\ \hline
\textbf{VarMiON}       & 2                 & (1,1,1), (1,1,1), (1,1,1), (2,5,5), (5,5), (1,1), (1,1), (40,40) \\ \hline
\textbf{Data NGO}              & 14                & (1,1,1), (1,1,1), (1,2,2), (2,5,5), (5,5), (2,2), (1,1), (20,20)  \\ \hline
\textbf{Data-free NGO} & 6                 & (2,2), (2,2), (5,5), (10,10), (10,10), (5,5), (2,2), (2,2) \\ \hline
\textbf{Model NGO}     & 6                 & (2,2), (2,2), (5,5), (10,10), (10,10), (5,5), (2,2), (2,2)             \\ \hline
\textbf{FNO} & 7                 & -                                                                                                            \\ \hline
\end{tabular}
\label{T system net settings tdd Joost}
\end{table}

For the time-dependent diffusion problem, the architecture of the data NGO is slightly different than for the steady diffusion problem: for the time-dependent diffusion problem, we provide the data NGOs the additional fixed input $F_{t,n}\equiv \int_\Omega \psi_n\left( (i+1)\Delta t \right) d\vb{x}'$. The reason why we do this is that otherwise the system net that we use, a U-Net with ReLU activations and zero bias, exhibits an unwanted symmetry that is not present in the training data. A ReLU activation function exhibits the scale equivariance property
\begin{equation}
    \mathrm{ReLU}(c\vb{x}) = c \mathrm{ReLU}(\vb{x}) \;\;\;\;\; \forall c \in \mathbb{R}^+,
\end{equation}
and since a U-Net with ReLU activations and zero bias consists only of linear maps and ReLU activations, the U-Net as a whole exhibits the same scale equivariance property
\begin{equation}
    \mathrm{NN}(c\vb{x}) = c \mathrm{NN}(\vb{x}) \;\;\;\;\; \forall c \in \mathbb{R}^+.
\end{equation}
The mapping that the system net intends to learn is not scale equivariant in this sense, so we do not want the system net to have this property either. Adding a constant input to $F_{t,n}\equiv \int_\Omega \psi_n\left( (i+1)\Delta t \right) d\vb{x}'$ breaks this unwanted symmetry. $\vb{F}_t$ enters the U-Net via a separate input channel. Adding $\vb{F}_t$ as input did not significantly change the accuracy of the data NGO on the tests performed in Section \ref{S time dependent diffusion}, however, it did result in slightly faster convergence of the training.

\subsection{Advection Diffusion}
\label{S model architectures advection-diffusion}
The architectures of the NGOs are largely unchanged from their architectures in Section~\ref{S Generalization Errors of Different NOs}. The differences are listed here:
\begin{itemize}
    \item Model NGOs construct the input matrix using the weak form from Bazilevs and Hughes~\cite{Bazilevs_Hughes_2005}, modified to handle the different boundary conditions (the cited paper assumes Dirichlet boundary conditions on the entire boundary). However, model NGOs omit the SUPG term and Dirichlet penalty term (the third and last terms of equation (10) of Bazilevs and Hughes). These were found not to have a positive effect on the model accuracy, and as such they are omitted to improve the computational efficiency of the model.
    \item Data NGOs construct the input matrix by integrating \(\theta\), \(c_x\), and \(c_y\) against the B-spline basis. The output matrix is multiplied by the same right-hand-side vector as is used in model NGOs. In particular, this means that the right-hand-side vector is also assembled using the unstabilized part of the weak form of Bazilevs and Hughes~\cite{Bazilevs_Hughes_2005}.
    \item For DeepONets and VarMiONs, the inputs are \(\theta, c_x, c_y, f, g\), and \(\eta\). The fields \(\theta, c_x, c_y\), and \(f\) are sampled on a \(12\times12\) uniform grid, whereas the boundary data \(g\) and \(\eta\) are sampled on \(2\times12\) points each, namely on 12 points per boundary segment. Note that this means that DeepONets and VarMiONs are given much coarser information about the inputs than other models. As explained, however, the architectures of these models do not allow for the inputs to be sampled more finely without drastically increasing the model's parameter count.
    \item For FNOs, CNOs, and U-Nets, the inputs are \(\theta, c_x, c_y, f, g\), and \(\eta\), all sampled on a \(100 \times 100\) uniform grid. Since \(\eta\) and \(g\) are only defined on the boundaries, their values are extended to the entire domain via nearest-neighbor resampling.
\end{itemize}

\section{Training Procedure}
\subsection{Steady Diffusion}
\label{sec:appendix-training-procedure-darcy}
\subsubsection{Training procedure for Tests in Sections \ref{S Generalization Errors of Different NOs}, \ref{S Train Speed} and \ref{S Precon}}
\label{A training procedure steady diffusion Hugo}
The training procedure is the same for each model. The model is trained on the first 8000 samples of the first dataset. After every epoch, the model is evaluated on the validation data (another 1000 samples from the first data). The final trained model is the model that performs best on the validation data. Table~\ref{tab:training_parameters} shows the training configuration.
\begin{table}[htb]
    \centering
    \caption{Configuration of the training procedure}
    \begin{tabular}{r | l}
        \toprule
        Loss function & Relative \(L^2\) error \\
        Optimizer & ADAM \\
        Learning rate & Default (\(10^{-3}\)) \\
        Batch size & 100 \\
        Training time & 20000 epochs \\
        Training data & Samples \(1-8000\) \\
        Validation data & Samples \(8001-9000\) \\
        Testing data (in distribution) & Samples \(9001-10000\) \\
        Testing data (out of distribution) & All 1000 manufactured samples \\
        \bottomrule
    \end{tabular}
    \label{tab:training_parameters}
\end{table}

\subsubsection{Training procedure for Tests in Sections \ref{S Generalization Errors Across Length Scales}, \ref{S Comparison of Different Bases and System Network Architectures} and \ref{S Sensitivity Analysis}}\label{A training procedure steady diffusion Joost}
The training settings of the models corresponding to the results of Sections \ref{S Generalization Errors Across Length Scales}, \ref{S Comparison of Different Bases and System Network Architectures} and \ref{S Sensitivity Analysis} are identical to those presented in Table \ref{tab:training_parameters}, except that 10000 samples have been used for training and 1000 samples for validation, and that the data-free NGO and model NGO have been trained for 5000 epochs only.

\subsection{Time-Dependent Diffusion}\label{A Training Procedure Time-Dependent Diffusion}
The training settings used for the time-dependent diffusion problem are identical to those of the steady diffusion problem as presented in Table \ref{tab:training_parameters}, except that 10000 samples have been used for training and 1000 samples for validation, and that the data-free NGO and model NGO have been trained for 5000 epochs only. The inductive bias for time stepping stability as presented in Section \ref{S Inductive Bias: Time Stepping Stability} is all included in the forward and backward passes of the training loop. To estimate the norm of the update matrix $||\hat{\vb{A}}^{(i)}\vb{M}_{\mathrm{lr}}||_{\vb{M}_{\mathrm{rr}}}$in the norm scaling layer in the data NGO, we use a power iteration \cite{golub2013matrix}
\begin{equation}
    \lambda, \vb{v} \approx \mathrm{PowerIteration\left(\vb{B};\lambda_0, \vb{v}_0 \right)},
\end{equation}
where $\vb{B}$ is a square matrix, $\lambda_0$ and $\vb{v}_0$ are initial guesses for the dominant eigenvalue and eigenvector of $\vb{B}$, respectively, and $\lambda$ and $\vb{v}$ are power iteration approximations of the dominant eigenvalue and eigenvector of $\vb{B}$, respectively. The initial guess for the dominant eigenvalue $\lambda_0$ is set to 1, and kept fixed. To speed up the convergence of the power iterations, we update the initial guess $\vb{v}_0$ during training after every epoch as
\begin{equation}
    \lambda_0, \vb{v}_0 \leftarrow \mathrm{PowerIteration\left(\hat{\vb{A}}[\expval{\theta}]\vb{M}_{\mathrm{lr}};\lambda_0, \vb{v}_0 \right)}.
\end{equation}
In words, we "learn" during training the dominant eigenvector corresponding to the matrix $\hat{\vb{A}}[\expval{\theta}]\vb{M}_{\mathrm{lr}}$, and use this as initial guess $\vb{v}_0$ for other matrices $\hat{\vb{A}}^{(i)}\vb{M}_{\mathrm{lr}}$. Since all material parameters $\theta$ are centered and reasonably close to $\expval{\theta}=1$, the dominant eigenvector $\vb{v}_0$ of the matrix $\hat{\vb{A}}[\expval{\theta}]\vb{M}_{\mathrm{lr}}$ is a good initial guess for the dominant eigenvector $\vb{v}$ of the matrices $\hat{\vb{A}}^{(i)}\vb{M}_{\mathrm{lr}}$. A power iteration then typically takes 2-10 iterations to converge.

\subsection{Advection-Diffusion}
The training procedure for advection-diffusion is the same as that for steady diffusion (Appendix~\ref{sec:appendix-training-procedure-darcy}).

\section{Scale Equivariance}\label{A Scale equivariance tests}
\begin{figure}[htb]
\centering
\includegraphics[width = 0.8\textwidth]{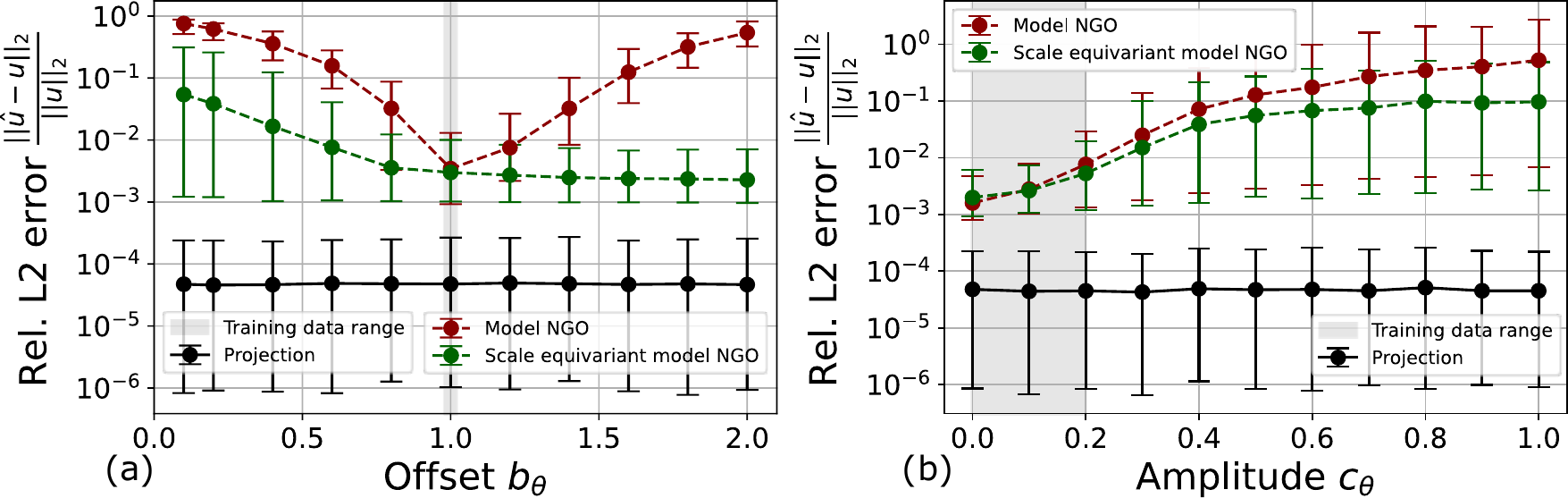}
\caption{\small{Relative test \(L^2\) error of model NGOs with and without embedded scale equivariance, versus the (a) offset $b_\theta$ and (b) amplitude $c_\theta$ of the material parameter $\theta(\vb{x})=b_\theta +c_\theta \mathrm{GRF}_{\lambda_\theta}(\vb{x})$. The model architectures are presented in \ref{A Model definition details of Section}, and the models have been trained on dataset C (Appendix \ref{A data generation}). The scale equivariance makes the NGO more robust against scale variations in the material parameter $\theta(\vb{x})$. Points and error bars are, respectively, averages and 95\% confidence intervals on 1000 manufactured steady diffusion solutions, generated in the same way as dataset C (Appendix \ref{A data generation}).}}
\label{F scaleequivariance}
\end{figure}

In this section, we show how to embed into the NGO formulation the scale equivariance of a PDE with respect to its material parameter, in the sense of Equation \ref{PDE scale equivariance}. If a PDE is scale equivariant in the sense of Equation \ref{PDE scale equivariance}, the Green's function exhibits a similar property. By substituting $\theta'=c\theta$ into \eqref{E definition G bvp}, we get
\begin{equation}\label{E SE step1}
    \mathcal{L}^*[\theta']G[\theta']=\mathcal{L}^*[c\theta]G[c\theta]= c\mathcal{L}^*[\theta]G[c\theta]= \delta, \;\;\;\;\; \forall c \in \mathbb{R}^+.
\end{equation}
Combining Equations \eqref{E SE step1} and \eqref{E definition G bvp} gives
\begin{equation}
    G[c\theta] = \frac{1}{c} G[\theta], \;\;\;\;\; \forall c \in \mathbb{R}^+.
\end{equation}
Equation \eqref{E Galerkin approximation Green's function} tells us that we can embed this scale equivariance property into the neural Green's function $\hat{G}$ by making sure that
\begin{equation}\label{E SE A}
    \hat{A}_{mn}[c\theta] = \frac{1}{c}\hat{A}_{mn}[\theta],  \;\;\;\;\; \forall c \in \mathbb{R}^+.
\end{equation}
If we redefine our neural network $\hat{A}_{mn}$ as
\begin{equation}\label{E SE NN}
    \hat{A}_{mn}'(\vb{F}[\theta]) \equiv \frac{1}{\norm{\theta}} \hat{A}_{mn}\left( \frac{\vb{F}[\theta]}{\norm{\theta}} \right),
\end{equation}
where $\norm{.}$ is a scale equivariant norm, $\hat{A}_{mn}$ satisfies the scale equivariance property \eqref{E SE A}, since
\begin{equation}
\begin{aligned}
    \hat{A}_{mn}'(\vb{F}[c\theta]) &= \frac{1}{\norm{c\theta}} \hat{A}_{mn}\left( \frac{\vb{F}[c\theta]}{\norm{c\theta}} \right) = \frac{1}{c\norm{\theta}} \hat{A}_{mn}\left(\frac{c\vb{F}[\theta]}{c\norm{\theta}} \right) = \frac{1}{c} \frac{1}{\norm{\theta}} \hat{A}_{mn}\left(\frac{\vb{F}[\theta]}{\norm{\theta}} \right) \\
    &= \frac{1}{c}\hat{A}_{mn}'(\vb{F}[\theta]).
\end{aligned}
\end{equation}
In Figure \ref{F scaleequivariance}, we show the effect if embedding scale equivariance into a preconditioned model NGO. 
Details of the model architectures are highlighted in Appendix \ref{A Model definition details of Section}. The models have been trained on dataset C, described in Appendix \ref{sec:darcy-data-generation}, using the training procedure described in Appendix \ref{A training procedure steady diffusion Joost}. It can be observed in Figure \ref{F scaleequivariance} that embedding scale equivariance into an NGO results in more robustness in scale variations in a material parameter $\theta$ manufactured according to \eqref{E MS}. The reason why the error still increases for small $b_\theta$ is that the output of a scale equivariant system net as given by \eqref{E SE NN} is divided by an increasingly small number $\norm{\theta}$, which amplifies the prediction error of $\hat{A}_{mn}$.

\section{Kronecker Product Assembly of the System Matrix}\label{A Kronecker product assembly of the system matrix}
We first project the material parameters onto the basis given by \eqref{E Kronecker product basis} as
\begin{equation}
    \theta(\vb{x},t) \approx \boldsymbol{\theta}^\mathrm{T} \boldsymbol{\psi}(\vb{x},t) = \boldsymbol{\theta}^\mathrm{T} (\boldsymbol{\psi}^{(t)}(t) \otimes_\mathrm{K} \boldsymbol{\psi}^{(x)}(x) \otimes_\mathrm{K} \boldsymbol{\psi}^{(y)}(y)),
\end{equation}
where $\boldsymbol{\theta}\equiv \{ \hat{\theta}_l\}_{l=1}^{N}$. This enables us to factorize the 3D integrals in the system matrix into three 1D integrals as 
\begin{multline}\label{E tensorized F}
    F_{ik}[\theta] \sim \int_{\Omega\times \Delta t_i} \phi_i \mathcal{L}^*[\theta] \psi_k \, d\vb{x}'dt' 
    \approx 
    \sum_{j=1}^N \hat{\theta}_j \int_{\Omega\times \Delta t_i} \phi_i \mathcal{L}^*[\psi_j] \psi_k \, d\vb{x}'dt'\\
    = \sum_{j=1}^N \hat{\theta}_j (\vb{F}^{(t)} \otimes_\mathrm{K} \vb{F}^{(x)} \otimes_\mathrm{K} \vb{F}^{(y)})_{ijk} \quad \forall i,k\in \{1,2,...,N\},
\end{multline}
where
\begin{equation}
\begin{aligned}
    \vb{F}^{(t)} &= \int_0^T \boldsymbol{\phi}^{(t)} \otimes \mathcal{L}^*[\boldsymbol{\psi}^{(t)}] \otimes \boldsymbol{\psi}^{(t)} dt', \\
    \vb{F}^{(x)} &= \int_0^L \boldsymbol{\phi}^{(x)} \otimes \mathcal{L}^*[\boldsymbol{\psi}^{(x)}] \otimes \boldsymbol{\psi}^{(x)} dx', \\
    \vb{F}^{(y)} &= \int_0^L \boldsymbol{\phi}^{(y)} \otimes \mathcal{L}^*[\boldsymbol{\psi}^{(y)}] \otimes \boldsymbol{\psi}^{(y)} dy',
\end{aligned} 
\end{equation}
and where $\otimes$ is the outer product. We compute $\vb{F}^{(t)}$, $\vb{F}^{(x)}$ and $\vb{F}^{(y)}$ once offline, and save them in memory, and assemble $F_{ik}[\theta]$ using \eqref{E tensorized F}.

\section{Relation to CNOs}
\label{appendix:relation-to-cnos}
A Neural Operator architecture similar to NGOs is the Convolutional Neural Operator (CNO) \cite{raonic2024convolutional}. In a CNO, the input function(s), defined on a rectangular domain in \(\mathbb{R}^d\), are passed through a low-pass filter, creating band-limited functions that can be reconstructed exactly from their samples on a relatively coarse grid. The band-limited function is then sampled on this grid, providing the inputs to a U-Net. The U-Net outputs are then treated as samples of band-limited functions from which the output functions are interpolated. The interpolation also assumes that the samples are of band-limited functions, so that there is a unique band-limited interpolated function.

Both the filtering and sampling of the inputs and the interpolation of the outputs are done through FFTs. In particular, both processes are linear meaning that applying them to standard basis vectors reveals an effective function basis that CNOs operate on. In 1D, the output interpolation can be seen as an expansion in terms of a function basis, where the basis functions are the Dirichlet Kernels \(D_n(x - x_k) = 1 + 2\sum_{i=1}^{\lfloor n / 2 \rfloor}\cos (2\pi i(x - x_k)) = \frac{\sin(n\pi (x - x_k))}{\sin(\pi (x - x_k))}\) with \(x_k = \frac{k}{n}L\). In 2D, the basis functions are the tensor products \(\phi_{kl}(x, y) = D_{n_1}(x - x_k)D_{n_2}(y - y_l)\). The input filtering, being up to a multiplicative constant the adjoint of the output interpolation, is equivalent to integration of the input functions against these basis functions with a uniform quadrature rule.

\begin{figure}[htb]
    \centering
    \includegraphics[width=0.80\linewidth]{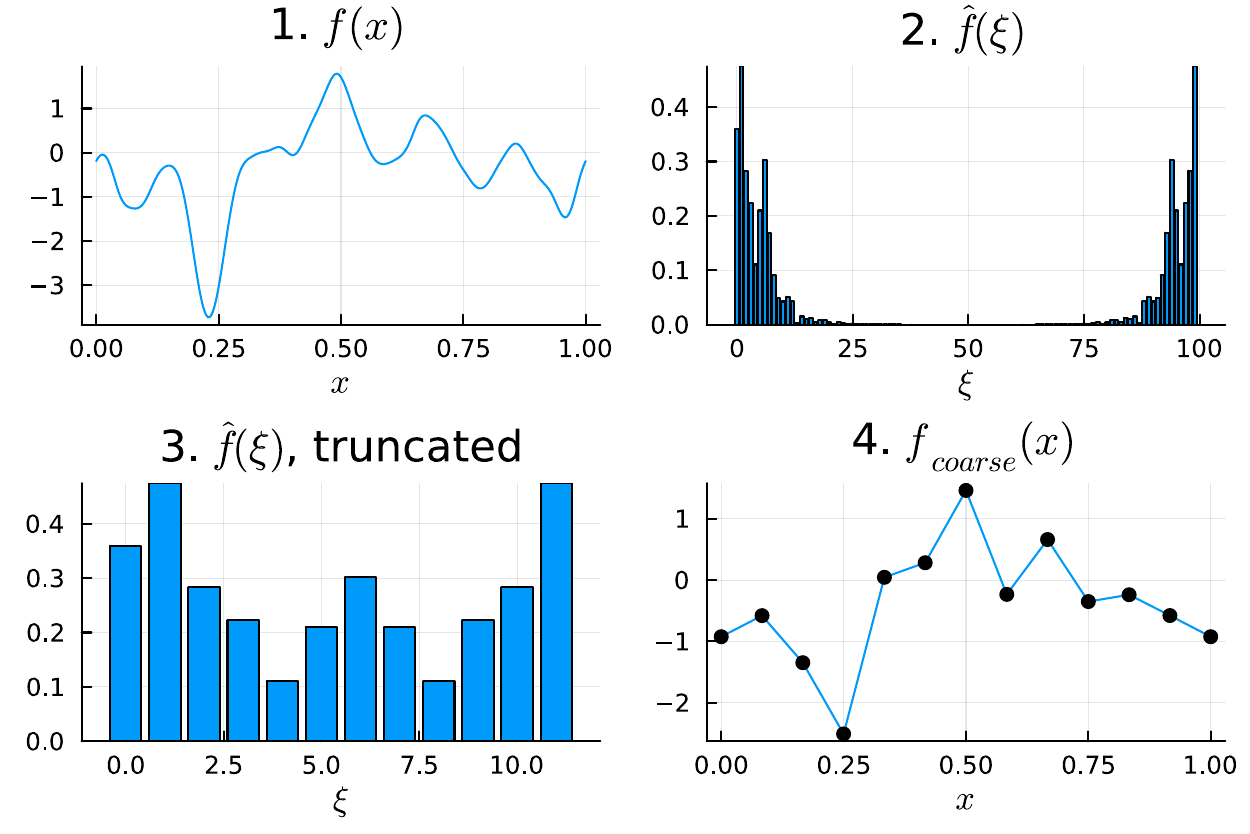}
    \caption{An overview of the filtering process as done in CNOs. A function \(f\) on a fine grid (1) is Fourier transformed (2). Then, only the lowest Fourier modes are retained (3). Finally, performing an inverse FFT produces the filtered function on a coarse grid (4).}
    \label{fig:cno-encode}
\end{figure}

\begin{figure}[htb]
    \centering
    \includegraphics[width=0.80\linewidth]{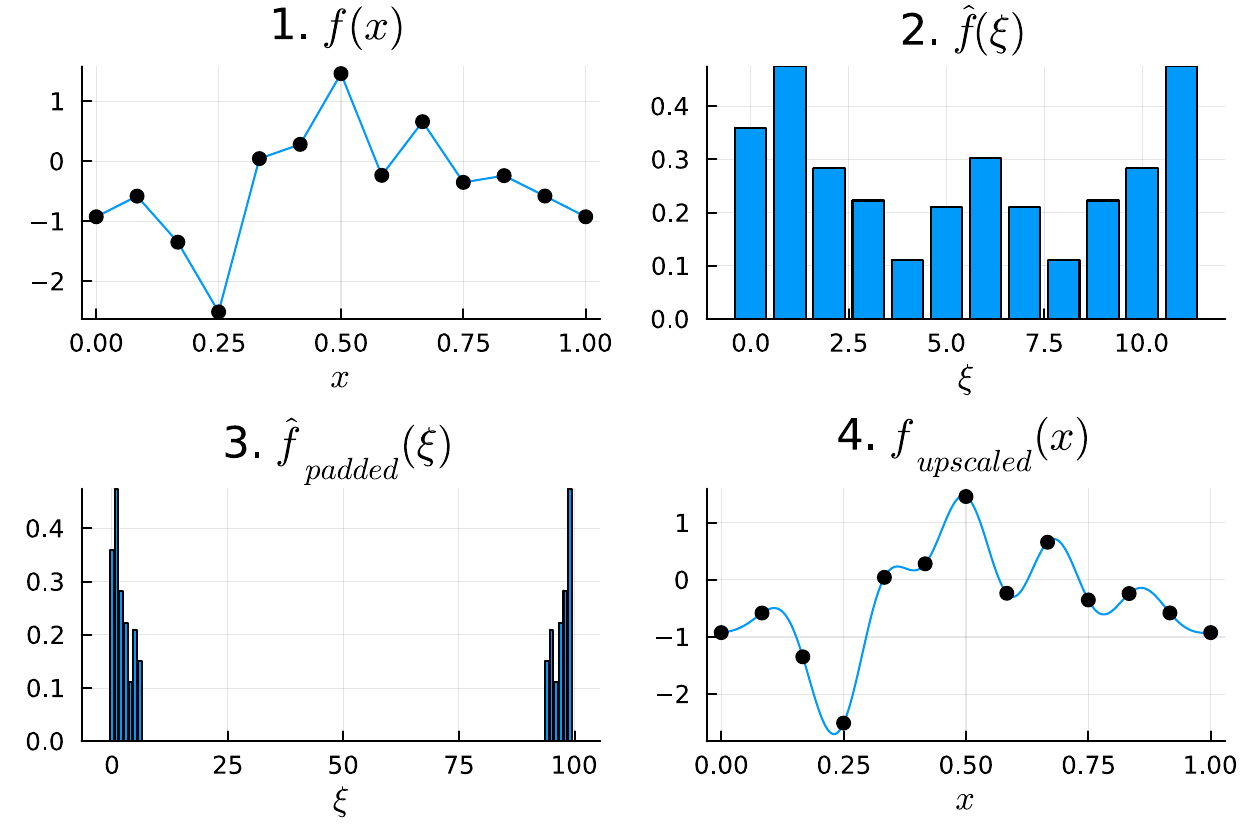}
    \caption{An overview of the interpolation process as done in CNOs. A function on a coarse grid (1) is Fourier transformed (2). Then, zeros are inserted for the high-frequency coefficients (3). Finally, performing an inverse FFT produces an interpolant on a fine grid (4).}
    \label{fig:cno-decode}
\end{figure}

\begin{figure}[htb]
    \centering
    \includegraphics[width=0.4\linewidth]{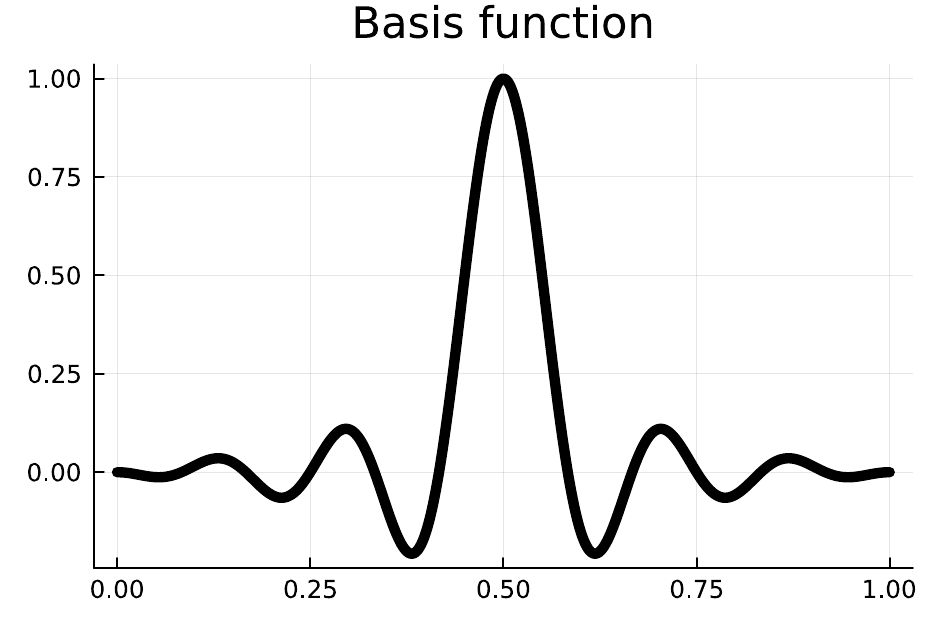}
    \includegraphics[width=0.4\linewidth]{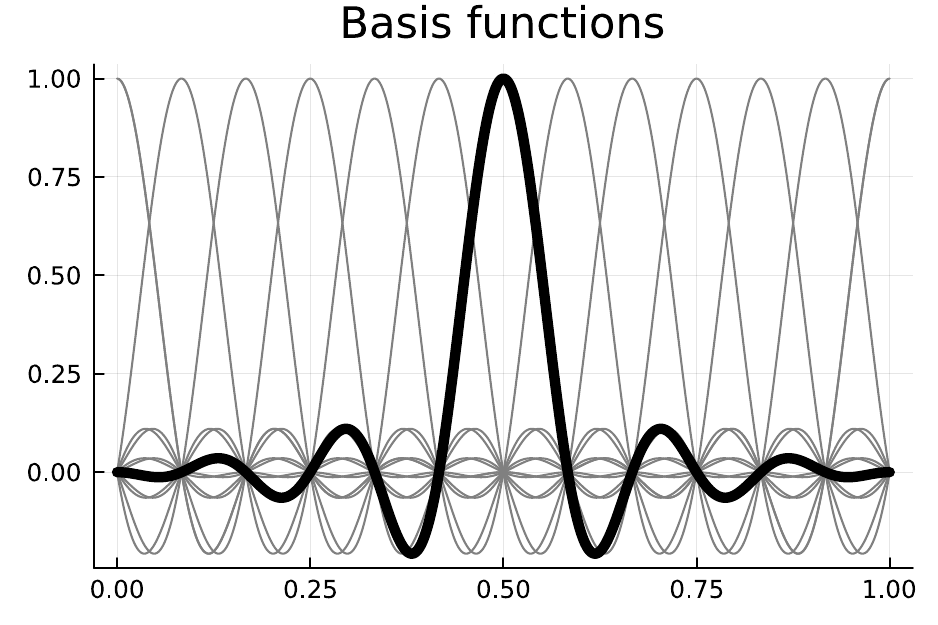}
    \caption{Left: a single Dirichlet kernel basis function. Right: a full function basis, showing that these basis functions are nodal and form a partition of unity.}
    \label{fig:cno-basis}
\end{figure}

Figures~\ref{fig:cno-encode}~and~\ref{fig:cno-decode} show the encoding and decoding process as done in CNOs. Figure~\ref{fig:cno-basis} shows the function basis corresponding to these encoding and decoding processes. This shows that CNOs use a specific function basis, which can actually be changed based on the problem. For example, while CNOs assume periodic domains, their Dirichlet kernel basis can be replaced by another basis, such as a B-spline basis, that conforms to the topology of the domain.

\end{document}